\documentclass{article}
\usepackage[utf8]{inputenc}

\usepackage[final,eandd]{neurips_2026}

\usepackage[utf8]{inputenc} %
\usepackage[T1]{fontenc}    %
\usepackage{hyperref}       %
\usepackage{url}            %
\usepackage{booktabs}       %
\usepackage{amsfonts}       %
\usepackage{nicefrac}       %
\usepackage{microtype}      %
\usepackage{xcolor}         %
\usepackage{amsmath}        %
\usepackage{graphicx}       %
\usepackage{subcaption}     %
\usepackage{bbm}
\usepackage{multirow}       %
\usepackage{makecell}       %
\usepackage{rotating}       %

\usepackage{_macros}        %
\setcitestyle{authoryear,round}
\usepackage{microtype}      %
\usepackage{comment}

\usepackage[fixed]{fontawesome5}

\usepackage{wrapfig}
\usepackage{etoc}
\definecolor{methodDirect}{HTML}{0072B2}
\definecolor{methodCode}{HTML}{D55E00}
\definecolor{methodSelfDistillation}{HTML}{009E73}
\definecolor{methodIntrinsic}{HTML}{E69F00}
\definecolor{methodRanking}{HTML}{CC79A7}
\definecolor{methodEmbedding}{HTML}{56B4E9}
\definecolor{methodPreTrained}{HTML}{999999}
\newcommand{\methodcolorsquare}[1]{\textcolor{#1}{\rule{0.55em}{0.55em}}}

\usepackage{xcolor}
\usepackage{tcolorbox}
\tcbuselibrary{listings, breakable, skins}

\definecolor{promptframe}{HTML}{4B5563}
\definecolor{promptbg}{HTML}{F8F8F6}
\definecolor{tagSystem}{HTML}{1F4E79}
\definecolor{tagUser}{HTML}{0F766E}
\definecolor{tagAssistant}{HTML}{B45309}
\definecolor{tagThought}{HTML}{6B21A8}
\definecolor{tagAction}{HTML}{B91C1C}
\definecolor{tagAnswer}{HTML}{15803D}
\definecolor{placeholder}{HTML}{C2410C}

\lstdefinestyle{promptlst}{
  basicstyle=\ttfamily\footnotesize,
  breaklines=true, breakindent=0pt, breakautoindent=false,
  columns=fullflexible, keepspaces=true, showstringspaces=false,
  upquote=true,
  escapeinside={(*@}{@*)},
  moredelim=[s][\color{tagSystem}\bfseries]{<system>}{</system>},
  moredelim=[s][\color{tagUser}\bfseries]{<user>}{</user>},
  moredelim=[s][\color{tagAssistant}\bfseries]{<assistant>}{</assistant>},
  moredelim=[s][\color{tagThought}\bfseries]{<thought>}{</thought>},
  moredelim=[s][\color{tagAction}\bfseries]{<action>}{</action>},
  moredelim=[s][\color{tagAnswer}\bfseries]{<answer>}{</answer>},
  moredelim=[s][\color{tagAnswer}\bfseries]{<explanation>}{</explanation>},
}

\newtcblisting{prompt}[2][]{%
  enhanced, breakable, listing only,
  listing engine=listings,
  listing style=promptlst,
  colback=promptbg, colframe=promptframe,
  colbacktitle=promptframe, coltitle=white,
  fonttitle=\bfseries\small\sffamily,
  title={#2},
  boxrule=0pt, leftrule=2.5pt,
  toprule=0pt, rightrule=0pt, bottomrule=0pt,
  arc=1mm, outer arc=1mm,
  attach boxed title to top left={xshift=-1pt, yshift=-1pt},
  boxed title style={enhanced, sharp corners=west, arc=1mm,
                     boxrule=0pt, frame hidden},
  left=2.5mm, right=2mm, top=1.2mm, bottom=1.2mm,
  before skip=4pt, after skip=4pt,
  #1
}

\newtcblisting{completion}[2][]{%
  enhanced, breakable, listing only,
  listing engine=listings,
  listing style=promptlst,
  colback=promptbg!50!white, colframe=tagAssistant,
  colbacktitle=tagAssistant, coltitle=white,
  fonttitle=\bfseries\small\sffamily,
  title={#2},
  boxrule=0pt, leftrule=2.5pt,
  toprule=0pt, rightrule=0pt, bottomrule=0pt,
  arc=1mm, outer arc=1mm,
  attach boxed title to top left={xshift=-1pt, yshift=-1pt},
  boxed title style={enhanced, sharp corners=west, arc=1mm,
                     boxrule=0pt, frame hidden},
  left=2.5mm, right=2mm, top=1.2mm, bottom=1.2mm,
  #1
}

\providecommand{\promptslot}[1]{%
  \textcolor{placeholder}{\{\textnormal{\itshape\detokenize{#1}}\}}%
}

\title{\benchname: Evaluating Dense Supervision Signals for Long-Horizon LLM Agents}

\definecolor{darkblue}{rgb}{0, 0, 0.5}
\hypersetup{colorlinks=true, citecolor=darkblue, linkcolor=darkblue, urlcolor=darkblue}

\newcommand{\affil}[1]{\textsuperscript{\normalfont #1}}

\author{
Sergio Hernández-Gutiérrez\affil{1} \quad
Matteo Merler \affil{2}\thanks{Equal Contribution \qquad $^\dagger$Equal Advising \qquad Correspondence to: {\small\texttt{sergio.hernandez@bethgelab.org}}} \quad
Ilze Amanda Auzina\affil{1}$^*$ \\
\textbf{Joschka Strüber}\affil{1} \quad
\textbf{Ameya Prabhu}\affil{1}$^\dagger$ \quad
\textbf{Matthias Bethge}\affil{1}$^\dagger$ \\[4pt]
\affil{1}Tübingen AI Center, University of Tübingen \quad
\affil{2}Fondazione Bruno Kessler
}

\begin{document}

\maketitle
\vspace{-0.75cm}
\begin{center}
    \begin{tabular}{c@{\hskip 19pt}c}
    \raisebox{-1pt}{\faGlobe} \href{https://q-val.com}{\texttt{Website}}
    \hspace{1.0cm}
    \raisebox{-1pt}{\faGithub} \href{https://github.com/bethgelab/qval}{\fontsize{8.8pt}{0pt}\texttt{Code}}
    \hspace{1.0cm}
    \raisebox{-1.5pt}{\faDatabase}\href{https://huggingface.co/datasets/bethgelab/qval/}{\fontsize{8.8pt}{0pt} \texttt{Datasets}} \\
\end{tabular}
\end{center}

\begin{abstract}

LLM agents increasingly act over long horizons, where outcome-only rewards say little about which intermediate actions were good.
Dense supervision methods address this by scoring intermediate steps, but they are usually evaluated through the downstream performance of a full training pipeline.
This is expensive, confounds signal quality with other training choices, and means that methods are rarely compared on common ground.
We introduce \benchname, a training-free testbed that measures the $Q$-alignment of a dense signal: whether it orders the candidate actions in each state the same way as their returns under a reference policy do.
We benchmark 21 methods from seven families with six open-weight backbones in four environments.
Simple prompting baselines are the most $Q$-aligned on average, and performance clusters strongly by family.
Code-generated signals do well only in small state and action spaces; embedding-based signals are weak; self-distillation signals are anti-aligned; and no method is $Q$-aligned in open-ended environments.
When five of these signals guide an actor at test time by choosing among four proposed actions, $Q$-alignment strongly correlates with the gain in task success.
\benchname is designed to be easily extensible to new environments and methods, enabling researchers to compare new signals cheaply before committing to a training run.

\end{abstract}

\begin{figure}[h!]
   \hspace{-0.65cm}\includegraphics[width=1.1\linewidth]{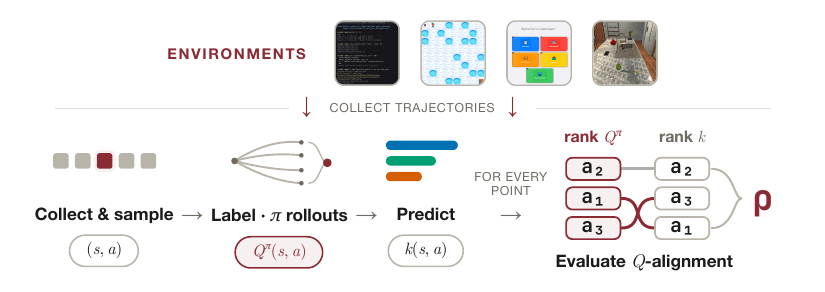}
   \caption{\benchname \textbf{design pipeline}. 
    We collect trajectories in multi-turn environments, sample candidate state--action pairs, and label them with estimated $Q$-values with respect to a reference policy $\pi$. We make each method predict action scores and measure $Q$-alignment between the predicted scores and the labels. \benchname isolates learning-signal quality before any downstream RL training.}
\label{fig:main}
\vspace{-0.2em}
\end{figure}

\section{Introduction}
\label{sec::introduction}

Large language models (LLMs) increasingly act as agents that write code, operate graphical interfaces, and navigate simulated environments.
These are long-horizon tasks, where a single trajectory can span hundreds or thousands of actions. 
Sparse rewards make learning intractable as horizons grow: an outcome-based reward gives little guidance about the goodness of individual steps and, more critically, it may never be observed if the task is beyond the frontier of the model's capabilities. 
Current reinforcement learning (RL) algorithms for LLM post-training primarily rely on sampling-based value estimation; for example, GRPO~\citep{shao2024deepseekmath} estimates the value of a completion by comparing samples within a group. 
This works best when trajectories are short, where only a reasonable number of samples is needed to attribute which actions caused the outcome. 
However, agent trajectories increasingly involve multi-step tool use, recursive decomposition, and context compaction; 
as trajectories lengthen, outcome rewards offer little guidance and group-based estimators do not distinguish individual actions’ contributions.

\enlargethispage{\baselineskip}
This has motivated methods that produce dense supervision signals. These include signals derived from token probabilities along reasoning traces~\citep{yoon2026pacr}, tool calls~\citep{xie2026tips}, or interaction feedback~\citep{auzina2026intrinsic}, as well as self-distillation approaches, which derive intermediate supervision from model-generated judgments or targets~\citep{hubotter_reinforcement_2026, shenfeld_self-distillation_2026, song_expanding_2026}.
Although these methods differ in how they construct their scores, and have largely been developed and evaluated in isolation, they all share the goal of assigning useful values to intermediate states or actions. 
We therefore study them as a common class of \emph{dense supervision methods} and group them into families according to how each method obtains its signal.

The primary bottleneck towards comprehensive evaluation is that we lack cheap and principled frameworks to compare dense supervision signals. Today, a method is evaluated by integrating it into a post-training pipeline and measuring the downstream performance improvement. This is expensive and often unachievable due to the distinct setups required by different methods. 
It further makes the results hard to interpret: the measured gain conflates the supervision quality and other engineering choices used for RL training, such as algorithmic or optimization features, normalization techniques, and balancing with other training signals. 
We pose the following research question: 
\begin{quote}
\centering
\textit{Can we evaluate dense supervision signals in isolation,\\ before any expensive post-training runs?}
\end{quote}
We introduce \benchname, a cheap, training-free testbed for dense supervision methods. This enables researchers to iterate on new approaches and their variants quickly, evaluating candidate signals before integrating them into guided search systems and post-training pipelines.
\benchname has a simple design, shown in Figure~\ref{fig:main}. For a given environment, we construct a dataset where each sample is a pair of state and candidate actions $(s, A)$ labelled with a \emph{reference $Q$-value} for each $(s, a \in A)$: the expected return of the trajectory that continues from $(s, a)$ under an expert \emph{reference policy}.
We use optimal policies or expert planners where available, and frontier models otherwise. 
We evaluate each method by its \emph{$Q$-alignment}: how well its predicted scores order the candidate actions relative to the reference $Q$-values.
Furthermore, we validate that $Q$-alignment is indicative of downstream performance by implementing an online guided search inference experiment, where the actor is guided by a subset of the methods we benchmark.

We instantiate this methodology as a benchmark, which at release covers four multi-turn environments across text and visual domains: programming and agentic terminal interaction in TerminalBench~\citep{merrill_terminal-bench_2025}, computer and application use in OpenApps~\citep{ullrich_openapps_2025}, embodied reasoning in ALFWorld~\citep{shridhar_alfworld_2020}, and goal-directed navigation in FrozenLake~\citep{towers_gymnasium_2025}. 
We use our datasets sampled from these environments to evaluate 21 methods spanning seven methodological families: direct prompting~\citep{liu_g-eval_2023, ma_vision_2024}, intrinsic signals~\citep{auzina2026intrinsic, kwok2026llmverifier}, code generation~\citep{ma_eureka_2023, li_auto_2024}, self-distillation~\citep{hubotter_reinforcement_2026}, ranking-based prediction, pretrained models~\citep{ma2022vip, ma2023liv}, and embedding similarity~\citep{rocamonde2024visionlanguage, baumli2023vision}.

We make the following core observations. First, $Q$-alignment shows that simple prompting methods perform best on average. Second, we observe that $Q$-alignment induces a clustering of dense supervision methods that corresponds with their methodological families. Third, $Q$-alignment generally weakens as environments become more open-ended. Fourth, in environments with both modalities, text and vision, methods recover reference values less reliably from images (Section~\ref{sec::modality_ablation}). Lastly, the ranking of dense supervision methods induced by \benchname strongly correlates with downstream performance in an online setting. \benchname enables practitioners to evaluate their own novel methods, facilitating the process of dense supervision signal design.

Overall, our contributions are as follows. 
First, we introduce \benchname, a training-free testbed that evaluates dense supervision methods by their \emph{$Q$-alignment} (Section~\ref{sec::qval}).
Second, we instantiate \benchname as a benchmark and evaluate 21 dense supervision methods across six open-weight model backbones scaling from 9B to 122B parameters (Sections~\ref{sec::benchmark} and~\ref{sec::experiments::results}), which shows when and which dense supervision methods work best. The benchmark is built to grow: it can easily be extended to new environments and methods, so they can be fairly evaluated alongside existing ones.
Third, on state--action pairs collected from guided search trajectories, the obtained method ranking aligns with downstream test-time selection success, verifying that \benchname provides meaningful signal for dense supervision method evaluation.

\section{\benchname: a training-free testbed for dense supervision methods}
\label{sec::qval}

We study supervision methods intended to densify an environment's sparse outcome reward. A single long-horizon trajectory contains many actions, and one that looks reasonable in isolation can still make the final goal harder to reach. For such signals, scores should reflect \emph{where a decision leads} rather than how good it looks locally.
\benchname asks exactly this: does a method assign higher scores to the actions that make eventual success more likely?

\textbf{Setup.}
We consider an environment modelled as a Markov decision process~\citep{bellman_markovian_1957}, with state space $\mathcal{S}$, action space $\mathcal{A}$, transition distribution $T(s'\mid s,a)$, reward function $r:\mathcal{S}\times\mathcal{A}\to\mathbb{R}$, and discount factor $\gamma\in[0,1]$. At each step the agent
observes a state $s\in\mathcal{S}$, takes an action $a\sim\pi(\cdot\mid s)$, receives reward $r(s,a)$, and the environment transitions to $s'\sim T(\cdot\mid s,a)$. A trajectory
$\tau=(s_0,a_0,r_0,s_1,\dots)$ is a realization of this process, with return $G(\tau)=\sum_{t\ge0}\gamma^t r_t$.
A \emph{policy} $\pi(a\mid s)$ maps states to distributions over actions. The \emph{$Q$-value function} of a policy $\pi$,
\begin{equation}
Q^\pi(s,a)=\mathbb{E}_{\tau\sim\pi}\!\left[G(\tau)\mid s_0=s,\,a_0=a\right],
\label{eq:qvalue}
\end{equation}
gives the expected return from state $s$ after committing to action $a$, and afterwards continuing to behave following $\pi$~\citep{sutton_reinforcement_2018}.
The analogous \emph{state-value function} $V^\pi(s)=\mathbb{E}_{\tau\sim\pi}[G(\tau)\mid s_0=s]$ scores a state without committing to a first action.

\textbf{Reference policies.}
$Q^\pi$ is only defined once a \emph{reference} continuation policy $\pi$ is fixed. \benchname uses $Q^\pi$ similarly to how a supervised dataset uses labels: it annotates each decision point with a reference value and evaluates a method based on how well its scores reproduce their ordering.
Importantly, $\pi$ labels state--action pairs; it need not be the policy later trained or guided by the signal. We choose strong reference policies, ideally optimal ones, to reduce the risk that good actions receive low values because of a poor continuation.
Outcome rewards say nothing about intermediate behaviour; $r_t=0$ at every non-terminal step, so an action's value is determined entirely by how $\pi$ rolls out the trajectory.
We do not assume a reference policy is available when a dense signal is later deployed, only that we can construct one here, in a controlled setting, to obtain trustworthy labels.
Section~\ref{sec::benchmark} describes how we obtain the reference policy $\pi$ for each chosen environment.

\textbf{$Q$-aligned signals.}
A dense supervision method assigns a score $k(s,a)$ to each state--action pair. 
To keep the original task signal, $k$ must complement $r$ rather than replace it, e.g., as a potential-based shaping term $\gamma k(s',a')-k(s,a)$~\citep{ng_policy_1999, wiewiora_principled_2003}.
\benchname studies whether, within a state, actions with higher reference $Q$-values $Q^\pi(s,a)$ receive higher scores. Formally, we call $k$ \emph{$Q$-aligned} under $\pi$ if, for every state $s$ and actions $a,b \in A$,
\begin{equation}
Q^\pi(s,a)>Q^\pi(s,b)\ \Longrightarrow\ k(s,a)>k(s,b).
\label{eq:q-aligned}
\end{equation}
This condition does not require scores to be comparable across states or equal-value actions to receive identical scores. We also provide measurements of $Q$-alignment across states (global $Q$-alignment) in Appendix~\ref{app::global_alignment}.

$Q$-alignment therefore offers a cheap way to assess whether a dense signal encodes the outcome-related guidance it is intended to provide during training or guided search. \benchname measures this property without a training run, separating alignment from the choices involved in integrating the signal into a downstream pipeline. Our guided search experiments show that $Q$-alignment measured on deployment-relevant decisions strongly predicts task success (Section~\ref{sec::guided_search}).

Appendix~\ref{app::value_targets} reports evaluations against reference state values $V^\pi(s)$.

\paragraph{Evaluating $Q$-alignment.}
$Q$-alignment (Eq.~\ref{eq:q-aligned}) describes an ideal: candidate actions with higher $Q^\pi(s,a)$ receive higher scores. In practice, we measure how closely a signal approaches this ideal using Spearman correlation $\rho$~\citep{spearman_proof_1904}, standard for meta-evaluating scorers against reference judgments~\citep{liu_g-eval_2023}. Predictions live on incompatible scales (e.g., raw LLM scores, code-generated outputs, token log-probabilities, embedding distances). We therefore compare the action orderings they induce, not their numerical agreement with $Q^\pi$. This is consistent with recent evidence that LLM and VLM judges order candidates reliably even when their absolute scores are poorly calibrated~\citep{kumar_vlm_2026}. Spearman correlation is unchanged by strictly increasing transformations of a signal's scores within a state.

For each state $s_i$, let $\mathbf{k}_i$ and $\mathbf{q}_i$ contain the candidate scores and their reference $Q^\pi$-values. We compute Spearman correlation within each state and average across states:
\begin{equation}
\rho_{\mathrm{local}}=\frac{1}{N}\sum_{i=1}^N \rho_{\mathrm{S}}(\mathbf{k}_i,\mathbf{q}_i),
\label{eq:local_spearman_main}
\end{equation}
where $\rho_{\mathrm{S}}$ denotes Spearman's rank correlation and $N$ is the number of valid state-level comparisons.

We employ additional metrics to gain insights into specific signal behaviors and properties of the data. Selection loss can complement $Q$-alignment by measuring how much value is potentially lost by picking the action with the highest score predicted by a given method; this metric can detect suboptimal behaviors during guided search, as explained in Section~\ref{sec::guided_search}. As shown in Figure~\ref{fig:calibration}, mean squared error and its affine-corrected counterpart can indicate whether normalization is needed before post-training runs. Other statistics can serve as useful context to understand the impact of the datasets and the environments on \benchname results: the number of label ties, the frequency of top-scores assigned to invalid actions, and the frequency at which a method produces unusable scores (e.g., due to failed runs or unparseable output). Detailed definitions are provided in Appendix~\ref{app::eval_metrics}.

\section{Benchmarking dense supervision methods with \benchname}
\label{sec::benchmark}

We instantiate the \benchname methodology as a benchmark, initially employing four environments and evaluating 21 dense supervision methods, designed to be extensible beyond its initial scope. We provide an overview of the datasets and methods next, with detailed descriptions of each environment in Appendix~\ref{app::environments} and each method in Appendix~\ref{app::methods}.

\subsection{Datasets}
\label{sec::datasets}

\paragraph{Environments.}
We choose four environments that vary in action-space structure, observation modality, and the amount of context needed to evaluate an action. 
The suite covers goal-directed navigation in FrozenLake~\citep{towers_gymnasium_2025}, embodied reasoning in ALFWorld~\citep{shridhar_alfworld_2020}, browser-based computer use in OpenApps~\citep{ullrich_openapps_2025}, and terminal-based problem solving in TerminalBench~\citep{merrill_terminal-bench_2025}. 
FrozenLake has four discrete actions, whereas TerminalBench accepts open-ended shell commands over rich textual observations. TerminalBench is text-only; the other environments provide both textual and visual observations. For TerminalBench, we use a subset of tasks from TBLite~\citep{OpenThoughts-TBLite}.

\textbf{Data collection.} For each environment, we collect trajectories and sample $N$ sets of state--action pairs in total (Table~\ref{tab:env_summary}, Appendix~\ref{app::environments}); each set consists of a single state and four candidate actions. 
We do not aim to maximize task success during collection; instead, we prioritize diversity, including both high- and low-value states and actions. 
For OpenApps and FrozenLake, we use scripted policies designed to be suboptimal to maximize coverage. 
For TerminalBench and ALFWorld, we generate trajectories with DeepSeek v3.2~\citep{deepseekai2025deepseekv32}. 
To further improve diversity, we select a limited set of state--action pairs from a range in the middle of each trajectory. 
This avoids early states with little task progress and late states where most actions have similar value. 

\textbf{Data labeling.} We obtain a reference action value for every candidate at each sampled state. To evaluate a candidate action $a$, we restore its state $s$, execute $a$, then continue under the reference policy $\pi$ and record the discounted return.
For stochastic references, we take the maximum observed return over repeated continuations as a proxy for near-optimal continuation value (Max-Value Monte Carlo, MVMC), rather than estimating the mean return under $\pi$. Appendix~\ref{app::mean_max} compares these targets for TerminalBench.
The reference policy $\pi$ is environment-specific. For OpenApps and FrozenLake, we use scripted optimal policies, and for ALFWorld, we use an expert planner. 
In TerminalBench, however, an optimal policy is intractable. We therefore estimate TerminalBench values via MVMC rollouts ($k=16$) with GPT-5.5~\citep{openai_gpt-55_2026} as a backbone. We verify that this creates a strong continuation policy: it reaches $100\%$ Pass@16 on our TerminalBench subset, and we further compare it with Claude Opus 4.7~\citep{anthropic2026claudemodels} in Appendix~\ref{app::reference_backbone}. To check that 16 continuations are enough, we re-estimate 100 of these labels with 128 continuations. The global $Q$-alignment matches the one at $k=16$ ($\rho=0.983$), with one swap in the middle of the table and the same best method. A label computed from 16 continuations varies with a standard deviation of 0.063, against 0.18 across all labels. Therefore, the labels provide a stable evaluation setup at $k=16$.
We also provide a reference state value $V(s_t)$ for FrozenLake, ALFWorld, and OpenApps, estimated with the same reference policies but without forcing the first action, to test for robustness to the choice of target value (Appendix~\ref{app::value_targets}).

\subsection{Dense supervision methods}
\label{sec::experiments::methods}

\begin{wraptable}{r}{0.55\textwidth}
\centering
\footnotesize
\vspace{-0.45cm}
\caption{\textbf{Benchmarked methods}. We group methods by the signal used to score state--action pairs and report each method source.}
\label{tab:method-groups}
\begin{tabular}{ll}
\toprule
Method group & Method name (source) \\
\midrule
Ranking & \methodcolorsquare{methodRanking}~\texttt{ranking} (Baseline) \\
\midrule
Direct & \methodcolorsquare{methodDirect}~\texttt{direct-16} (Baseline) \\
 & \methodcolorsquare{methodDirect}~\texttt{direct-batched} (Baseline) \\
 & \methodcolorsquare{methodDirect}~\texttt{direct-sequential} (Baseline) \\
 & \methodcolorsquare{methodDirect}~\texttt{direct-single} (Baseline) \\
 & \methodcolorsquare{methodDirect}~\texttt{gvl}~\citep{ma_vision_2024} \\
\midrule
Intrinsic scoring & \methodcolorsquare{methodIntrinsic}~\texttt{verifier}~\citep{kwok2026llmverifier} \\
 & \methodcolorsquare{methodIntrinsic}~\texttt{\(\Delta\)belief}~\citep{auzina2026intrinsic} \\
\midrule
Self-distillation & \methodcolorsquare{methodSelfDistillation}~\texttt{sdpo}~\citep{hubotter_reinforcement_2026} \\
 & \methodcolorsquare{methodSelfDistillation}~\texttt{sdpo-gt} (Extension) \\
\midrule
Pretrained & \methodcolorsquare{methodPreTrained}~\texttt{vip}~\citep{ma2022vip} \\
 & \methodcolorsquare{methodPreTrained}~\texttt{liv-cos}~\citep{ma2023liv} \\
 & \methodcolorsquare{methodPreTrained}~\texttt{liv-l2}~\citep{ma2023liv} \\
 & \methodcolorsquare{methodPreTrained}~\texttt{liv-txt}~\citep{ma2023liv} \\
\midrule
Embedding & \methodcolorsquare{methodEmbedding}~\texttt{vlm-sor-softmax}~\citep{baumli2023vision} \\
 & \methodcolorsquare{methodEmbedding}~\texttt{vlm-sor}~\citep{baumli2023vision} \\
 & \methodcolorsquare{methodEmbedding}~\texttt{vlm-rm-cos}~\citep{rocamonde2024visionlanguage} \\
 & \methodcolorsquare{methodEmbedding}~\texttt{vlm-rm}~\citep{rocamonde2024visionlanguage} \\
\midrule
Code & \methodcolorsquare{methodCode}~\texttt{eureka}~\citep{ma_eureka_2023} \\
 & \methodcolorsquare{methodCode}~\texttt{codegen}~\citep{li_auto_2024} \\
 & \methodcolorsquare{methodCode}~\texttt{codegen-avg} (Extension) \\
\bottomrule
\end{tabular}
\vspace{-0.45cm}
\end{wraptable}

Dense supervision methods can induce different signals and employ available information in varied ways.
Using \benchname, we evaluate 21 dense supervision methods and group them into seven families by the information they use to score state--action pairs, summarized in Table~\ref{tab:method-groups}. 
The methods include direct implementations of prior work, adaptations of prior methods to our setting, as well as additional baselines introduced in this paper. The latter serve either as simple baselines or as stronger probes within a method family. 
We next describe each family briefly (details are provided in Appendix~\ref{app::methods}).

\textbf{Ranking methods} prompt an LLM to directly compare a set of candidate actions from the same state. This group contains \texttt{ranking}, a direct LLM action-ranking baseline.

\textbf{Direct methods} prompt an LLM or VLM to output a numeric value for a datapoint. This is reminiscent of the LLM-as-judge numeric-scoring paradigm~\citep{liu_g-eval_2023}. 
The simplest variant uses one datapoint per prompt (\texttt{direct-single}). 
We also test variants that predict multiple datapoints at once, with points from the same environment but not necessarily from the same trajectories. 
The \texttt{direct-batched} variant provides multiple datapoints in a single prompt, while \texttt{direct-sequential} provides multiple datapoints in a multi-turn format, appending new datapoints after the previous answer. The purpose of both variants is to test whether scoring multiple points at once helps to ground each score in a common scale.
Another variant averages 16 independent estimates for the same datapoint (\texttt{direct-16}). 
We also adapt \texttt{gvl}~\citep{ma_vision_2024} to text-based environments and value prediction by giving the model a shuffled full trajectory around the target transition before asking for a value.

\textbf{Intrinsic scoring methods} derive scores from the model's own confidence rather than from an explicit value estimate. 
We adapt \texttt{$\Delta$belief}~\citep{auzina2026intrinsic}, which scores an action by the change in the probability the model assigns to eventual success once the
action's outcome is observed. 
We also adapt LLM-as-a-Verifier~\citep{kwok2026llmverifier} (\texttt{verifier}), which prompts the model to score a $(s,a,s')$ tuple on a rubric of per-environment quality criteria such as correctness, efficiency, and error-freeness, using the probabilities it assigns across an ordered grading scale and averaging them into a scalar score.

\textbf{Self-distillation methods} score a candidate action by how much more likely the model
is to have produced that action once it sees additional privileged information about the action's outcome.
The intuition is that a good action becomes more probable in hindsight when its favorable
outcome is shown, whereas a poor action does not. This differs from the intrinsic scoring methods, which read the model's probability for a dedicated prompt asking about the agent's success rather than over the action itself. 
The family contains two offline re-ranking signals: \texttt{sdpo}~\citep{hubotter_reinforcement_2026}, where the additional information is the candidate's immediate next state, and \texttt{sdpo-gt}, an oracle ablation that additionally reveals the next expert action and an outcome summary from a stored expert continuation.

\textbf{Pretrained methods} use fixed representations from value-pretrained vision-language models. \texttt{Vip}~\citep{ma2022vip} scores an image state by its negative distance to a goal-image embedding. \texttt{Liv}~\citep{ma2023liv} contributes three variants: \texttt{liv-cos}, which uses the paper's original cosine similarity score with an image goal; \texttt{liv-l2}, which uses a \texttt{vip}-style negative Euclidean distance to an image goal; and \texttt{liv-txt}, which compares the state image embedding to a textual goal embedding.

\textbf{Embedding methods} use frozen vision-language encoders to score image states by their similarity to a text goal. 
\texttt{Vlm-rm}~\citep{rocamonde2024visionlanguage} contributes \texttt{vlm-rm-cos}, which scores a state by cosine similarity between its image embedding and the goal-text embedding, and \texttt{vlm-rm}, which projects out baseline visual features before measuring progress toward the goal. \texttt{Vlm-sor}~\citep{baumli2023vision} contributes \texttt{vlm-sor-softmax}, which uses the softmax probability assigned to the target goal among negative goals, and \texttt{vlm-sor}, which thresholds that probability into a binary success reward.

\textbf{Code methods} prompt an LLM to generate executable Python code for a scoring function with state--action pairs as input. The code is generated once and then executed to produce predicted scores for all datapoints. 
The family contains three methods that differ in how code is generated. \texttt{Codegen}, our adaptation of Auto MC-Reward~\citep{li_auto_2024}, generates $k$
candidate functions independently (reporting per-function correlation), and \texttt{codegen-avg} averages their predictions first, and then reports correlation. \texttt{Eureka}~\citep{ma_eureka_2023} instead refines the function iteratively:
Across several rounds, an LLM judge scores the previous round's candidates and the best is used as seed for the next, along with feedback produced by the same LLM judge and program-execution diagnostics.

\subsection{Experimental details}

\textbf{Backbones.} We fix the LLM/VLM backbone across methods wherever possible, so that differences in $Q$-alignment reflect the scoring methods themselves rather than the underlying models' performance.
We also prioritize open-weights models with accessible internal information, so that methods requiring hidden states or token log-probabilities can be evaluated. We leverage the Qwen3.5 family~\citep{qwen35blog} at 9B, 27B, 35B-A3B, and 122B-A10B parameter scales, and the Gemma 4 family~\citep{google_gemma_2026} at 26B-A4B and 31B parameter scales. The \texttt{vip} and \texttt{liv} methods use their corresponding pretrained models. The \texttt{vlm-rm} and \texttt{vlm-sor} methods use CLIP~\citep{radford_learning_2021} and SigLIP~\citep{zhai_sigmoid_2023}. For the visual ablation in Section~\ref{sec::modality_ablation}, we use the same Qwen3.5 and Gemma 4 models. Appendix~\ref{app::methods} specifies model instantiations and sampling parameters. 

\textbf{Contextual information.} To ensure a fair comparison, we give prompt-based methods the same context: a high-level task description, the same environment dynamics, a textual reward specification, and descriptions of the state and action spaces. Appendix~\ref{app::methods} provides full prompts and parameterization details. 
In visual domains, we keep the same context but provide states as images rather than textual descriptions.

\textbf{Guided search: downstream performance.} To relate $Q$-alignment to downstream performance, we use a signal to guide an actor at test time. At each step, the actor independently proposes four actions, the signal scores them, and the highest-scoring action is executed. This selection uses only the order of the scores within a state, which is what $Q$-alignment measures. The unguided baseline is the same actor executing one sampled action per step. We run five methods (\texttt{direct-single}, \texttt{ranking}, \texttt{verifier}, \texttt{codegen-avg} and \texttt{sdpo}) on FrozenLake (100 maps, Gemma 4 31B) and ALFWorld (24 games with three episodes each, Qwen3.5 27B), and record task success. We compare success with $Q$-alignment on datasets sampled from the state--action pairs visited by the actors (Appendix~\ref{app::guided_search}).

\section{Results}
\label{sec::experiments::results}

\subsection{$Q$-alignment of benchmarked methods}

\begin{figure}[!t]
   \centering
   \vspace{-0.4cm}
   \includegraphics[width=\linewidth]{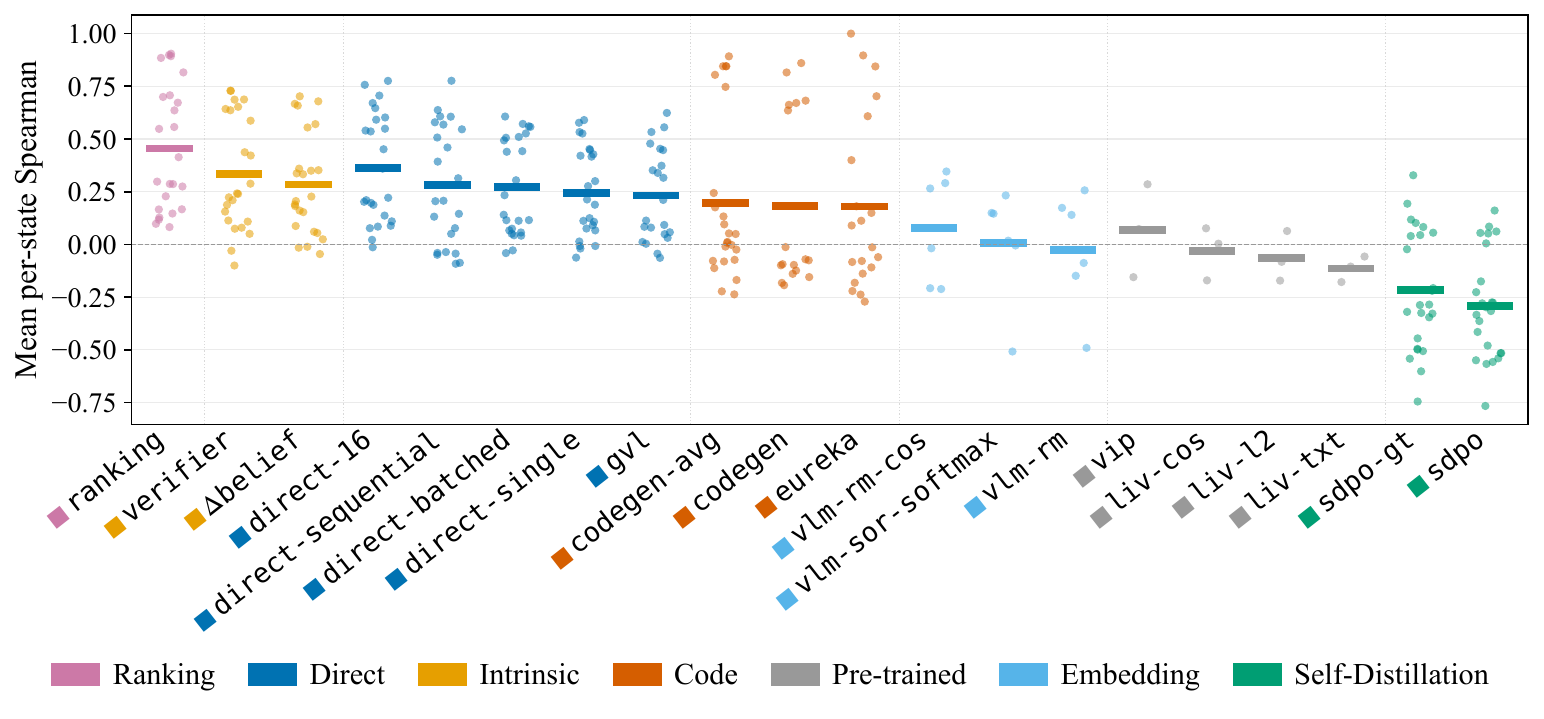}
   \caption{\textbf{Spearman correlations by dense supervision method.} Each point is one environment-model evaluation pair; horizontal bars show means across evaluations. \texttt{Ranking}, \texttt{direct-16}, and \texttt{verifier} methods align best with reference policy values on average. Methods in the same family show similar $Q$-alignment patterns, supporting our method taxonomy.}
   \label{fig:strip-per-cell}
\end{figure}

\begin{figure}
   \includegraphics[width=0.975\linewidth]{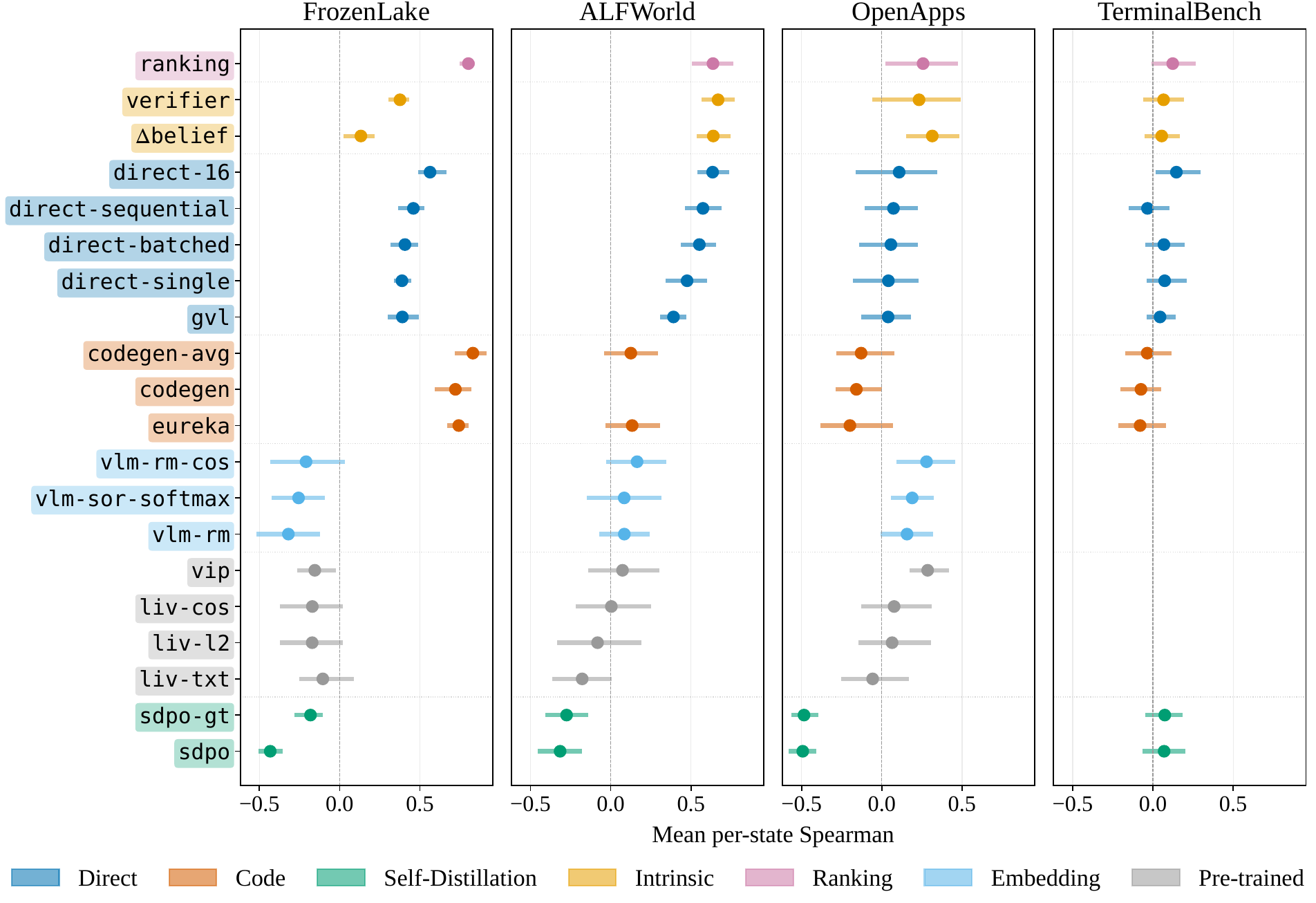}
   \caption{\textbf{Per-method Spearman correlation by environment.}
   Correlations are computed against reference values and averaged across model backbones within each environment; error bars show 95\% confidence intervals.
   Visual methods are not evaluated on TerminalBench.
   Code-based signals show strong environment dependence, and alignment is weak across methods in open-ended tasks.}
   \vspace{-0.4cm}
   \label{fig:forest-per-env}
\end{figure}

\textbf{Simple methods provide strong baselines.}
Figure~\ref{fig:strip-per-cell} shows that \texttt{ranking}, \texttt{direct-16}, and \texttt{verifier} are the strongest approaches overall. Direct value prediction is thus a surprisingly strong baseline for dense supervision.
Methods also cluster by family, obtaining similar correlation ranges within each, which suggests our taxonomy captures real differences in the signal each family extracts.

\textbf{How additional computation is used matters.}
For methods which can use additional computation via multiple estimate generation, averaging improves alignment in 23/24 environment--backbone pairs for \texttt{direct} and 13/17 for \texttt{codegen} (Figure~\ref{fig:strip-per-cell}). In contrast, sequential prompting (\texttt{direct-sequential}) lowers alignment on TerminalBench for 4/5 backbones. 
These results highlight \benchname utility on estimating whether additional compute results in improved signal quality.

\textbf{Environment structure shapes alignment.}
Figure~\ref{fig:forest-per-env} reports correlations per environment, ordered roughly from simpler closed-action settings (FrozenLake) to open-ended ones (TerminalBench) from left to right.
Code-based signals work best when progress can be expressed through compact rules over structured observations. In open-ended environments, a fixed generated function must cover a much broader range of state and action semantics; methods that invoke an LLM at each decision can interpret these inputs directly and are more consistently competitive. Nevertheless, alignment remains weak across methods in the most open-ended tasks we evaluate.

\textbf{Raw method scores can be poorly calibrated}.
$Q$-alignment ignores signal value magnitude, which can affect post-training. Therefore, we also measure calibration via mean squared error (MSE) between the reference values and the method signals. On ALFWorld, \texttt{direct-16} and \texttt{eureka} have raw MSE of $0.07$ and $0.61$, respectively. A least-squares affine fit to the same reference labels reduces these to $0.002$ and $0.003$ (Figure~\ref{fig:calibration}). As such, good $Q$-alignment does not guarantee a suitable scale; nonetheless, affine normalization can largely address this issue for the evaluated methods.

\par\medskip
\noindent\begin{minipage}{\linewidth}
   \centering
   \includegraphics[width=\linewidth]{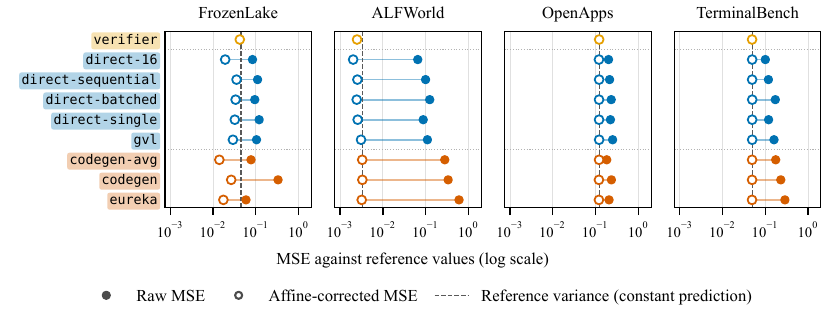}
   \captionof{figure}{\textbf{Raw and affine-corrected MSE against reference values}, averaged over backbones. Filled points show raw MSE for direct and code methods; open points show in-sample affine-corrected MSE. Dashed lines show variance across the full reference set (constant-prediction error). Full results and fitting details are in Appendix~\ref{app::calibration_results}.}
   \label{fig:calibration}
\end{minipage}\par\medskip

\subsection{Can \benchname predict downstream performance?}
\label{sec::guided_search}

\begin{figure}[!ht]
   \centering
   \includegraphics[width=\linewidth]{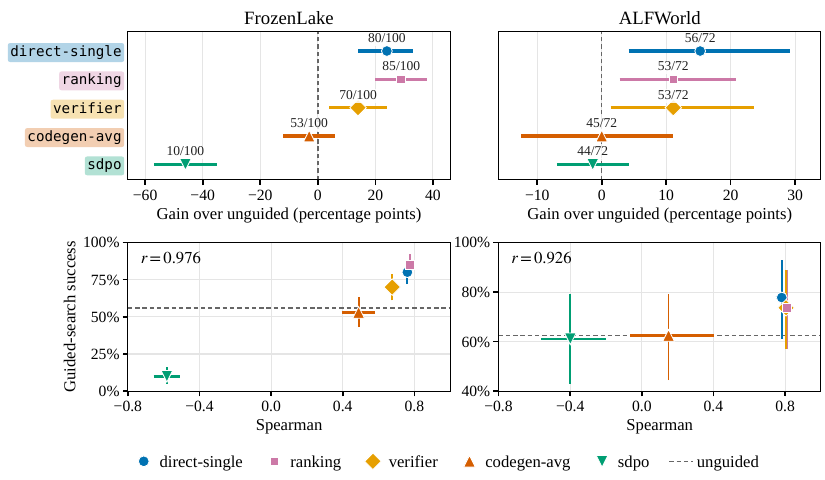}
   \caption{\textbf{Guided search success and its relation to $Q$-alignment.} Top: success gains over the unguided actor; labels give success counts, and bars are paired 95\% bootstrap confidence intervals. Bottom: success against Spearman, with all methods evaluated on the same dataset, built from trajectories collected with and without guidance. Bars show 95\% bootstrap confidence intervals, and $r$ is the Pearson correlation across the five methods.}
   \label{fig:guided_search}
\end{figure}

\textbf{$Q$-alignment predicts downstream performance.}
$Q$-alignment strongly correlates with guided search success across the five methods (Pearson $r=0.976$ in FrozenLake and $r=0.926$ in ALFWorld; Figure~\ref{fig:guided_search}, bottom). In particular, \texttt{direct-single}, \texttt{ranking} and \texttt{verifier} improve task success in both environments (Figure~\ref{fig:guided_search}, top). \texttt{Codegen-avg} shows no clear gain in either environment, and \texttt{sdpo} reduces FrozenLake success to 10\%.
In FrozenLake, \texttt{sdpo} selected an invalid-action candidate in 78\% of groups containing both valid and invalid proposals, suggesting one reason for its poor performance.

\textbf{The state--action distribution matters.}
When using the FrozenLake dataset described in Section~\ref{sec::datasets}, \texttt{codegen-avg} and \texttt{verifier} receive similar $Q$-alignment scores ($\rho=0.787$ and $0.764$), despite achieving 53\% and 70\% guided search success. When evaluated on a dataset with state--action pairs sampled from the trajectories observed in the guided search experiments, their scores are instead $0.489$ and $0.676$, matching their downstream ordering. This highlights the key role that the state--action distribution of the datasets plays in \benchname: $Q$-alignment is best measured on decisions that match the expected situations experienced by downstream actors. Appendix~\ref{app::matching_analysis} provides matching details.

\textbf{Selection loss complements $Q$-alignment.} Spearman scores the full candidate ordering, whereas guided search executes only the top-scoring proposal; selection loss measures the reference value lost by that choice. On ALFWorld, the \texttt{ranking} method achieves a Spearman correlation of 0.81 while picking the optimal action at every step, so its selection loss is zero; its Spearman falls short of 1 only because the method cannot express ties by design. Across the five methods, selection loss tracks guided search success about as closely as Spearman (Pearson $r=-0.977$ in FrozenLake and $r=-0.953$ in ALFWorld; Table~\ref{tab:guided_search_metric_agreement}). Complete selection loss results are provided in Appendix~\ref{app:selection_loss}.

\subsection{Input modality ablation}
\label{sec::modality_ablation}

\noindent
\begin{minipage}[t]{0.50\linewidth}
\vspace{0pt}
We compare $Q$-alignment under text and image representations of the same state, on the environments that provide both. Figure~\ref{fig:modality_ablation} shows Spearman correlations per method--environment--model combination, text on the $x$-axis and image on the $y$-axis, so points below the diagonal favor text, and vice versa. 
The results indicate that the evaluated methods recover reference values more reliably from text than from images.
This suggests that, in our setting, parsing visual information is more challenging, and the potential additional context from it often does not help.
These results must be interpreted through the information available to each method. Text observations often provide compact state abstractions, whereas image-based methods must also infer the relevant state information from pixels. Weaker alignment from images therefore does not show that visual feedback is intrinsically inferior; rather, it highlights that value estimation often requires the right abstraction, especially when progress depends on symbolic, relational, or hidden state information.
\end{minipage}\hfill
\begin{minipage}[t]{0.47\linewidth}
\vspace{0pt}
\centering
\includegraphics[width=\linewidth]{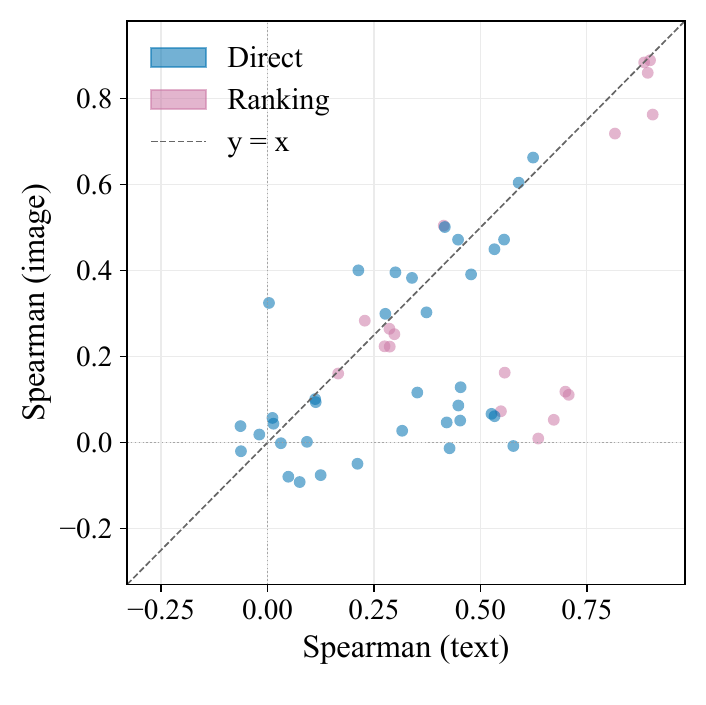}
\captionof{figure}{\textbf{Text vs.\ image observations.} Each point is a method--environment--model evaluation. Points above the diagonal benefit from visual input; points below are hurt by it.}
\label{fig:modality_ablation}
\end{minipage}

\vspace{-0.2cm}
\section{Related works}
\vspace{-0.15cm}
\label{sec::related}

\textbf{Downstream evaluation.}
Most dense feedback methods are evaluated by using the proposed signal inside a downstream training or selection pipeline and reporting task return, or pass-rate improvements. For instance, self-evaluation signals that guide reasoning search are typically validated only by final task accuracy~\citep{xie_self-evaluation_2023}, and generative verifiers are assessed by downstream selection accuracy under a fixed inference budget~\citep{singhi_when_2025}. Although this demonstrates end-to-end utility, it makes signal quality hard to isolate: measured gains depend on the policy model, optimizer, exploration process, environment distribution, amount of generated data, and implementation details of the training loop. Some work evaluates progress or reward-model quality more directly in restricted settings, especially in robotics and visual progress estimation~\citep{ma_vision_2024,budzianowski_opengvl_2026,roy_revisiting_2025}. \benchname is complementary: as a training-free testbed, rather than asking whether a full training recipe improves, it fixes datasets, model backbones, and environment context, then directly measures the $Q$-alignment of a signal with respect to reference $Q$-values.

\textbf{Reward and critic benchmarks.}
There is also a growing literature on evaluating reward models and critic models for language and vision-language systems. RewardBench, RM-Bench, and RewardBench 2 evaluate reward models with response-comparison or accuracy-based tasks spanning chat, reasoning, safety, subtle errors, and style biases~\citep{lambert_rewardbench_2024,liu_rm-bench_2024,malik_rewardbench_2025}. VL-RewardBench and Multimodal RewardBench extend this style of evaluation to multimodal reward models and VLM judges~\citep{li_vl-rewardbench_2025,yasunaga_multimodal_2025}. ProcessBench and PRMBench evaluate models' ability to identify erroneous reasoning steps~\citep{zheng_processbench_2025, song_prmbench_2025}, CriticBench evaluates models' ability to critique and correct solutions~\citep{lin_criticbench_2024}, while AgentRewardBench studies automatic evaluation of complete web-agent trajectories~\citep{lu_agentrewardbench_2025}. These benchmarks are closely related in spirit, but they primarily assess final responses, pairwise preferences, critiques, step-level correctness labels, or whole-trajectory judgments. \benchname instead provides a testbed for dense signals used to train and guide multi-turn agents, measuring their alignment with reference $Q$-values in interactive environments.

\vspace{-0.2cm}
\section{Conclusions}
\vspace{-0.15cm}
\label{sec::conclusion}

We introduce \benchname, a training-free testbed that measures whether dense signals order a state’s candidate actions like a strong reference policy’s returns do.
Across 21 methods spanning seven families from the literature, four environments and six backbones, simple prompting baselines are the most
$Q$-aligned on average, averaging estimates helps where elaborate pipelines do not, methods cluster reliably by family, and all methods are weak in open-ended settings.
In test-time action selection, $Q$-alignment highly correlates with downstream success. \benchname also provides additional metrics for what rank correlation does not capture: calibration, the cost of selection errors, and statistics to contextualize the evaluation setup.
\benchname is built to grow: new methods plug in by emitting a single score per state--action pair, and practitioners can apply our methodology to build datasets for their own tasks.
We hope \benchname supports faster, cheaper iteration on the dense signals needed to guide and supervise long-horizon agents.

\textbf{Limitations.} \benchname provides a principled and cheap approach to evaluating the alignment of dense supervision methods in isolation. Nonetheless, the quality of a signal is not the only driver of RL post-training effectiveness, and it is not isolated from the rest of the pipeline. When methods are compared directly through Q-alignment, simple direct value prediction is surprisingly competitive with more specialized mechanisms. The confounding elements we want to isolate the signal from when evaluating it (e.g., optimization and algorithmic choices, strategies for integrating the signal into the loss function, interactions with other signals, etc.) must also be studied to build successful post-training systems. Furthermore, some signals that poorly align with a particular environmental objective might still be beneficial to learning agents (e.g., exploration incentives). Therefore, future work should treat direct prompting as a baseline when evaluating signal quality.

\newpage

\section*{Acknowledgments}

The authors thank (in alphabetical order): Hardik Bhatnagar, Nikhil Chandak, Shyamgopal Karthik, Shashwat Goel, Matthias K\"ummerer, Ronald Skorobogat and Vishaal Udandarao for valuable feedback on the project. IA and JS acknowledge support by the Tübingen AI Center. JS and SHG thank the International Max Planck Research School for Intelligent Systems (IMPRS-IS) for support. SHG and AP acknowledge funding by the Federal Ministry of Research, Technology and Space (BMFTR), FKZ: 16IS24085B. AP and MB acknowledge Coefficient Giving funded by the Good Ventures Foundation. MB acknowledges funding by the Federal Ministry of Research, Technology and Space (BMFTR), FKZ: 16IS24079A. MB is a member of the Machine Learning Cluster of Excellence, funded by the Deutsche Forschungsgemeinschaft (DFG, German Research Foundation) under Germany’s Excellence Strategy – EXC number 2064/1 – Project number 390727645.

\bibliographystyle{plainnat}
\bibliography{main}

\clearpage
\appendix
\part{Appendix}

\localtableofcontents
\clearpage

\section{Additional results}
\label{app::results}

\subsection{Global Spearman correlation}
\label{app::global_spearman}

\begin{figure}[t]
\centering
\includegraphics[width=\linewidth]{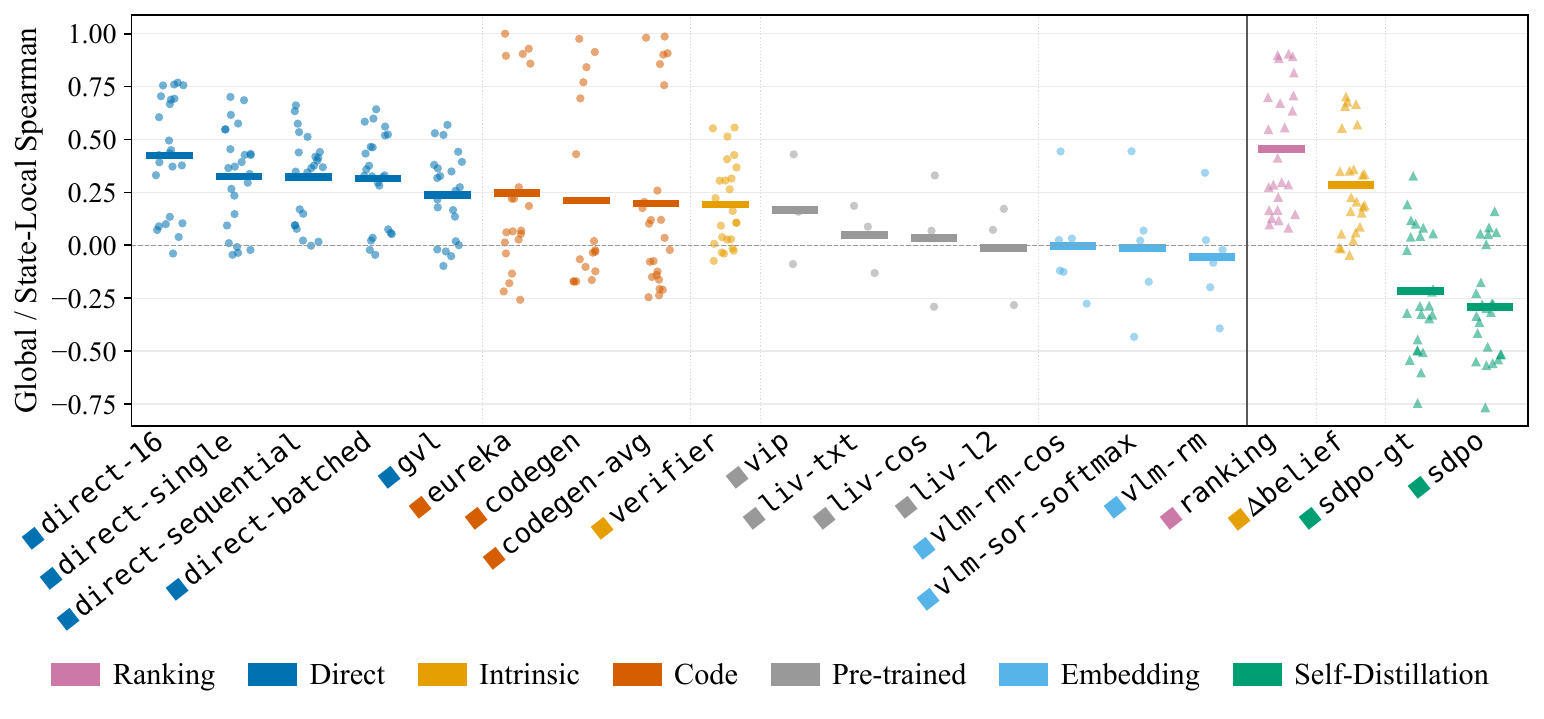}
\caption{\textbf{Global Spearman leaderboard.} The divider separates global pointwise correlations from state-local correlations.}
\label{fig:global_leaderboard}
\end{figure}

Figure~\ref{fig:global_leaderboard} reflects the \benchname leaderboard with global Spearman correlation: computed over all state--action pairs in the dataset pooled together, instead of grouped and averaged over states. Some methods can only use state-local Spearman, since they natively rank actions for a given fixed state; they are shown as a separate group on the right. Results for global Spearman are computed over the same predictions and references as for state-local Spearman; complete results are given in Tables~\ref{tab:global_spearman_terminal_bench_text}--\ref{tab:global_spearman_frozen_lake_vision}.

\subsection{Robustness analyses}
\label{sec::experiments::ablations}

\begin{figure}[t]
   \centering
   \includegraphics[width=0.95\linewidth]{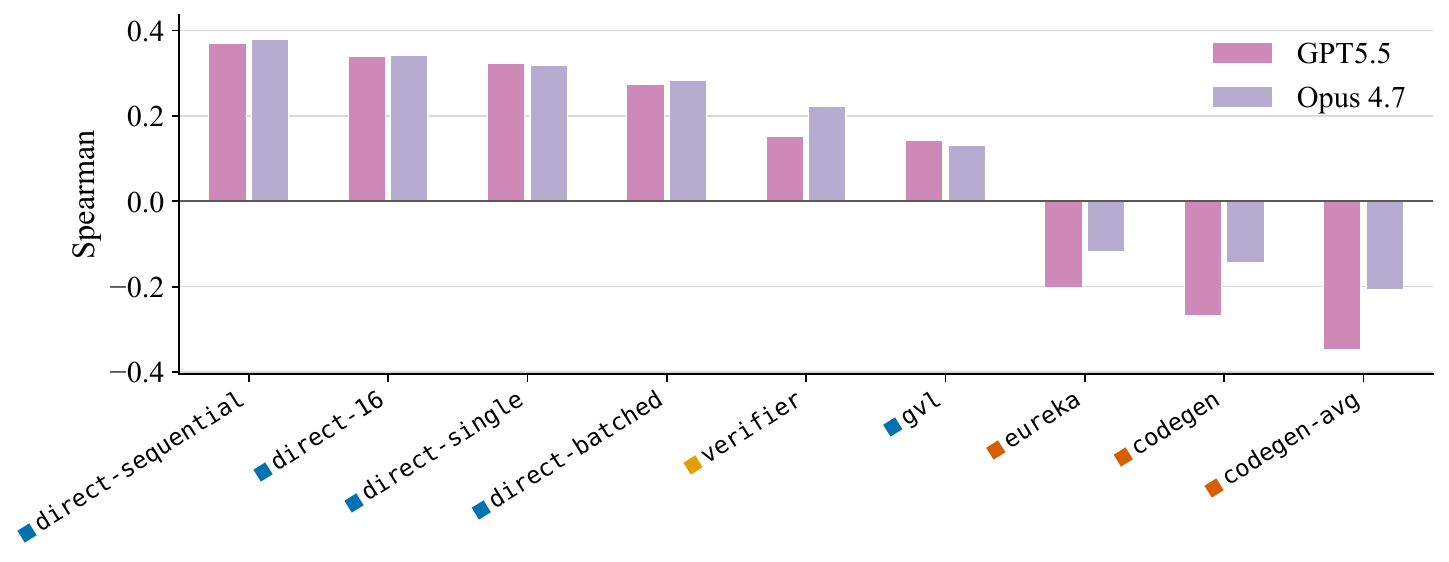}
    \caption{
    \textbf{Backbone ablation for TerminalBench value labels.}
    Global Spearman of the same numerical predictions against GPT-5.5 and Claude Opus 4.7 MVMC labels, computed over 100 actions and averaged over six backbones. Both reference policies preserve the ranking of methods induced by \benchname.
    }
   \label{fig:gt-ablation}
\end{figure}

The reference-backbone comparison evaluates whether two different models used as reference policies to label the TerminalBench dataset yield the same \benchname ranking (Figure~\ref{fig:gt-ablation}).

\begin{wrapfigure}{r}{0.50\textwidth}
\vspace{-\intextsep}
\centering
\includegraphics[width=\linewidth]{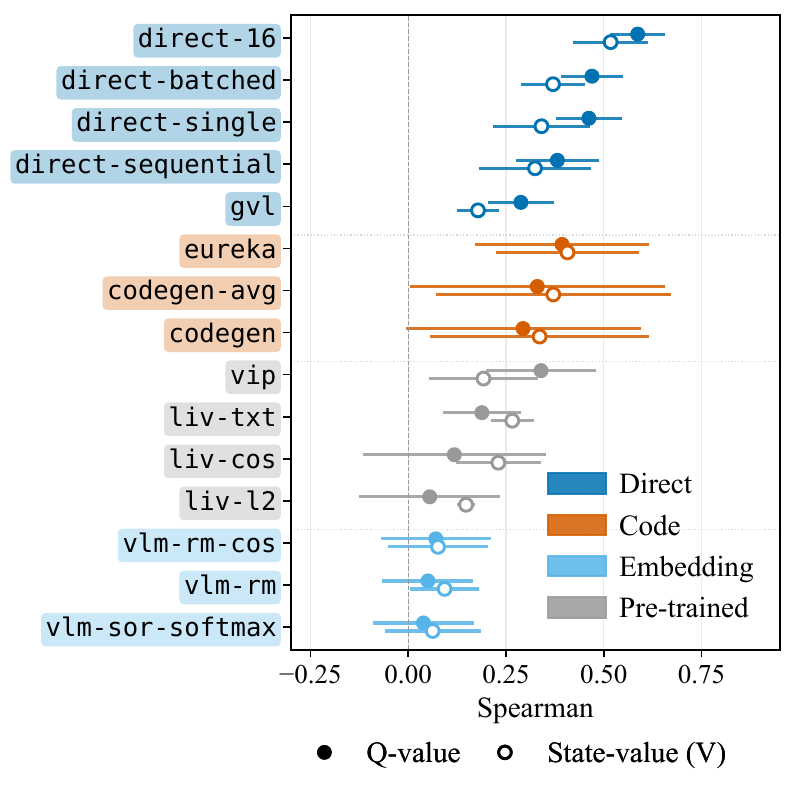}
\caption{\textbf{Q-value (filled) vs.\ state-value (shallow).} Signal type
        averaged over models and environments (OpenApps, ALFWorld, FrozenLake).  Points show means; bars show ±1 SE.}
\label{fig:ablations}
\vspace{-\intextsep}
\end{wrapfigure}
\phantomsection\label{app::value_targets}
\textbf{Reference value type.}
\benchname provides state-value labels $V(s_t)$ alongside $Q$-values for OpenApps, ALFWorld, and FrozenLake, letting us test whether method rankings depend on the choice of target value.
Figure~\ref{fig:ablations} compares the two across the direct, code, and pretrained
families, averaged over models and environments. 
The relative ordering of methods is largely
preserved, so our conclusions do not rely on a particular value target. 
Absolute correlations do change: code and pretrained methods align better with state values, and direct prompting with $Q$-values. 
This likely reflects differences in how methods consume the input: code-based methods may more naturally express state-level heuristics as executable functions, while direct prompting can explicitly reason about actions.

\phantomsection\label{app::reference_backbone}
\textbf{MVMC backbones.}
We compare the same numerical predictions against TerminalBench labels estimated using GPT-5.5~\citep{openai_gpt-55_2026} and Claude Opus 4.7~\citep{anthropic2026claudemodels} with Max-Value Monte Carlo.
Figure~\ref{fig:gt-ablation} reports global Spearman on 100 actions, for which both policies provide labels.
Averaged over the six backbones, all nine methods retain the sign of their correlation: direct methods remain positively aligned and code-based methods negatively aligned under both reference policies.
The magnitudes change (e.g., \texttt{codegen-avg} scores $-0.347$ with GPT-5.5 and $-0.206$ with Opus 4.7, whereas \texttt{direct-16} scores $0.340$ and $0.341$), but the ranking induced by \benchname is preserved.
These results support the same broad comparison between method families, while showing sensitivity to the reference policy.

\subsection{Mean versus maximum continuation return}
\label{app::mean_max}

The average and maximum return over the same stochastic continuations estimate different targets: the mean estimates the continuation policy's expected return, whereas the maximum is a best-continuation proxy. Table~\ref{tab:mean_max} compares Spearman for every method with respect to these two targets. Method rankings can change: \texttt{ranking} rises from $0.123$ to $0.162$, while \texttt{sdpo} falls from $0.070$ to $-0.032$. Neither target yields strong alignment on TerminalBench.

\begin{table}[!htbp]
\centering\small
\caption{State-local Spearman on TerminalBench for maximum and mean returns from the same 16 continuations, keeping method scores fixed.}
\label{tab:mean_max}
\begin{tabular}{lrrr}
\toprule
Method & $n$ & Maximum return & Mean return \\
\midrule
\texttt{direct-single} & 6 & 0.073 & 0.073 \\
\texttt{direct-16} & 6 & 0.146 & 0.122 \\
\texttt{direct-batched} & 6 & 0.068 & 0.084 \\
\texttt{direct-sequential} & 5 & -0.034 & 0.011 \\
\texttt{gvl} & 4 & 0.044 & 0.047 \\
\texttt{eureka} & 5 & -0.080 & -0.036 \\
\texttt{codegen-avg} & 6 & -0.036 & -0.006 \\
\texttt{verifier} & 6 & 0.066 & 0.058 \\
\texttt{codegen} & 6 & -0.075 & -0.053 \\
\texttt{$\Delta$belief} & 6 & 0.055 & 0.063 \\
\texttt{ranking} & 6 & 0.123 & 0.162 \\
\texttt{sdpo} & 6 & 0.070 & -0.032 \\
\texttt{sdpo-gt} & 6 & 0.074 & -0.005 \\
\bottomrule
\end{tabular}
\end{table}

\subsection{Calibration of numerical scores}
\label{app::calibration_results}

Table~\ref{tab:calibration} reports MSE on the candidate actions, using the definitions in Appendix~\ref{app::eval_metrics}. We include all available numerical outputs, including constant scores, for which MSE remains defined even when rank correlation is not. Raw MSE is reported only for the direct and code methods, whose outputs are intended to estimate returns; the other numerical scorers are reported after the affine fit only. Permutation-based and likelihood-ratio signals are excluded from this calibration comparison.

\begin{table}[t]
\centering
\caption{Calibration against reference values. Each entry is the mean over six backbones for LLM-based methods, two encoders for embedding methods, and one checkpoint for pretrained methods. Each affine fit uses the same candidate labels on which its error is reported; for \texttt{codegen}, fitting and MSE are computed separately for each function before averaging. Dashes mark inapplicable raw errors or unevaluated environments.}
\label{tab:calibration}
\scriptsize
\setlength{\tabcolsep}{3pt}
\begin{tabular}{lrrrrrrrr}
\toprule
Method & \multicolumn{2}{c}{FrozenLake} & \multicolumn{2}{c}{ALFWorld} & \multicolumn{2}{c}{OpenApps} & \multicolumn{2}{c}{TerminalBench} \\
 & Raw & Affine & Raw & Affine & Raw & Affine & Raw & Affine \\
\midrule
\texttt{gvl} & 0.104 & 0.0291 & 0.11 & 0.00308 & 0.252 & 0.122 & 0.16 & 0.0494 \\
\texttt{direct-single} & 0.12 & 0.0323 & 0.0877 & 0.00257 & 0.224 & 0.121 & 0.12 & 0.0493 \\
\texttt{direct-batched} & 0.0946 & 0.0337 & 0.125 & 0.00244 & 0.234 & 0.121 & 0.172 & 0.0495 \\
\texttt{direct-sequential} & 0.11 & 0.0354 & 0.0998 & 0.00253 & 0.214 & 0.122 & 0.118 & 0.0492 \\
\texttt{direct-16} & 0.0837 & 0.0192 & 0.0654 & 0.00201 & 0.202 & 0.121 & 0.1 & 0.0493 \\
\texttt{eureka} & 0.0589 & 0.0174 & 0.605 & 0.00319 & 0.208 & 0.121 & 0.288 & 0.049 \\
\texttt{codegen} & 0.332 & 0.0266 & 0.337 & 0.00331 & 0.236 & 0.121 & 0.231 & 0.0491 \\
\texttt{codegen-avg} & 0.0773 & 0.014 & 0.28 & 0.00328 & 0.184 & 0.121 & 0.175 & 0.0491 \\
\texttt{verifier} & --- & 0.0422 & --- & 0.00248 & --- & 0.122 & --- & 0.0491 \\
\texttt{vlm-rm} & --- & 0.0413 & --- & 0.00303 & --- & 0.122 & --- & --- \\
\texttt{vlm-rm-cos} & --- & 0.0358 & --- & 0.00291 & --- & 0.121 & --- & --- \\
\texttt{vlm-sor} & --- & 0.0575 & --- & 0.00333 & --- & 0.122 & --- & --- \\
\texttt{vlm-sor-softmax} & --- & 0.0444 & --- & 0.00288 & --- & 0.122 & --- & --- \\
\texttt{vip} & --- & 0.0575 & --- & 0.0027 & --- & 0.122 & --- & --- \\
\texttt{liv-cos} & --- & 0.0473 & --- & 0.00285 & --- & 0.122 & --- & --- \\
\texttt{liv-l2} & --- & 0.0407 & --- & 0.00307 & --- & 0.122 & --- & --- \\
\texttt{liv-txt} & --- & 0.0574 & --- & 0.00318 & --- & 0.114 & --- & --- \\
\bottomrule
\end{tabular}
\end{table}

\subsection{Alignment over an episode}
\label{app::temporal_alignment}

We group each sampled state by its position in the trajectory. For zero-based step $t$ in a trajectory of $L$ transitions, its normalized position is $t/(L-1)$; single-transition trajectories are assigned $0.5$. The three groups are early $[0,1/3)$, middle $[1/3,2/3)$, and late $[2/3,1]$. Table~\ref{tab:temporal_counts} gives their sizes for each environment. Within each group, we average the same Spearman correlations used in the main results. We retain the main leaderboard's reporting rule and require at least ten valid states per cell and group.

\begin{table}[t]
\centering
\caption{Sampled states by episode position, for each environment.}
\label{tab:temporal_counts}
\small
\begin{tabular}{lrrr}
\toprule
Environment & Early & Middle & Late \\
\midrule
FrozenLake & 33 & 38 & 29 \\
ALFWorld & 34 & 43 & 23 \\
OpenApps & 28 & 36 & 30 \\
TerminalBench & 14 & 64 & 22 \\
\bottomrule
\end{tabular}
\end{table}

\begin{figure}[p]
\centering
\includegraphics[width=\linewidth]{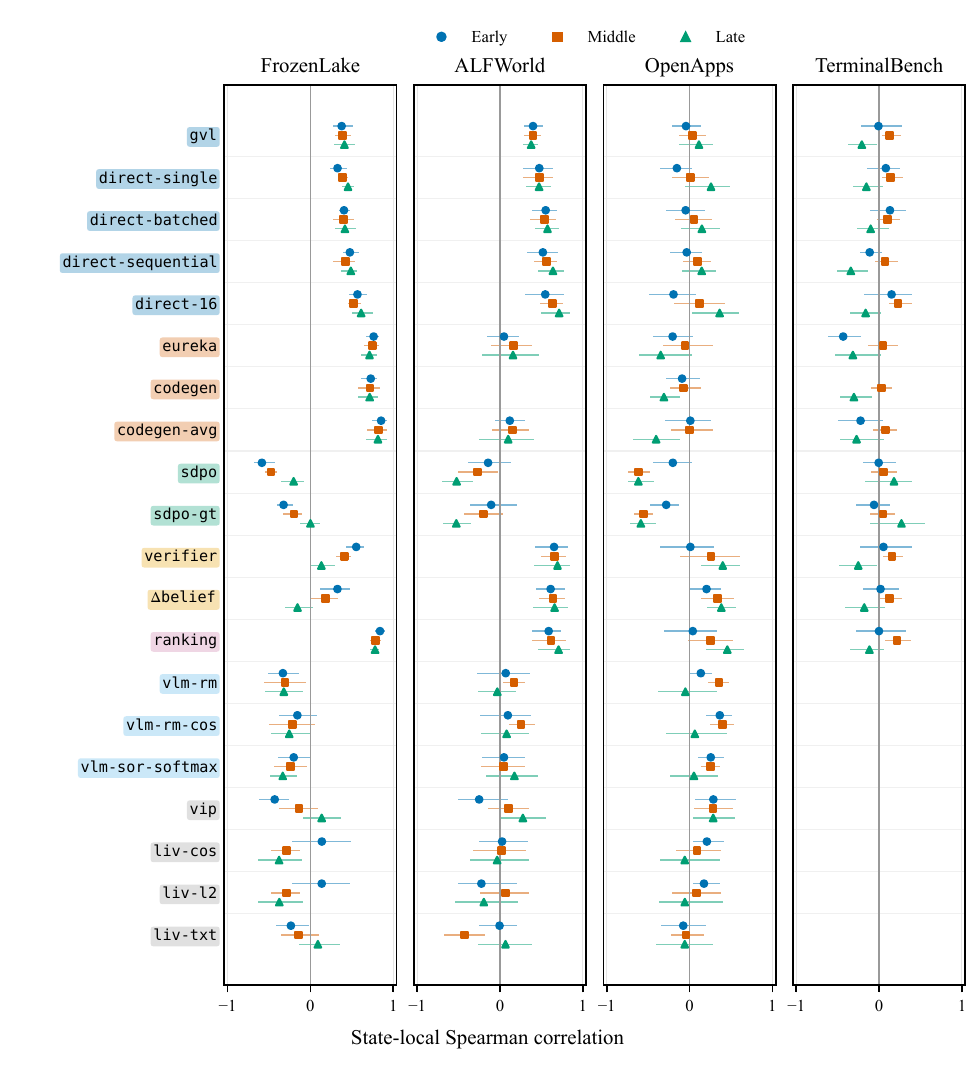}
\caption{\textbf{$Q$-alignment over an episode.} Each method is shown separately for early, middle and late collection states; points average backbones, and bars show 95\% bootstrap confidence intervals from 2,000 draws. The saved code functions and reference labels are held fixed. A cell--phase with fewer than ten valid states is omitted, so the contributing backbones can differ between phases. Methods with no eligible cell are absent; blank environment entries indicate no evaluation.}
\label{fig:temporal_alignment}
\end{figure}

Figure~\ref{fig:temporal_alignment} shows that alignment does not consistently improve toward episode completion. In FrozenLake, \texttt{codegen-avg} remains strongly aligned from early to late states ($0.85$ to $0.81$), whereas \texttt{verifier} declines ($0.55$ to $0.13$). In ALFWorld, \texttt{direct-single} remains close to $0.47$ in all three groups, while \texttt{ranking} rises from $0.59$ to $0.71$. In OpenApps, \texttt{ranking} rises from $0.04$ to $0.46$. These are descriptive comparisons of different sampled states and task mixtures, rather than a controlled effect of advancing time.

\subsection{Complete benchmark tables}

Cells report Spearman ($\rho$) correlation between predicted
and ground-truth signals. Primary $Q$-value tables use the common state-local
metric for every method: Spearman is computed over candidate actions within
each valid state, then reported as $\mathrm{mean} \pm \mathrm{std}$ across states.
Other tables are explicitly global-metric
ablations. Significance markers
for global cells are $^{*}\,p<.10$, $^{**}\,p<.01$, $^{***}\,p<.001$ (no marker
means not significant). Em-dash (---) marks unavailable or undefined scores,
including scores excluded due to not having enough valid states; primary $Q$-value scores
require at least 50 valid states and at least 50\% of compared states. \texttt{Codegen}
aggregate cells show $\mathrm{mean} \pm \mathrm{std}$ on the top line and
$[\min, \max]$ on the bottom over 16 sample correlations.
Rows and columns that are conceptually inapplicable to a given slice
(e.g.\ \texttt{$\Delta$belief} or \texttt{ranking} in $V$-value tables; \textsc{CLIP} /
\textsc{SigLIP} columns in $V$-value vision tables) are omitted from the table schema.
Method rows, metric rows, and model columns with no values are omitted.

\begin{sidewaystable}[p]
\centering
\small
\setlength{\tabcolsep}{3pt}
\caption{Correlations on TerminalBench, Q-value, text modality, using GPT-5.5 Max-Value Monte Carlo for label generation.}
\label{tab:corr_terminalbench_qv_text_codex_55}
\begin{tabular}{l l c c c c c c}
\toprule
Method & Metric & \rotatebox{45}{Gemma4 26B-A4B} & \rotatebox{45}{Gemma4 31B} & \rotatebox{45}{Qwen3.5 9B} & \rotatebox{45}{Qwen3.5 27B} & \rotatebox{45}{Qwen3.5 35B-A3B} & \rotatebox{45}{Qwen3.5 122B-A10B} \\
\midrule
\texttt{gvl} & $\rho$ & --- & --- & $0.058 \pm 0.569$ & $0.084 \pm 0.541$ & $0.080 \pm 0.533$ & $-0.044 \pm 0.607$ \\
\texttt{direct-single} & $\rho$ & $0.067 \pm 0.615$ & $0.091 \pm 0.629$ & $-0.007 \pm 0.542$ & $0.107 \pm 0.534$ & $-0.007 \pm 0.594$ & $0.188 \pm 0.556$ \\
\texttt{direct-batched} & $\rho$ & $0.067 \pm 0.591$ & $0.141 \pm 0.603$ & $-0.041 \pm 0.627$ & $0.112 \pm 0.541$ & $0.057 \pm 0.581$ & $0.075 \pm 0.596$ \\
\texttt{direct-sequential} & $\rho$ & --- & $-0.044 \pm 0.621$ & $-0.087 \pm 0.621$ & $-0.040 \pm 0.597$ & $0.050 \pm 0.615$ & $-0.049 \pm 0.629$ \\
\texttt{direct-16} & $\rho$ & $0.210 \pm 0.570$ & $0.202 \pm 0.573$ & $0.022 \pm 0.620$ & $0.137 \pm 0.609$ & $0.085 \pm 0.557$ & $0.222 \pm 0.596$ \\
\midrule
\texttt{eureka} & $\rho$ & $-0.014 \pm 0.651$ & $-0.138 \pm 0.613$ & --- & $-0.079 \pm 0.623$ & $-0.109 \pm 0.649$ & $-0.060 \pm 0.619$ \\
\texttt{codegen} & $\rho$ & \makecell{$-0.070 \pm 0.080$ \\ $[-0.218, 0.081]$} & \makecell{$-0.012 \pm 0.076$ \\ $[-0.122, 0.155]$} & \makecell{$-0.099 \pm 0.094$ \\ $[-0.245, 0.072]$} & \makecell{$-0.094 \pm 0.053$ \\ $[-0.202, 0.002]$} & \makecell{$-0.097 \pm 0.075$ \\ $[-0.265, 0.064]$} & \makecell{$-0.075 \pm 0.055$ \\ $[-0.210, -0.010]$} \\
\texttt{codegen-avg} & $\rho$ & $-0.081 \pm 0.652$ & $0.012 \pm 0.611$ & $-0.078 \pm 0.622$ & $0.008 \pm 0.605$ & $-0.073 \pm 0.607$ & $-0.002 \pm 0.605$ \\
\midrule
\texttt{sdpo} & $\rho$ & $0.085 \pm 0.611$ & $0.062 \pm 0.586$ & $0.052 \pm 0.602$ & $0.005 \pm 0.610$ & $0.161 \pm 0.565$ & $0.055 \pm 0.607$ \\
\texttt{sdpo-gt} & $\rho$ & $0.040 \pm 0.608$ & $0.056 \pm 0.594$ & $0.044 \pm 0.566$ & $0.102 \pm 0.606$ & $0.118 \pm 0.627$ & $0.083 \pm 0.650$ \\
\midrule
\texttt{verifier} & $\rho$ & $0.075 \pm 0.649$ & $0.080 \pm 0.617$ & $-0.030 \pm 0.668$ & $0.108 \pm 0.621$ & $0.050 \pm 0.605$ & $0.114 \pm 0.591$ \\
\texttt{$\Delta$belief} & $\rho$ & $-0.015 \pm 0.638$ & $0.024 \pm 0.583$ & $-0.011 \pm 0.602$ & $0.060 \pm 0.591$ & $0.088 \pm 0.612$ & $0.182 \pm 0.604$ \\
\midrule
\texttt{ranking} & $\rho$ & $0.118 \pm 0.627$ & $0.082 \pm 0.616$ & $0.147 \pm 0.629$ & $0.098 \pm 0.626$ & $0.128 \pm 0.592$ & $0.167 \pm 0.592$ \\
\bottomrule
\end{tabular}
\end{sidewaystable}

\begin{sidewaystable}[p]
\centering
\small
\setlength{\tabcolsep}{3pt}
\caption{Correlations on OpenApps, Q-value, text modality, using a scripted policy with Max-Value Monte Carlo for label generation.}
\label{tab:corr_openapps_qv_text_scripted}
\begin{tabular}{l l c c c c c c}
\toprule
Method & Metric & \rotatebox{45}{Gemma4 26B-A4B} & \rotatebox{45}{Gemma4 31B} & \rotatebox{45}{Qwen3.5 9B} & \rotatebox{45}{Qwen3.5 27B} & \rotatebox{45}{Qwen3.5 35B-A3B} & \rotatebox{45}{Qwen3.5 122B-A10B} \\
\midrule
\texttt{gvl} & $\rho$ & $0.012 \pm 0.599$ & $0.113 \pm 0.644$ & $0.049 \pm 0.644$ & $0.093 \pm 0.620$ & $-0.064 \pm 0.568$ & $0.031 \pm 0.606$ \\
\texttt{direct-single} & $\rho$ & $0.014 \pm 0.665$ & $-0.062 \pm 0.647$ & $0.112 \pm 0.577$ & $0.076 \pm 0.616$ & $-0.019 \pm 0.602$ & $0.125 \pm 0.603$ \\
\texttt{direct-batched} & $\rho$ & $0.041 \pm 0.656$ & $-0.028 \pm 0.701$ & $0.115 \pm 0.636$ & $0.051 \pm 0.658$ & $0.043 \pm 0.671$ & $0.115 \pm 0.648$ \\
\texttt{direct-sequential} & $\rho$ & $-0.036 \pm 0.687$ & $-0.092 \pm 0.689$ & $0.145 \pm 0.624$ & $0.077 \pm 0.661$ & $0.132 \pm 0.583$ & $0.207 \pm 0.632$ \\
\texttt{direct-16} & $\rho$ & $0.077 \pm 0.662$ & $-0.013 \pm 0.656$ & $0.197 \pm 0.585$ & $0.110 \pm 0.683$ & $0.089 \pm 0.639$ & $0.188 \pm 0.628$ \\
\midrule
\texttt{eureka} & $\rho$ & $-0.221 \pm 0.681$ & $-0.182 \pm 0.635$ & --- & $-0.238 \pm 0.681$ & $-0.271 \pm 0.701$ & $-0.083 \pm 0.606$ \\
\texttt{codegen} & $\rho$ & \makecell{$-0.182 \pm 0.241$ \\ $[-0.441, 0.621]$} & \makecell{$-0.155 \pm 0.114$ \\ $[-0.455, 0.084]$} & --- & \makecell{$-0.124 \pm 0.169$ \\ $[-0.351, 0.216]$} & \makecell{$-0.193 \pm 0.242$ \\ $[-0.456, 0.274]$} & \makecell{$-0.139 \pm 0.180$ \\ $[-0.621, 0.158]$} \\
\texttt{codegen-avg} & $\rho$ & $-0.237 \pm 0.635$ & $-0.010 \pm 0.703$ & $-0.169 \pm 0.676$ & $-0.112 \pm 0.646$ & $-0.024 \pm 0.643$ & $-0.222 \pm 0.639$ \\
\midrule
\texttt{sdpo} & $\rho$ & $-0.549 \pm 0.487$ & $-0.557 \pm 0.454$ & $-0.316 \pm 0.644$ & $-0.516 \pm 0.500$ & $-0.480 \pm 0.509$ & $-0.540 \pm 0.497$ \\
\texttt{sdpo-gt} & $\rho$ & $-0.446 \pm 0.519$ & $-0.542 \pm 0.446$ & $-0.320 \pm 0.569$ & $-0.495 \pm 0.487$ & $-0.506 \pm 0.499$ & $-0.602 \pm 0.459$ \\
\midrule
\texttt{verifier} & $\rho$ & $0.240 \pm 0.641$ & $0.210 \pm 0.640$ & $0.188 \pm 0.603$ & $0.288 \pm 0.586$ & $0.242 \pm 0.638$ & $0.223 \pm 0.672$ \\
\texttt{$\Delta$belief} & $\rho$ & $0.352 \pm 0.551$ & $0.350 \pm 0.544$ & $0.333 \pm 0.555$ & $0.153 \pm 0.608$ & $0.337 \pm 0.553$ & $0.360 \pm 0.529$ \\
\midrule
\texttt{ranking} & $\rho$ & $0.298 \pm 0.618$ & $0.166 \pm 0.616$ & $0.275 \pm 0.606$ & $0.286 \pm 0.630$ & $0.228 \pm 0.634$ & $0.287 \pm 0.611$ \\
\bottomrule
\end{tabular}
\end{sidewaystable}

\begin{table}[htbp]
\centering
\small
\setlength{\tabcolsep}{3pt}
\caption{Correlations on OpenApps, Q-value, vision modality, using a scripted policy with Max-Value Monte Carlo for label generation.}
\label{tab:corr_openapps_qv_vision_scripted}
\begin{tabular}{l l c c c}
\toprule
 &  & \multicolumn{3}{c}{Non-LLM} \\
\cmidrule(lr){3-5}
Method & Metric & \rotatebox{45}{Self} & \rotatebox{45}{CLIP} & \rotatebox{45}{SigLIP} \\
\midrule
\texttt{vlm-rm} & $\rho$ & --- & $0.140 \pm 0.678$ & $0.173 \pm 0.621$ \\
\texttt{vlm-rm-cos} & $\rho$ & --- & $0.266 \pm 0.669$ & $0.291 \pm 0.601$ \\
\texttt{vlm-sor-softmax} & $\rho$ & --- & $0.146 \pm 0.655$ & $0.233 \pm 0.628$ \\
\midrule
\texttt{vip} & $\rho$ & $0.285 \pm 0.583$ & --- & --- \\
\texttt{liv-cos} & $\rho$ & $0.076 \pm 0.605$ & --- & --- \\
\texttt{liv-l2} & $\rho$ & $0.064 \pm 0.614$ & --- & --- \\
\texttt{liv-txt} & $\rho$ & $-0.057 \pm 0.615$ & --- & --- \\
\bottomrule
\end{tabular}
\end{table}

\begin{sidewaystable}[p]
\centering
\small
\setlength{\tabcolsep}{3pt}
\caption{Correlations on ALFWorld, Q-value, text modality, using symbolic planner Max-Value Monte Carlo for label generation.}
\label{tab:corr_alfworld_qv_text_planner}
\begin{tabular}{l l c c c c c c}
\toprule
Method & Metric & \rotatebox{45}{Gemma4 26B-A4B} & \rotatebox{45}{Gemma4 31B} & \rotatebox{45}{Qwen3.5 9B} & \rotatebox{45}{Qwen3.5 27B} & \rotatebox{45}{Qwen3.5 35B-A3B} & \rotatebox{45}{Qwen3.5 122B-A10B} \\
\midrule
\texttt{gvl} & $\rho$ & $0.316 \pm 0.608$ & $0.478 \pm 0.516$ & $0.454 \pm 0.499$ & $0.533 \pm 0.424$ & $0.211 \pm 0.552$ & $0.352 \pm 0.553$ \\
\texttt{direct-single} & $\rho$ & $0.453 \pm 0.489$ & $0.577 \pm 0.404$ & $0.448 \pm 0.510$ & $0.526 \pm 0.476$ & $0.428 \pm 0.526$ & $0.421 \pm 0.523$ \\
\texttt{direct-batched} & $\rho$ & $0.506 \pm 0.544$ & $0.607 \pm 0.456$ & $0.510 \pm 0.449$ & $0.558 \pm 0.477$ & $0.560 \pm 0.501$ & $0.572 \pm 0.456$ \\
\texttt{direct-sequential} & $\rho$ & $0.637 \pm 0.426$ & $0.546 \pm 0.522$ & $0.507 \pm 0.400$ & $0.608 \pm 0.462$ & $0.568 \pm 0.414$ & $0.579 \pm 0.474$ \\
\texttt{direct-16} & $\rho$ & $0.592 \pm 0.494$ & $0.671 \pm 0.419$ & $0.602 \pm 0.457$ & $0.757 \pm 0.304$ & $0.540 \pm 0.491$ & $0.647 \pm 0.482$ \\
\midrule
\texttt{eureka} & $\rho$ & --- & --- & $0.090 \pm 0.572$ & $0.181 \pm 0.628$ & $0.112 \pm 0.660$ & $0.150 \pm 0.659$ \\
\texttt{codegen-avg} & $\rho$ & $0.244 \pm 0.593$ & $0.052 \pm 0.719$ & $0.095 \pm 0.684$ & $0.133 \pm 0.609$ & $0.050 \pm 0.667$ & $0.176 \pm 0.628$ \\
\midrule
\texttt{sdpo} & $\rho$ & $-0.226 \pm 0.646$ & $-0.334 \pm 0.569$ & $-0.276 \pm 0.636$ & $-0.279 \pm 0.556$ & $-0.415 \pm 0.491$ & $-0.363 \pm 0.532$ \\
\texttt{sdpo-gt} & $\rho$ & $-0.220 \pm 0.599$ & $-0.328 \pm 0.575$ & $-0.207 \pm 0.586$ & $-0.285 \pm 0.552$ & $-0.287 \pm 0.634$ & $-0.325 \pm 0.561$ \\
\midrule
\texttt{verifier} & $\rho$ & $0.688 \pm 0.441$ & $0.729 \pm 0.347$ & $0.587 \pm 0.463$ & $0.728 \pm 0.379$ & $0.643 \pm 0.444$ & $0.636 \pm 0.416$ \\
\texttt{$\Delta$belief} & $\rho$ & $0.667 \pm 0.396$ & $0.679 \pm 0.343$ & $0.555 \pm 0.529$ & $0.659 \pm 0.384$ & $0.571 \pm 0.486$ & $0.703 \pm 0.382$ \\
\midrule
\texttt{ranking} & $\rho$ & $0.557 \pm 0.454$ & $0.700 \pm 0.398$ & $0.548 \pm 0.531$ & $0.672 \pm 0.421$ & $0.636 \pm 0.461$ & $0.707 \pm 0.407$ \\
\bottomrule
\end{tabular}
\end{sidewaystable}

\begin{table}[htbp]
\centering
\small
\setlength{\tabcolsep}{3pt}
\caption{Correlations on ALFWorld, Q-value, vision modality, using symbolic planner Max-Value Monte Carlo for label generation.}
\label{tab:corr_alfworld_qv_vision_planner}
\begin{tabular}{l l c c c}
\toprule
 &  & \multicolumn{3}{c}{Non-LLM} \\
\cmidrule(lr){3-5}
Method & Metric & \rotatebox{45}{Self} & \rotatebox{45}{CLIP} & \rotatebox{45}{SigLIP} \\
\midrule
\texttt{vlm-rm} & $\rho$ & --- & $0.257 \pm 0.640$ & $-0.088 \pm 0.688$ \\
\texttt{vlm-rm-cos} & $\rho$ & --- & $0.346 \pm 0.614$ & $-0.018 \pm 0.700$ \\
\texttt{vlm-sor-softmax} & $\rho$ & --- & $0.149 \pm 0.691$ & $0.018 \pm 0.683$ \\
\midrule
\texttt{vip} & $\rho$ & $0.073 \pm 0.634$ & --- & --- \\
\texttt{liv-cos} & $\rho$ & $0.004 \pm 0.694$ & --- & --- \\
\texttt{liv-l2} & $\rho$ & $-0.082 \pm 0.672$ & --- & --- \\
\texttt{liv-txt} & $\rho$ & $-0.178 \pm 0.599$ & --- & --- \\
\bottomrule
\end{tabular}
\end{table}

\begin{sidewaystable}[p]
\centering
\small
\setlength{\tabcolsep}{3pt}
\caption{Correlations on FrozenLake, Q-value, text modality, using a scripted policy with Max-Value Monte Carlo for label generation.}
\label{tab:corr_frozenlake_qv_text_scripted}
\begin{tabular}{l l c c c c c c}
\toprule
Method & Metric & \rotatebox{45}{Gemma4 26B-A4B} & \rotatebox{45}{Gemma4 31B} & \rotatebox{45}{Qwen3.5 9B} & \rotatebox{45}{Qwen3.5 27B} & \rotatebox{45}{Qwen3.5 35B-A3B} & \rotatebox{45}{Qwen3.5 122B-A10B} \\
\midrule
\texttt{gvl} & $\rho$ & $0.003 \pm 0.568$ & $0.624 \pm 0.412$ & $0.373 \pm 0.571$ & $0.448 \pm 0.570$ & $0.339 \pm 0.531$ & $0.556 \pm 0.493$ \\
\texttt{direct-single} & $\rho$ & $0.300 \pm 0.511$ & $0.590 \pm 0.469$ & $0.277 \pm 0.564$ & $0.416 \pm 0.503$ & $0.213 \pm 0.545$ & $0.533 \pm 0.442$ \\
\texttt{direct-batched} & $\rho$ & $0.234 \pm 0.617$ & $0.526 \pm 0.547$ & $0.442 \pm 0.546$ & $0.440 \pm 0.548$ & $0.304 \pm 0.619$ & $0.494 \pm 0.578$ \\
\texttt{direct-sequential} & $\rho$ & $0.314 \pm 0.592$ & $0.776 \pm 0.448$ & $0.393 \pm 0.483$ & $0.460 \pm 0.540$ & $0.205 \pm 0.680$ & $0.606 \pm 0.448$ \\
\texttt{direct-16} & $\rho$ & $0.536 \pm 0.367$ & $0.776 \pm 0.227$ & $0.549 \pm 0.442$ & $0.706 \pm 0.345$ & $0.360 \pm 0.539$ & $0.452 \pm 0.522$ \\
\midrule
\texttt{eureka} & $\rho$ & $0.608 \pm 0.303$ & $1.000 \pm 0.000$ & $0.400 \pm 0.425$ & $0.897 \pm 0.161$ & $0.844 \pm 0.216$ & $0.703 \pm 0.260$ \\
\texttt{codegen} & $\rho$ & \makecell{$0.816 \pm 0.109$ \\ $[0.562, 0.930]$} & \makecell{$0.860 \pm 0.069$ \\ $[0.648, 0.930]$} & \makecell{$0.671 \pm 0.167$ \\ $[0.317, 0.868]$} & \makecell{$0.662 \pm 0.147$ \\ $[0.247, 0.864]$} & \makecell{$0.635 \pm 0.213$ \\ $[0.251, 0.851]$} & \makecell{$0.682 \pm 0.139$ \\ $[0.396, 0.894]$} \\
\texttt{codegen-avg} & $\rho$ & $0.892 \pm 0.172$ & $0.845 \pm 0.246$ & $0.845 \pm 0.246$ & $0.747 \pm 0.281$ & $0.804 \pm 0.261$ & $0.845 \pm 0.246$ \\
\midrule
\texttt{sdpo} & $\rho$ & $-0.567 \pm 0.448$ & $-0.766 \pm 0.253$ & $-0.275 \pm 0.464$ & $-0.515 \pm 0.454$ & $-0.298 \pm 0.606$ & $-0.175 \pm 0.556$ \\
\texttt{sdpo-gt} & $\rho$ & $-0.499 \pm 0.440$ & $-0.745 \pm 0.257$ & $0.193 \pm 0.473$ & $-0.346 \pm 0.546$ & $0.328 \pm 0.502$ & $-0.023 \pm 0.507$ \\
\midrule
\texttt{verifier} & $\rho$ & $0.687 \pm 0.268$ & $0.653 \pm 0.264$ & $0.422 \pm 0.501$ & $0.437 \pm 0.460$ & $-0.100 \pm 0.563$ & $0.156 \pm 0.558$ \\
\texttt{$\Delta$belief} & $\rho$ & $0.206 \pm 0.602$ & $0.227 \pm 0.507$ & $-0.045 \pm 0.586$ & $0.190 \pm 0.601$ & $0.054 \pm 0.575$ & $0.161 \pm 0.536$ \\
\midrule
\texttt{ranking} & $\rho$ & $0.414 \pm 0.549$ & $0.893 \pm 0.141$ & $0.816 \pm 0.314$ & $0.885 \pm 0.146$ & $0.905 \pm 0.124$ & $0.899 \pm 0.126$ \\
\bottomrule
\end{tabular}
\end{sidewaystable}

\begin{table}[htbp]
\centering
\small
\setlength{\tabcolsep}{3pt}
\caption{Correlations on FrozenLake, Q-value, vision modality, using a scripted policy with Max-Value Monte Carlo for label generation.}
\label{tab:corr_frozenlake_qv_vision_scripted}
\begin{tabular}{l l c c c}
\toprule
 &  & \multicolumn{3}{c}{Non-LLM} \\
\cmidrule(lr){3-5}
Method & Metric & \rotatebox{45}{Self} & \rotatebox{45}{CLIP} & \rotatebox{45}{SigLIP} \\
\midrule
\texttt{vlm-rm} & $\rho$ & --- & $-0.491 \pm 0.405$ & $-0.149 \pm 0.626$ \\
\texttt{vlm-rm-cos} & $\rho$ & --- & $-0.212 \pm 0.544$ & $-0.207 \pm 0.623$ \\
\texttt{vlm-sor-softmax} & $\rho$ & --- & $-0.508 \pm 0.355$ & $-0.005 \pm 0.594$ \\
\midrule
\texttt{vip} & $\rho$ & $-0.155 \pm 0.574$ & --- & --- \\
\texttt{liv-cos} & $\rho$ & $-0.170 \pm 0.607$ & --- & --- \\
\texttt{liv-l2} & $\rho$ & $-0.171 \pm 0.609$ & --- & --- \\
\texttt{liv-txt} & $\rho$ & $-0.105 \pm 0.578$ & --- & --- \\
\bottomrule
\end{tabular}
\end{table}

\clearpage
\begin{table}[p]
\subsection{Additional ablation tables}

The reference-backbone, modality and state-value comparisons below use the protocol stated in each table: global correlations over all evaluation points or state-local correlations over candidate actions.

\medskip
\centering
\rotatebox{90}{%
\begin{minipage}{0.88\textheight}
\centering
\small
\setlength{\tabcolsep}{3pt}
\caption{Global Spearman on TerminalBench against Claude Opus 4.7 MVMC labels, using the benchmark's numerical scores for 100 actions. \texttt{Codegen} reports mean $\pm$ standard deviation and range across its individual functions.}
\label{tab:corr_terminalbench_qv_text_opus_47}
\begin{tabular}{l l c c c c c c}
\toprule
Method & Metric & \rotatebox{45}{Gemma4 26B-A4B} & \rotatebox{45}{Gemma4 31B} & \rotatebox{45}{Qwen3.5 9B} & \rotatebox{45}{Qwen3.5 27B} & \rotatebox{45}{Qwen3.5 35B-A3B} & \rotatebox{45}{Qwen3.5 122B-A10B} \\
\midrule
\texttt{gvl} & $\rho$ & $0.093$ & $0.114$ & $0.043$ & $0.129$ & $0.264^{**}$ & $0.142$ \\
\texttt{direct-single} & $\rho$ & $0.258^{**}$ & $0.463^{***}$ & $0.125$ & $0.341^{***}$ & $0.322^{**}$ & $0.400^{***}$ \\
\texttt{direct-batched} & $\rho$ & $0.268^{**}$ & $0.309^{**}$ & $0.255^{*}$ & $0.309^{**}$ & $0.325^{***}$ & $0.229^{*}$ \\
\texttt{direct-sequential} & $\rho$ & $0.311^{**}$ & $0.499^{***}$ & $0.402^{***}$ & $0.391^{***}$ & $0.377^{***}$ & $0.292^{**}$ \\
\texttt{direct-16} & $\rho$ & $0.349^{***}$ & $0.428^{***}$ & $0.285^{**}$ & $0.304^{**}$ & $0.310^{**}$ & $0.371^{***}$ \\
\midrule
\texttt{eureka} & $\rho$ & $-0.030$ & $-0.270^{**}$ & $-0.056$ & $-0.178^{*}$ & $-0.170^{*}$ & $-0.001$ \\
\texttt{codegen} & $\rho$ & \makecell{$-0.147 \pm 0.073$ \\ $[-0.243, 0.010]$} & \makecell{$-0.146 \pm 0.101$ \\ $[-0.382, 0.047]$} & \makecell{$-0.154 \pm 0.097$ \\ $[-0.318, 0.070]$} & \makecell{$-0.155 \pm 0.063$ \\ $[-0.247, -0.023]$} & \makecell{$-0.109 \pm 0.070$ \\ $[-0.194, 0.069]$} & \makecell{$-0.154 \pm 0.063$ \\ $[-0.238, -0.047]$} \\
\texttt{codegen-avg} & $\rho$ & $-0.244^{*}$ & $-0.202^{*}$ & $-0.209^{*}$ & $-0.207^{*}$ & $-0.165$ & $-0.205^{*}$ \\
\midrule
\texttt{verifier} & $\rho$ & $0.292^{**}$ & $0.308^{**}$ & $0.163$ & $0.223^{*}$ & $0.060$ & $0.289^{**}$ \\
\bottomrule
\end{tabular}
\end{minipage}%
}
\end{table}

\begin{sidewaystable}[p]
\centering
\small
\setlength{\tabcolsep}{3pt}
\caption{Correlations on OpenApps, state-value, text modality, using a scripted policy with Max-Value Monte Carlo for label generation.}
\label{tab:corr_openapps_sv_text_scripted}
\begin{tabular}{l l c c c c c c}
\toprule
Method & Metric & \rotatebox{45}{Gemma4 26B-A4B} & \rotatebox{45}{Gemma4 31B} & \rotatebox{45}{Qwen3.5 9B} & \rotatebox{45}{Qwen3.5 27B} & \rotatebox{45}{Qwen3.5 35B-A3B} & \rotatebox{45}{Qwen3.5 122B-A10B} \\
\midrule
\texttt{gvl} & $\rho$ & $0.037$ & $-0.070$ & $-0.310^{**}$ & $0.376^{***}$ & $0.014$ & $0.316^{**}$ \\
\texttt{direct-single} & $\rho$ & $0.187^{*}$ & $0.139$ & $0.271^{**}$ & $0.152$ & $0.110$ & $0.188^{*}$ \\
\texttt{direct-batched} & $\rho$ & $0.363^{***}$ & $0.509^{***}$ & $0.046$ & $0.423^{***}$ & $0.330^{**}$ & $0.255^{*}$ \\
\texttt{direct-sequential} & $\rho$ & $0.140$ & $0.131$ & $0.072$ & $0.227^{*}$ & $0.208^{*}$ & $0.137$ \\
\texttt{direct-16} & $\rho$ & $0.296^{**}$ & $0.578^{***}$ & $0.385^{***}$ & $0.478^{***}$ & $0.210^{*}$ & $0.419^{***}$ \\
\midrule
\texttt{eureka} & $\rho$ & $0.567^{***}$ & $0.440^{***}$ & $-0.053$ & $-0.120$ & $0.429^{***}$ & $0.528^{***}$ \\
\texttt{codegen} & $\rho$ & \makecell{$0.277 \pm 0.218$ \\ $[-0.050, 0.564]$} & \makecell{$0.267 \pm 0.167$ \\ $[-0.043, 0.555]$} & \makecell{$0.051 \pm 0.187$ \\ $[-0.208, 0.383]$} & \makecell{$0.094 \pm 0.203$ \\ $[-0.190, 0.449]$} & \makecell{$0.246 \pm 0.198$ \\ $[-0.182, 0.493]$} & \makecell{$0.140 \pm 0.206$ \\ $[-0.209, 0.482]$} \\
\texttt{codegen-avg} & $\rho$ & $0.535^{***}$ & $0.440^{***}$ & $0.064$ & $0.306^{**}$ & $0.353^{***}$ & $0.013$ \\
\bottomrule
\end{tabular}
\end{sidewaystable}

\begin{sidewaystable}[p]
\centering
\small
\setlength{\tabcolsep}{3pt}
\caption{Correlations on ALFWorld, Q-value, text modality, using a scripted policy with Max-Value Monte Carlo for label generation.}
\label{tab:corr_alfworld_qv_text_scripted}
\begin{tabular}{l l c c c c c c}
\toprule
Method & Metric & \rotatebox{45}{Gemma4 26B-A4B} & \rotatebox{45}{Gemma4 31B} & \rotatebox{45}{Qwen3.5 9B} & \rotatebox{45}{Qwen3.5 27B} & \rotatebox{45}{Qwen3.5 35B-A3B} & \rotatebox{45}{Qwen3.5 122B-A10B} \\
\midrule
\texttt{gvl} & $\rho$ & $0.327^{***}$ & $0.611^{***}$ & $0.515^{***}$ & $0.413^{***}$ & $0.463^{***}$ & $0.317^{**}$ \\
\texttt{direct-single} & $\rho$ & $0.598^{***}$ & $0.665^{***}$ & $0.567^{***}$ & $0.625^{***}$ & $0.587^{***}$ & $0.546^{***}$ \\
\texttt{direct-batched} & $\rho$ & $0.563^{***}$ & $0.649^{***}$ & $0.594^{***}$ & $0.594^{***}$ & $0.595^{***}$ & $0.206^{*}$ \\
\texttt{direct-sequential} & $\rho$ & $0.678^{***}$ & $0.716^{***}$ & $0.564^{***}$ & $0.623^{***}$ & $0.519^{***}$ & $0.436^{***}$ \\
\texttt{direct-16} & $\rho$ & $0.658^{***}$ & $0.738^{***}$ & $0.570^{***}$ & $0.585^{***}$ & $0.388^{***}$ & $0.647^{***}$ \\
\midrule
\texttt{eureka} & $\rho$ & $-0.063$ & $-0.029$ & $0.110$ & $0.389^{***}$ & $0.084$ & $0.208^{*}$ \\
\texttt{codegen} & $\rho$ & \makecell{$-0.054 \pm 0.259$ \\ $[-0.313, 0.397]$} & \makecell{$-0.012 \pm 0.227$ \\ $[-0.342, 0.356]$} & \makecell{$0.125 \pm 0.192$ \\ $[-0.103, 0.488]$} & \makecell{$0.205 \pm 0.151$ \\ $[-0.020, 0.469]$} & \makecell{$0.036 \pm 0.157$ \\ $[-0.284, 0.249]$} & \makecell{$0.080 \pm 0.189$ \\ $[-0.359, 0.365]$} \\
\texttt{codegen-avg} & $\rho$ & $-0.207^{*}$ & $0.091$ & $0.251^{*}$ & $0.169^{*}$ & $0.274^{**}$ & $0.231^{*}$ \\
\midrule
\texttt{sdpo} & $\rho$ & $-0.138 \pm 0.626$ & $-0.195 \pm 0.619$ & $-0.226 \pm 0.568$ & $-0.172 \pm 0.630$ & $-0.200 \pm 0.575$ & $-0.209 \pm 0.637$ \\
\texttt{sdpo-gt} & $\rho$ & $-0.129 \pm 0.631$ & $-0.209 \pm 0.598$ & $-0.181 \pm 0.587$ & $-0.195 \pm 0.613$ & $-0.187 \pm 0.551$ & --- \\
\midrule
\texttt{verifier} & $\rho$ & $0.592^{***}$ & $0.676^{***}$ & $0.495^{***}$ & $0.621^{***}$ & $0.506^{***}$ & $0.617^{***}$ \\
\texttt{$\Delta$belief} & $\rho$ & $0.333 \pm 0.639$ & $0.423 \pm 0.551$ & $0.321 \pm 0.674$ & $0.429 \pm 0.562$ & $0.416 \pm 0.535$ & $0.386 \pm 0.654$ \\
\midrule
\texttt{ranking} & $\rho$ & $0.456 \pm 0.492$ & $0.513 \pm 0.559$ & $0.501 \pm 0.520$ & $0.483 \pm 0.564$ & $0.538 \pm 0.555$ & $0.513 \pm 0.583$ \\
\bottomrule
\end{tabular}
\end{sidewaystable}

\begin{sidewaystable}[p]
\centering
\small
\setlength{\tabcolsep}{3pt}
\caption{Correlations on ALFWorld, Q-value, vision modality, using a scripted policy with Max-Value Monte Carlo for label generation.}
\label{tab:corr_alfworld_qv_vision_scripted}
\begin{tabular}{l l c c c c c c c}
\toprule
 &  & \multicolumn{6}{c}{LLM backbones} & \multicolumn{1}{c}{Non-LLM} \\
\cmidrule(lr){3-8}\cmidrule(lr){9-9}
Method & Metric & \rotatebox{45}{Gemma4 26B-A4B} & \rotatebox{45}{Gemma4 31B} & \rotatebox{45}{Qwen3.5 9B} & \rotatebox{45}{Qwen3.5 27B} & \rotatebox{45}{Qwen3.5 35B-A3B} & \rotatebox{45}{Qwen3.5 122B-A10B} & \rotatebox{45}{Self} \\
\midrule
\texttt{gvl} & $\rho$ & $0.275^{**}$ & $0.385^{***}$ & --- & $0.162$ & $0.080$ & $0.292^{**}$ & --- \\
\texttt{direct-single} & $\rho$ & $0.446^{***}$ & $0.454^{***}$ & $0.228^{*}$ & $0.266^{**}$ & $0.028$ & $0.322^{**}$ & --- \\
\texttt{direct-batched} & $\rho$ & $0.302^{**}$ & $0.210^{*}$ & --- & $0.215^{*}$ & --- & $0.151$ & --- \\
\texttt{direct-sequential} & $\rho$ & $0.136$ & $0.377^{***}$ & --- & $0.111$ & $-0.173$ & $0.202^{*}$ & --- \\
\texttt{direct-16} & $\rho$ & $0.269^{**}$ & $0.546^{***}$ & $0.155$ & $0.279^{**}$ & $0.150$ & $0.472^{***}$ & --- \\
\midrule
\texttt{ranking} & $\rho$ & $0.197 \pm 0.630$ & $0.122 \pm 0.630$ & $0.154 \pm 0.662$ & $0.147 \pm 0.663$ & --- & $0.184 \pm 0.636$ & --- \\
\midrule
\texttt{vip} & $\rho$ & --- & --- & --- & --- & --- & --- & $0.302^{**}$ \\
\texttt{liv-cos} & $\rho$ & --- & --- & --- & --- & --- & --- & $0.288^{**}$ \\
\texttt{liv-l2} & $\rho$ & --- & --- & --- & --- & --- & --- & $0.241^{*}$ \\
\texttt{liv-txt} & $\rho$ & --- & --- & --- & --- & --- & --- & $0.049$ \\
\bottomrule
\end{tabular}
\end{sidewaystable}

\begin{sidewaystable}[p]
\centering
\small
\setlength{\tabcolsep}{3pt}
\caption{Correlations on ALFWorld, state-value, text modality, using a scripted policy with Max-Value Monte Carlo for label generation.}
\label{tab:corr_alfworld_sv_text_scripted}
\begin{tabular}{l l c c c c c c}
\toprule
Method & Metric & \rotatebox{45}{Gemma4 26B-A4B} & \rotatebox{45}{Gemma4 31B} & \rotatebox{45}{Qwen3.5 9B} & \rotatebox{45}{Qwen3.5 27B} & \rotatebox{45}{Qwen3.5 35B-A3B} & \rotatebox{45}{Qwen3.5 122B-A10B} \\
\midrule
\texttt{gvl} & $\rho$ & $0.185^{*}$ & $0.436^{***}$ & $0.142$ & $0.326^{***}$ & $0.161$ & $0.211^{*}$ \\
\texttt{direct-single} & $\rho$ & $0.560^{***}$ & $0.469^{***}$ & $0.386^{***}$ & $0.410^{***}$ & $0.480^{***}$ & $0.335^{***}$ \\
\texttt{direct-batched} & $\rho$ & $0.329^{**}$ & $0.458^{***}$ & $0.443^{***}$ & $0.318^{**}$ & $0.196^{*}$ & $0.332^{***}$ \\
\texttt{direct-sequential} & $\rho$ & $0.500^{***}$ & $0.449^{***}$ & $0.394^{***}$ & $0.590^{***}$ & $0.460^{***}$ & $0.463^{***}$ \\
\texttt{direct-16} & $\rho$ & $0.423^{***}$ & $0.628^{***}$ & $0.457^{***}$ & $0.459^{***}$ & $0.453^{***}$ & $0.492^{***}$ \\
\midrule
\texttt{eureka} & $\rho$ & $0.095$ & $0.158$ & $0.275^{**}$ & $0.356^{***}$ & $0.083$ & $0.243^{*}$ \\
\texttt{codegen} & $\rho$ & \makecell{$-0.223 \pm 0.112$ \\ $[-0.302, -0.144]$} & \makecell{$0.029 \pm 0.190$ \\ $[-0.302, 0.181]$} & \makecell{$0.161 \pm 0.188$ \\ $[-0.172, 0.398]$} & \makecell{$0.022 \pm 0.147$ \\ $[-0.255, 0.327]$} & \makecell{$0.108 \pm 0.192$ \\ $[-0.262, 0.363]$} & \makecell{$0.160 \pm 0.171$ \\ $[-0.175, 0.443]$} \\
\texttt{codegen-avg} & $\rho$ & $-0.239^{*}$ & $0.015$ & $0.361^{***}$ & $0.075$ & $0.335^{***}$ & $0.275^{**}$ \\
\bottomrule
\end{tabular}
\end{sidewaystable}

\begin{sidewaystable}[p]
\centering
\small
\setlength{\tabcolsep}{3pt}
\caption{Correlations on FrozenLake, state-value, text modality, using a scripted policy with Max-Value Monte Carlo for label generation.}
\label{tab:corr_frozenlake_sv_text_scripted}
\begin{tabular}{l l c c c c c c}
\toprule
Method & Metric & \rotatebox{45}{Gemma4 26B-A4B} & \rotatebox{45}{Gemma4 31B} & \rotatebox{45}{Qwen3.5 9B} & \rotatebox{45}{Qwen3.5 27B} & \rotatebox{45}{Qwen3.5 35B-A3B} & \rotatebox{45}{Qwen3.5 122B-A10B} \\
\midrule
\texttt{gvl} & $\rho$ & $0.170$ & $0.001$ & $0.236^{*}$ & $-0.038$ & $0.007$ & $0.240^{*}$ \\
\texttt{direct-single} & $\rho$ & $0.116$ & $0.065$ & $0.503^{***}$ & $0.022$ & $0.178^{*}$ & $0.263^{**}$ \\
\texttt{direct-batched} & $\rho$ & --- & $0.508^{***}$ & $0.558^{***}$ & $0.303^{*}$ & $-0.087$ & $0.193^{*}$ \\
\texttt{direct-sequential} & $\rho$ & $0.082$ & $0.453^{***}$ & $0.457^{***}$ & $0.211^{*}$ & $0.226^{*}$ & $0.241^{*}$ \\
\texttt{direct-16} & $\rho$ & $0.017$ & $0.150$ & $0.608^{***}$ & $-0.007$ & $0.653^{***}$ & $0.180^{*}$ \\
\midrule
\texttt{eureka} & $\rho$ & $0.805^{***}$ & $0.911^{***}$ & $0.986^{***}$ & $0.873^{***}$ & $0.803^{***}$ & $0.940^{***}$ \\
\texttt{codegen} & $\rho$ & \makecell{$0.831 \pm 0.177$ \\ $[0.431, 1.000]$} & \makecell{$0.899 \pm 0.092$ \\ $[0.668, 0.987]$} & \makecell{$0.810 \pm 0.216$ \\ $[0.296, 0.991]$} & \makecell{$0.847 \pm 0.172$ \\ $[0.384, 0.973]$} & \makecell{$0.825 \pm 0.190$ \\ $[0.421, 0.985]$} & \makecell{$0.877 \pm 0.122$ \\ $[0.543, 0.973]$} \\
\texttt{codegen-avg} & $\rho$ & $0.874^{***}$ & $0.942^{***}$ & $0.937^{***}$ & $0.921^{***}$ & $0.932^{***}$ & $0.933^{***}$ \\
\bottomrule
\end{tabular}
\end{sidewaystable}

\begin{sidewaystable}[p]
\centering\small
\setlength{\tabcolsep}{4pt}
\caption{Global Spearman on TerminalBench, text modality. The four ranking-style methods are reported with state-local Spearman.}
\label{tab:global_spearman_terminal_bench_text}
\begin{tabular}{lcccccc}
\toprule
Method & \rotatebox{45}{Gemma4 26B-A4B} & \rotatebox{45}{Gemma4 31B} & \rotatebox{45}{Qwen3.5 9B} & \rotatebox{45}{Qwen3.5 27B} & \rotatebox{45}{Qwen3.5 35B-A3B} & \rotatebox{45}{Qwen3.5 122B-A10B} \\
\midrule
\texttt{gvl} & --- & --- & $0.166$ & $0.258$ & $0.274$ & $0.216$ \\
\texttt{direct-single} & $0.372$ & $0.393$ & $0.266$ & $0.296$ & $0.337$ & $0.365$ \\
\texttt{direct-batched} & $0.376$ & $0.358$ & $0.329$ & $0.326$ & $0.330$ & $0.296$ \\
\texttt{direct-sequential} & --- & $0.439$ & $0.369$ & $0.344$ & $0.377$ & $0.348$ \\
\texttt{direct-16} & $0.426$ & $0.435$ & $0.372$ & $0.331$ & $0.392$ & $0.378$ \\
\texttt{eureka} & $0.013$ & $-0.257$ & --- & $-0.179$ & $-0.218$ & $-0.038$ \\
\texttt{codegen} & \makecell{$-0.172 \pm 0.052$ \\ $[-0.268, -0.056]$} & \makecell{$-0.123 \pm 0.071$ \\ $[-0.254, -0.035]$} & \makecell{$-0.164 \pm 0.086$ \\ $[-0.313, -0.032]$} & \makecell{$-0.171 \pm 0.056$ \\ $[-0.294, -0.059]$} & \makecell{$-0.102 \pm 0.069$ \\ $[-0.218, 0.065]$} & \makecell{$-0.170 \pm 0.059$ \\ $[-0.257, -0.055]$} \\
\texttt{codegen-avg} & $-0.237$ & $-0.162$ & $-0.245$ & $-0.207$ & $-0.150$ & $-0.210$ \\
\texttt{verifier} & $0.305$ & $0.265$ & $0.093$ & $0.162$ & $0.038$ & $0.223$ \\
\bottomrule
\end{tabular}
\end{sidewaystable}

\begin{sidewaystable}[p]
\centering\small
\setlength{\tabcolsep}{4pt}
\caption{Global Spearman on OpenApps, text modality. The four ranking-style methods are reported with state-local Spearman.}
\label{tab:global_spearman_open_apps_text}
\begin{tabular}{lcccccc}
\toprule
Method & \rotatebox{45}{Gemma4 26B-A4B} & \rotatebox{45}{Gemma4 31B} & \rotatebox{45}{Qwen3.5 9B} & \rotatebox{45}{Qwen3.5 27B} & \rotatebox{45}{Qwen3.5 35B-A3B} & \rotatebox{45}{Qwen3.5 122B-A10B} \\
\midrule
\texttt{gvl} & $-0.051$ & $-0.020$ & $-0.029$ & $0.019$ & $0.000$ & $0.136$ \\
\texttt{direct-single} & $-0.036$ & $-0.045$ & $0.093$ & $-0.022$ & $-0.009$ & $0.010$ \\
\texttt{direct-batched} & $0.075$ & $-0.022$ & $0.036$ & $0.022$ & $0.053$ & $0.058$ \\
\texttt{direct-sequential} & $0.094$ & $0.017$ & $-0.002$ & $0.022$ & $0.149$ & $0.077$ \\
\texttt{direct-16} & $0.072$ & $0.089$ & $0.104$ & $-0.039$ & $0.100$ & $0.134$ \\
\texttt{eureka} & $0.027$ & $0.054$ & --- & $-0.134$ & $0.069$ & $0.066$ \\
\texttt{codegen} & \makecell{$-0.031 \pm 0.117$ \\ $[-0.185, 0.341]$} & \makecell{$-0.035 \pm 0.090$ \\ $[-0.243, 0.101]$} & --- & \makecell{$-0.023 \pm 0.134$ \\ $[-0.190, 0.261]$} & \makecell{$0.020 \pm 0.171$ \\ $[-0.240, 0.330]$} & \makecell{$-0.066 \pm 0.133$ \\ $[-0.338, 0.183]$} \\
\texttt{codegen-avg} & $-0.078$ & $0.035$ & $-0.075$ & $-0.124$ & $0.120$ & $-0.141$ \\
\texttt{verifier} & $0.107$ & $0.028$ & $-0.015$ & $0.027$ & $-0.026$ & $-0.039$ \\
\bottomrule
\end{tabular}
\end{sidewaystable}

\begin{table}[htbp]
\centering\small
\setlength{\tabcolsep}{4pt}
\caption{Global Spearman on OpenApps, vision modality. The four ranking-style methods are reported with state-local Spearman.}
\label{tab:global_spearman_open_apps_vision}
\begin{tabular}{lccc}
\toprule
Method & \rotatebox{45}{Self} & \rotatebox{45}{CLIP} & \rotatebox{45}{SigLIP} \\
\midrule
\texttt{vlm-rm} & --- & $-0.082$ & $0.024$ \\
\texttt{vlm-rm-cos} & --- & $-0.126$ & $0.025$ \\
\texttt{vlm-sor-softmax} & --- & $-0.013$ & $0.069$ \\
\texttt{vip} & $0.158$ & --- & --- \\
\texttt{liv-cos} & $0.069$ & --- & --- \\
\texttt{liv-l2} & $0.073$ & --- & --- \\
\texttt{liv-txt} & $-0.131$ & --- & --- \\
\bottomrule
\end{tabular}
\end{table}

\begin{sidewaystable}[p]
\centering\small
\setlength{\tabcolsep}{4pt}
\caption{Global Spearman on ALFWorld, text modality. The four ranking-style methods are reported with state-local Spearman.}
\label{tab:global_spearman_alfworld_text}
\begin{tabular}{lcccccc}
\toprule
Method & \rotatebox{45}{Gemma4 26B-A4B} & \rotatebox{45}{Gemma4 31B} & \rotatebox{45}{Qwen3.5 9B} & \rotatebox{45}{Qwen3.5 27B} & \rotatebox{45}{Qwen3.5 35B-A3B} & \rotatebox{45}{Qwen3.5 122B-A10B} \\
\midrule
\texttt{gvl} & $0.380$ & $0.522$ & $0.364$ & $0.394$ & $0.318$ & $0.327$ \\
\texttt{direct-single} & $0.686$ & $0.701$ & $0.432$ & $0.616$ & $0.548$ & $0.454$ \\
\texttt{direct-batched} & $0.599$ & $0.643$ & $0.463$ & $0.561$ & $0.584$ & $0.433$ \\
\texttt{direct-sequential} & $0.535$ & $0.634$ & $0.414$ & $0.512$ & $0.575$ & $0.419$ \\
\texttt{direct-16} & $0.760$ & $0.755$ & $0.605$ & $0.769$ & $0.693$ & $0.667$ \\
\texttt{eureka} & --- & --- & $0.061$ & $0.220$ & $0.220$ & $0.186$ \\
\texttt{codegen-avg} & $0.102$ & $-0.022$ & $0.205$ & $0.119$ & $0.259$ & $0.176$ \\
\texttt{verifier} & $0.514$ & $0.553$ & $0.407$ & $0.556$ & $0.315$ & $0.426$ \\
\bottomrule
\end{tabular}
\end{sidewaystable}

\begin{table}[htbp]
\centering\small
\setlength{\tabcolsep}{4pt}
\caption{Global Spearman on ALFWorld, vision modality. The four ranking-style methods are reported with state-local Spearman.}
\label{tab:global_spearman_alfworld_vision}
\begin{tabular}{lccc}
\toprule
Method & \rotatebox{45}{Self} & \rotatebox{45}{CLIP} & \rotatebox{45}{SigLIP} \\
\midrule
\texttt{vlm-rm} & --- & $0.342$ & $-0.022$ \\
\texttt{vlm-rm-cos} & --- & $0.443$ & $0.032$ \\
\texttt{vlm-sor-softmax} & --- & $0.444$ & $0.022$ \\
\texttt{vip} & $0.430$ & --- & --- \\
\texttt{liv-cos} & $0.330$ & --- & --- \\
\texttt{liv-l2} & $0.172$ & --- & --- \\
\texttt{liv-txt} & $0.187$ & --- & --- \\
\bottomrule
\end{tabular}
\end{table}

\begin{sidewaystable}[p]
\centering\small
\setlength{\tabcolsep}{4pt}
\caption{Global Spearman on FrozenLake, text modality. The four ranking-style methods are reported with state-local Spearman.}
\label{tab:global_spearman_frozen_lake_text}
\begin{tabular}{lcccccc}
\toprule
Method & \rotatebox{45}{Gemma4 26B-A4B} & \rotatebox{45}{Gemma4 31B} & \rotatebox{45}{Qwen3.5 9B} & \rotatebox{45}{Qwen3.5 27B} & \rotatebox{45}{Qwen3.5 35B-A3B} & \rotatebox{45}{Qwen3.5 122B-A10B} \\
\midrule
\texttt{gvl} & $-0.098$ & $0.530$ & $0.442$ & $0.349$ & $0.179$ & $0.569$ \\
\texttt{direct-single} & $0.148$ & $0.575$ & $0.427$ & $0.427$ & $0.235$ & $0.547$ \\
\texttt{direct-batched} & $-0.045$ & $0.465$ & $0.523$ & $0.519$ & $0.282$ & $0.322$ \\
\texttt{direct-sequential} & $0.170$ & $0.661$ & $0.365$ & $0.441$ & $0.095$ & $0.400$ \\
\texttt{direct-16} & $0.039$ & $0.688$ & $0.756$ & $0.704$ & $0.450$ & $0.495$ \\
\texttt{eureka} & $0.896$ & $1.000$ & $0.274$ & $0.858$ & $0.929$ & $0.904$ \\
\texttt{codegen} & \makecell{$0.913 \pm 0.110$ \\ $[0.607, 0.994]$} & \makecell{$0.976 \pm 0.039$ \\ $[0.825, 0.994]$} & \makecell{$0.431 \pm 0.269$ \\ $[0.066, 0.825]$} & \makecell{$0.770 \pm 0.086$ \\ $[0.613, 0.923]$} & \makecell{$0.842 \pm 0.114$ \\ $[0.587, 0.982]$} & \makecell{$0.694 \pm 0.147$ \\ $[0.316, 0.876]$} \\
\texttt{codegen-avg} & $0.981$ & $0.987$ & $0.757$ & $0.856$ & $0.907$ & $0.901$ \\
\texttt{verifier} & $0.306$ & $0.368$ & $0.007$ & $0.106$ & $-0.034$ & $-0.073$ \\
\bottomrule
\end{tabular}
\end{sidewaystable}

\begin{table}[htbp]
\centering\small
\setlength{\tabcolsep}{4pt}
\caption{Global Spearman on FrozenLake, vision modality. The four ranking-style methods are reported with state-local Spearman.}
\label{tab:global_spearman_frozen_lake_vision}
\begin{tabular}{lccc}
\toprule
Method & \rotatebox{45}{Self} & \rotatebox{45}{CLIP} & \rotatebox{45}{SigLIP} \\
\midrule
\texttt{vlm-rm} & --- & $-0.393$ & $-0.198$ \\
\texttt{vlm-rm-cos} & --- & $-0.276$ & $-0.120$ \\
\texttt{vlm-sor-softmax} & --- & $-0.433$ & $-0.172$ \\
\texttt{vip} & $-0.089$ & --- & --- \\
\texttt{liv-cos} & $-0.290$ & --- & --- \\
\texttt{liv-l2} & $-0.283$ & --- & --- \\
\texttt{liv-txt} & $0.088$ & --- & --- \\
\bottomrule
\end{tabular}
\end{table}

\clearpage
\section{Environment details}
\label{app::environments}

\begin{figure}[h]
\centering
\begin{subfigure}[t]{0.48\linewidth}\includegraphics[width=\linewidth]{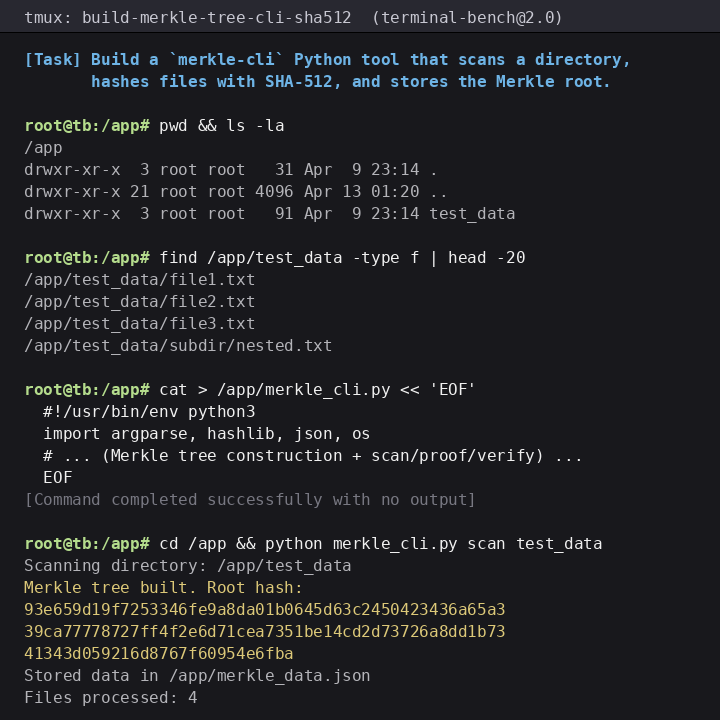}\caption{TerminalBench}\end{subfigure}\hfill
\begin{subfigure}[t]{0.48\linewidth}\includegraphics[width=\linewidth]{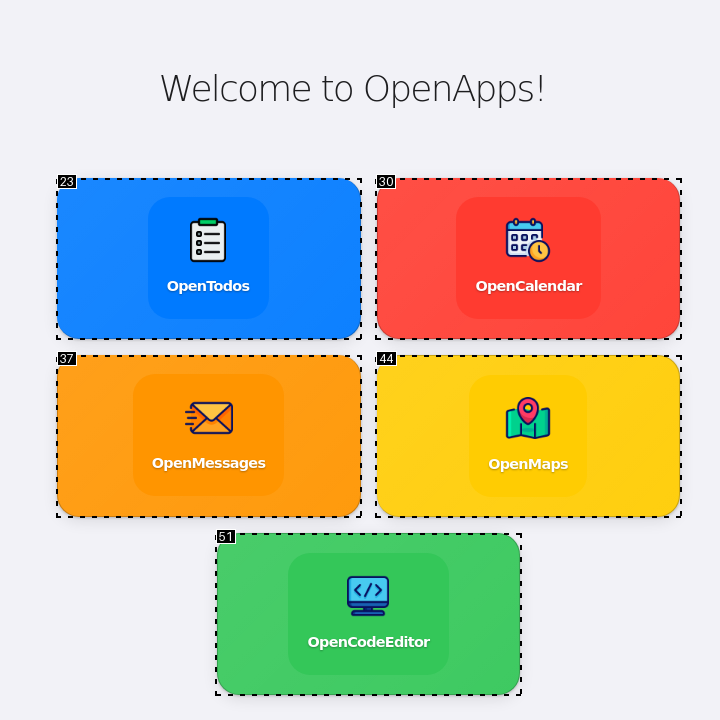}\caption{OpenApps}\end{subfigure}

\vspace{0.5em}

\begin{subfigure}[t]{0.48\linewidth}\includegraphics[width=\linewidth]{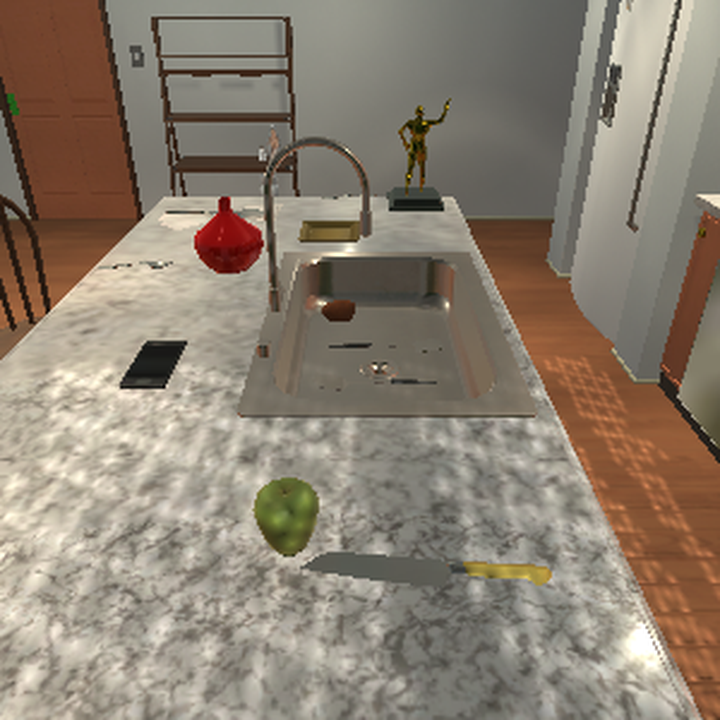}\caption{ALFWorld}\end{subfigure}\hfill
\begin{subfigure}[t]{0.48\linewidth}\includegraphics[width=\linewidth]{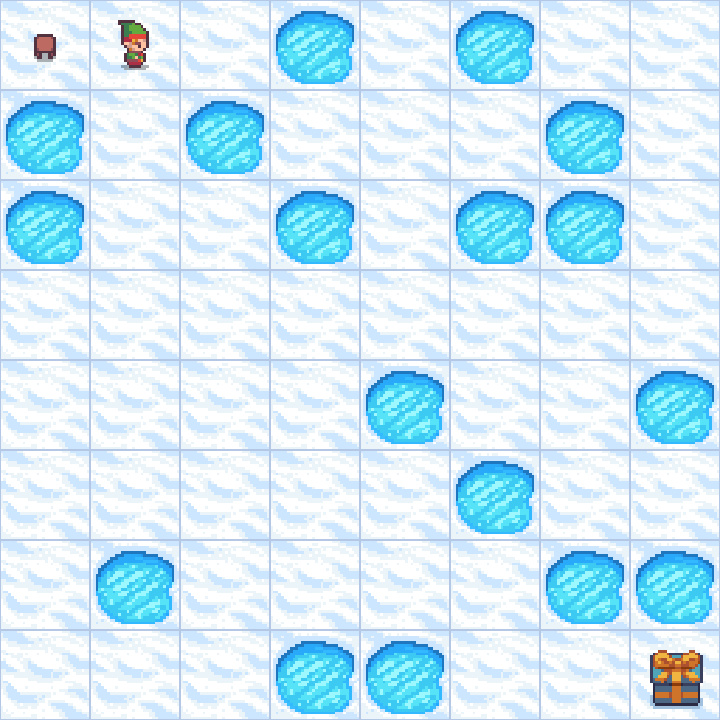}\caption{FrozenLake}\end{subfigure}

\caption{Representative state observations for each environment. TerminalBench is text-only and the image is for presentation only; the remaining three environments support either text or image observations, with the rendered image shown.}
\label{fig:env_renderings}
\end{figure}

This appendix details the four environments comprising \benchname{} at release. For each, we describe the source, observation and action spaces, reward function, episode horizon, and the parameters used for trajectory collection and reference labelling. Table~\ref{tab:env_summary} summarises the per-environment configuration; Figure~\ref{fig:env_renderings} shows a
representative state for each environment.

\begin{table}[h]
\centering
\caption{\benchname{} environment summary. Each evaluated state has a set of four candidate actions. LLM-based methods are evaluated on the first $100$ states (all $94$ for OpenApps); the encoder methods use all states of the image-capable environments. \emph{Candidates}: \emph{LLM} $=$ alternatives sampled using DeepSeek v3.2 as a policy; \emph{enumerated} $=$ generated by an environment-specific sampler. \emph{Reference-tied states} are states whose candidates all have the same reference $Q$-value, counted on the LLM state set (encoder set in parentheses).}
\label{tab:env_summary}
\footnotesize
\setlength{\tabcolsep}{4pt}
\begin{tabular}{l@{\hspace{6pt}}cccc}
\toprule
                  & \textbf{TerminalBench}       & \textbf{OpenApps}            & \textbf{ALFWorld}             & \textbf{FrozenLake}          \\
\midrule
Source            & TBLite (easy)       & OpenApps (BrowserGym) & ALFWorld OOD       & Gymnasium 8$\times$8 \\
Modalities        & text                & text, image         & text, image         & text, image        \\
\# distinct tasks & 19                  & 8                   & 24 games, 4 scenes  & 8 maps              \\
\# trajectories   & 118                 & 40                  & 40                  & 50                  \\
\# successful trajectories & 60        & 37                  & 34                  & 41                  \\
\# states (LLM methods) & 100           & 94                  & 100                 & 100                 \\
\# states (encoder methods) & --        & 94                  & 106                 & 239                 \\
Candidates per state & 4 (LLM)          & 4 (enumerated)      & 3--4 (LLM)          & 4 (all actions)     \\
Reference-tied states & 25              & 3                   & 43 (47)             & 0 (0)               \\
Episode horizon   & 40 steps            & 45 steps            & 40 steps            & 30 steps            \\
Collection actor  & DeepSeek v3.2       & scripted ($\varepsilon{=}0.25$) & DeepSeek v3.2 & scripted ($\varepsilon{=}0.1$) \\
Reference policy  & GPT-5.5, best of 16 & scripted            & PDDL planner        & value iteration     \\
\bottomrule
\end{tabular}
\end{table}

\subsection{TerminalBench}
\label{app::env_terminalbench}

We use the \emph{easy} difficulty split of TBLite~\citep{OpenThoughts-TBLite}, a curated subset of TerminalBench~\citep{merrill_terminal-bench_2025}. We specifically choose TBLite over the full TerminalBench as we find many tasks to be unsolvable even by the best current models, mostly due to timeouts cancelling long operations (e.g., package installs in R). We use 19 of the 22 tasks in the split, spanning system administration, cryptography, log processing, data engineering, and ML pipeline tasks: \texttt{amuse-install}, \texttt{broken-python},
\texttt{build-merkle-tree-cli-sha512}, \texttt{convolutional-layers},
\texttt{cosign-keyless-signing}, \texttt{cryptographic-protocol-verifier},
\texttt{hydra-debug-slurm-mode}, \texttt{image-tile-identification},
\texttt{jq-data-processing}, \texttt{jsonl-aggregator}, \texttt{log-summary},
\texttt{mlflow-register}, \texttt{pandas-etl},
\texttt{playing-card-recognition}, \texttt{protein-sequence},
\texttt{raft-log-repair-concurrent-access}, \texttt{schedule-vacation},
\texttt{smiles-data-lab}, \texttt{systemd-log-monitoring}.

\paragraph{Observation/Action.} The agent operates a persistent \texttt{tmux}-backed shell inside an Apptainer container sandbox for safety. Observations are the trailing terminal scrollback after each command; valid actions are any valid shell command (in text). Shell state (current directory, exported variables, background processes) persists across
commands within a trajectory.

\paragraph{Reward.} Binary verifier output: $1.0$ if the task verifier passes on submission or step truncation, $0.0$ otherwise.

\paragraph{Collection.} Trajectories are collected with DeepSeek v3.2~\citep{deepseekai2025deepseekv32} as the actor (thinking disabled, temperature $0.1$), with a maximum of 40 steps per episode, 180s per command, and a 900s wall-clock cap per trajectory that also counts model generation time. From nine trajectories per task, we retain one state per trajectory, sampled uniformly at random from the turns outside the first and last $3$ (trajectories too short for this use all turns). For each state, we sample $3$ alternative commands from the same model at temperature $0.2$ and top-$p$ $1.0$, and keep only states with $4$ distinct candidates. This leaves $118$ trajectories ($60$ successful, $58$ failed) with one state each; the first $100$ states are evaluated.

\paragraph{Reference policy.}
\label{app:tb_optimal}
An optimal policy is not available for TerminalBench, so we label each candidate by Max-Value Monte Carlo: after the candidate command is executed, GPT-5.5~\citep{openai_gpt-55_2026} (low reasoning effort) completes $k=16$ continuations, and the label is the highest discounted return ($\gamma = 0.95$) among them. At least one continuation succeeds for $374$ of the $400$ candidates, including $96$ of the $100$ actor actions. Appendix~\ref{app::reference_backbone} compares these labels with labels obtained with Claude Opus 4.7 as the continuation model.

\subsection{OpenApps}
\label{app::env_openapps}

OpenApps~\citep{ullrich_openapps_2025} is a suite of synthetic web applications (calendar, todo, messenger, map, code editor). The 8 tasks are: \texttt{add\_meeting\_with\_dennis} (calendar),
\texttt{add\_christmas\_shopping\_event} (calendar),
\texttt{add\_paper\_reading\_meeting\_with\_einstein} (calendar),
\texttt{remove\_wacv\_abstract\_deadline} (calendar),
\texttt{add\_call\_mom\_to\_my\_todo} (todo),
\texttt{mark\_water\_plants\_as\_done} (todo),
\texttt{message\_bob\_to\_meet} (messenger),
\texttt{save\_paris\_to\_my\_favorite\_places} (map).

\paragraph{Observation/Action.} Each step the agent sees either an accessibility-tree (AXTree) text representation of the page or a
$1280{\times}720$ screenshot with annotations drawn on
visible interactive elements~\citep{yang2023setofmark}. Actions are: \texttt{click(element)}, \texttt{fill(element,
text)}, \texttt{select\_option(element, option)}, \texttt{scroll(x, y)}, \texttt{press(key)}, \texttt{hover(element)}, \texttt{go\_back()}, and \texttt{noop()}.

\paragraph{Reward.} Binary outcome reward: $1.0$ once the goal state is reached, $0.0$ otherwise.

\paragraph{Collection.} The collection actor is a hand-crafted scripted policy that consults the AXTree to perform each task, with random actions mixed in at $\varepsilon{=}0.25$. We collect $40$ trajectories ($5$ per task, $37$ successful), with $\text{max\_steps}{=}45$ and a 60s timeout (180s during reference labelling). We sample up to $3$ evenly spaced states per trajectory, excluding the first and last turn, and remove $13$ states whose reference value is zero, which leaves $94$ states. Candidate next states are obtained by replaying the trajectory up to the state and executing the candidate.

\paragraph{Reference policy.} The same scripted policy without
$\varepsilon$-noise serves as the reference. Because the policy and the environment are deterministic, we use a single rollout per state.

\subsection{ALFWorld}
\label{app::env_alfworld}

ALFWorld~\citep{shridhar_alfworld_2020} aligns the embodied THOR
simulator~\citep{kolve2017ai2thor} with a TextWorld interface, exposing household tasks through both natural-language observations and rendered images of the agent's egocentric view.

\paragraph{Task scope.} ALFWorld defines six task types; we restrict the dataset to \texttt{pick\_and\_place\_simple} and use all $24$ games of this type in the out-of-distribution evaluation split (\texttt{valid\_unseen}). They are set in $4$ scenes with $8$ distinct $(\text{object}, \text{receptacle})$ pairings.

\paragraph{Observation/Action.} Text observations include the current scene description and a list of admissible commands (navigation, object pickup, placement, container open/close, object-state changes such as cleaning). In the image modality, states are $300{\times}300$ first-person THOR renders at the agent's recorded location. Actions are text commands. The admissible commands are listed but not enforced: an inadmissible command leaves the state unchanged and returns ``Nothing happens.'' ($124$ of the $567$ collected transitions and $9$ of the $100$ evaluated actor actions), and a response without a parsable command becomes a no-op.

\paragraph{Reward.} Binary outcome reward: $1.0$ when the task goal is reached, $0.0$ otherwise.

\paragraph{Collection.} We collect $40$ trajectories with DeepSeek v3.2 ($34$ successful), $\text{max\_steps}{=}40$, sampling up to $5$ states per trajectory with the first turn and last $2$ turns excluded. For each state, we sample $3$ alternative commands from the same model at temperature $0.2$ and top-$p$ $1.0$, and keep only states with at least $3$ distinct candidates including the actor's action. This leaves $106$ of $153$ sampled states, $82$ with $4$ candidates and $24$ with $3$; the LLM-based methods use the first $100$.

\paragraph{Reference policy.} The reference is a symbolic planner: TextWorld's Fast Downward planner over the game's PDDL description, with unit action costs and random successors disabled. It is deterministic, so we use a single rollout per state.

\subsection{FrozenLake}
\label{app::env_frozenlake}

We use the $8{\times}8$ FrozenLake variant from
Gymnasium~\citep{towers_gymnasium_2025} with deterministic dynamics
(\texttt{is\_slippery}=False). To test agents on more than a single fixed layout, we generate $8$ distinct random maps.

\paragraph{Observation/Action.} Text observations are an $8{\times}8$ ASCII grid in which the agent's current cell is marked with \texttt{@}, holes with \texttt{H}, and the goal with \texttt{G}; the visual modality instead provides a $512{\times}512$ pygame render of the same grid. The action space is four discrete moves: \texttt{left, down, right, up}; off-grid moves leave the agent in place.

\paragraph{Reward.} Binary outcome reward: $1.0$ for reaching the goal, $0.0$ for falling into a hole or exhausting the step budget.

\paragraph{Collection.} The collection actor is a scripted shortest-path policy mixed with $\varepsilon{=}0.1$ uniform-random noise. $41$ of the $50$ collected trajectories reach the goal and $9$ fall into a hole; no trajectory is longer than $20$ steps, so the $30$-step horizon never binds. We sample up to $5$ states per trajectory, yielding $239$ states; the LLM-based methods use the first $100$ (matching the other environments) and the encoder methods use all $239$. The candidates are the four discrete actions.

\paragraph{Reference policy.} The reference is the shortest-path policy without $\varepsilon$-noise. The environment and the policy are deterministic, so we use a single rollout per state.

\clearpage
\section{Evaluation metrics}
\label{app::eval_metrics}

State-local Spearman is the main metric of \benchname. Selection loss measures the reference value sacrificed by the selected candidate, and raw and affine-corrected MSE measure calibration. Global Spearman serves as a robustness ablation, where correlation is computed over the pool of all state--action pairs in a dataset, instead of grouping pairs by shared states and averaging across state groups (state-local variant, which is the main metric as mentioned above).

\benchname evaluates every method within states. Each evaluated state $s$ has a candidate set $C_s$: the action taken by the collection actor and up to three alternative actions (Appendix~\ref{app::environments}). Each candidate $a \in C_s$ carries a reference label $Q(s,a)$: the discounted return ($\gamma = 0.95$) obtained by taking $a$ in $s$ and then following the environment's reference policy, maximised over $16$ continuations for TerminalBench (Section~\ref{sec::benchmark}, Appendix~\ref{app::environments}). It also receives a score $f(s,a)$ from the evaluated method. Pointwise methods score each candidate separately; methods that output a permutation over $C_s$ are converted to scores by negating the rank, so the top-ranked candidate receives the highest score; methods that score states rather than actions score the candidate's next state. A score that a method fails to produce (e.g.\ an extraction failure, or a runtime error in generated code) is simply recorded as missing.

We compare methods by the ordering their scores induce within a state rather than by absolute values: scores from different methods live on incompatible scales (raw LLM outputs, generated code, token log-probabilities, embedding similarities). Action selection and ranking depend only on score order and are invariant to strictly increasing score transformations. Group-normalised advantages are only affine-invariant; more general training losses can also depend on score magnitudes. We report the following metrics. Let
\begin{equation*}
U_s = \{a \in C_s : Q(s,a) \text{ and } f(s,a) \text{ are both available}\}
\end{equation*}
denote the jointly usable candidates of state $s$.

\paragraph{State-local Spearman.}
A state is \emph{valid} if $|U_s| \ge 3$ and neither $Q(s,\cdot)$ nor $f(s,\cdot)$ is constant on $U_s$. For a valid state, $\rho_s$ is Spearman's correlation~\citep{spearman_proof_1904} over $U_s$, i.e.\ the Pearson correlation~\citep{pearson_vii_1896} between the ranks of $Q(s,\cdot)$ and of $f(s,\cdot)$, with tied values assigned their average rank. With $\mathcal{V}$ the set of valid states,
\begin{equation}
\bar{\rho} \;=\; \frac{1}{|\mathcal{V}|} \sum_{s \in \mathcal{V}} \rho_s .
\label{eq::local_spearman}
\end{equation}
Because reference-tied candidates share a rank, local Spearman also penalises a method for ordering candidates that the reference considers equal.

\paragraph{Selection loss.}
Selection loss measures the reference value sacrificed by the candidate a method would select. Let $F_s \subseteq C_s$ be the candidates with an available score. The method selects
\begin{equation*}
A_s = \operatorname*{arg\,max}_{a \in F_s} f(s,a) \quad \text{(all maximisers, if tied)}, \qquad A_s = C_s \ \text{ if } F_s = \emptyset,
\end{equation*}
where the second case is a uniform fallback. With $\bar{q}_s = \frac{1}{|A_s|}\sum_{a \in A_s} Q(s,a)$ and $b_s = \max_{a \in C_s} Q(s,a)$,
\begin{equation}
\ell_s \;=\; b_s - \bar{q}_s .
\label{eq::selection}
\end{equation}
The selection loss $\ell_s \ge 0$ (lower is better) is defined for every state whose candidates all have reference labels, and is averaged over those states. Constant scores remain evaluable through uniform selection, and reference-tied states contribute $\ell_s = 0$. Loss is expressed in units of the reference return, so its magnitude is comparable only within an environment and its configuration settings.

\paragraph{Calibration.}
Raw MSE measures whether a signal's values lie on the scale of the reference values. Over the $M$ jointly usable candidate state--action pairs of a cell, with scores $f_j$ and reference values $Q_j$, it is $\frac{1}{M}\sum_{j}(f_j-Q_j)^2$. Affine-corrected MSE, $\min_{\alpha,\beta}\frac{1}{M}\sum_{j}(\alpha f_j+\beta-Q_j)^2$, is the error left after a least-squares fit of the scores to the reference values in that cell, so the gap between the two is the error due to scale and offset. A constant prediction of the mean reference value has an MSE equal to the variance of the reference values. Raw MSE is meaningful only for methods whose scores estimate returns; token probabilities, embedding similarities and pretrained value heads have a different scale by construction. Both errors use the same usable pairs. The affine fit permits either sign of $\alpha$ and is evaluated on the labels used to fit it, so it is a diagnostic rather than a held-out calibration test. For \texttt{codegen}, each saved function is fitted separately and its MSE is then averaged; \texttt{codegen-avg} is fitted after averaging the candidate scores.

\paragraph{Valid comparisons and reporting rule.}
A state is excluded for one of three reasons: its reference labels are tied, the method's scores are constant, or fewer than three candidates have usable scores. We report valid-state counts and never score excluded states as zero. Reference ties bound the number of valid states in each environment (Table~\ref{tab:env_summary}). A cell (environment, method, backbone) is reported only if it has at least $50$ valid states and these make up at least $50\%$ of the states compared. Cells that fail the rule are marked in the result tables as such.

\paragraph{Aggregation.}
Each metric of a cell is the mean over the states where that metric is defined, so the metrics can rest on slightly different sets of states. Aggregates over backbones or environments are means over reported cells; a mean pooled over environments therefore weights each environment by its number of reported cells, which is why we also report per-environment values.

\subsection{Global alignment}
\label{app::global_alignment}

The global notion of alignment requires one strictly increasing function shared across all states:
\begin{equation}
k(s,a)=\phi\!\left(Q^\pi(s,a)\right).
\label{eq:global_q_aligned}
\end{equation}
Unlike Eq.~\ref{eq:q-aligned}, this requires comparability across states and equal scores for equal reference values. A signal may order all unequal-value actions correctly within each state without satisfying this global condition. Conversely, high global correlation can arise from identifying easy versus hard states without reliably distinguishing the quality of available actions at any one state.

For $M$ jointly usable evaluation points, let $R_j$ and $S_j$ be the average ranks of the predicted scores and reference values. Global Spearman is
\begin{equation}
\rho_{\mathrm{global}}=\frac{\sum_{j=1}^M(R_j-\bar R)(S_j-\bar S)}{\sqrt{\sum_{j=1}^M(R_j-\bar R)^2\sum_{j=1}^M(S_j-\bar S)^2}}.
\end{equation}
It is undefined when either vector is constant. Pairs missing either value are excluded. Results for global Spearman are provided in Appendix~\ref{app::global_spearman}.

\subsection{Diagnostics for guided decisions}
\label{app::decision_diagnostics}

Candidate diversity, measured as the number of distinct proposals per decision, is reported in Table~\ref{tab:guided_search_candidate_diversity}. Reference-tie counts are given in Table~\ref{tab:env_summary}, with ALFWorld tie patterns discussed in Appendix~\ref{app::reference_ties}.

\clearpage
\section{Method details}
\label{app::methods}

This appendix details how every method in Section~\ref{sec::experiments::methods} is parameterised and run. Section~\ref{app::methods_common} describes the context shared by the prompt-based methods, Section~\ref{app::methods_candidates} describes how methods score the candidate set of a state, and Sections~\ref{app::methods_direct}--\ref{app::methods_pretrained} describe each method family. Table~\ref{tab:method_hparams} summarises the hyperparameters, and Section~\ref{app::methods_prompts} lists the prompt templates.

\subsection{Shared method context}
\label{app::methods_common}

Every LLM-based method receives the same environment context: a task description, the episode step limit, and a reward description (Prompt~C.1). Each datapoint consists of the current state, the action, and the resulting next state, preceded by the episode's state--action history. The history contains up to the $40$ most recent turns (none for FrozenLake, whose state is fully observed, and none for \texttt{gvl}, which has its own trajectory context). When a prompt exceeds the context budget (Appendix~\ref{app::models}), text is shortened in the middle, in this order: history observations (oldest first), history actions, the next state, and the current state. Whole turns are never dropped, and \texttt{direct-sequential} is never truncated.

All prompts define the target as the $Q$-value under optimal play thereafter. The discount factor $\gamma = 0.95$ of the reference labels is not disclosed: prompts ask for the ``expected discounted cumulative reward'' and state that shorter successful trajectories should be valued higher. All prompts except the \texttt{eureka} judge's also ask for an informed approximation rather than an exact computation. Numeric answers are read from the last \texttt{<answer>} tag of the response, so numbers in the model's reasoning are never used; a response without a parsable answer yields a missing score.

In the main results, LLM-based methods see states as text. In the vision ablation, states and next states are given as images placed where the state text would otherwise appear, and the rest of the context is unchanged. The encoder methods (Sections~\ref{app::methods_vle} and~\ref{app::methods_pretrained}) take images and goal texts directly and use no prompt.

\subsection{Scoring candidate sets}
\label{app::methods_candidates}

All methods are evaluated on the candidate set $C_s$ of each state (Appendix~\ref{app::eval_metrics}). \texttt{Ranking}, \texttt{$\Delta$belief}, \texttt{sdpo} and \texttt{sdpo-gt} are defined on candidate sets directly. Every other method scores one state--action pair at a time. Methods that score states rather than actions (the encoder methods) score $s'_a$, i.e.\ $\hat{Q}(s,a) = \hat{V}(s'_a)$. The code methods generate their functions once and apply them to every candidate.

\subsection{Direct prediction family}
\label{app::methods_direct}

The four direct variants share one datapoint template (Prompt~C.1) and differ in how datapoints are grouped into LLM calls.

\paragraph{\texttt{direct-single}.} One datapoint per prompt and one sample per datapoint.

\paragraph{\texttt{direct-16}.} The \texttt{direct-single} prompt is sent $16$ times independently. The score is the mean of the parsed answers; it is missing only if no answer parses.

\paragraph{\texttt{direct-batched}.} Four datapoints are placed in one prompt (Prompt~C.2), and the model returns a single \texttt{<answer>} tag holding exactly four comma-separated values in datapoint order. A list of any other length makes all four scores missing. Datapoints are grouped in input order, so a prompt can mix candidates from different states.

\paragraph{\texttt{direct-sequential}.} Eight datapoints are presented as successive user turns of one conversation (Prompt~C.2), each answered with one value. Earlier answers are replayed as assistant turns with the reasoning removed. If an answer fails to parse, that datapoint and all later ones in the conversation are missing. Grouping follows input order, as for \texttt{direct-batched}.

\paragraph{\texttt{gvl}.} Our adaptation of \texttt{gvl}~\citep{ma_vision_2024} scores one target transition per prompt (Prompt~C.3). The context holds all other transitions of the collected trajectory that contains the state shuffled, both earlier and later ones, as state and action only. If the prompt exceeds the budget, the context transitions farthest from the target are removed first. Unlike the original, which predicts task progress for every shuffled frame, our version asks only for the $Q$-value of the target.

\subsection{\texttt{verifier}}
\label{app::methods_verifier}

Our adaptation of LLM-as-a-Verifier~\citep{kwok2026llmverifier} asks the model to grade a datapoint on an ordered scale of $20$ letter bins, \texttt{A} (best) to \texttt{T} (worst), inside a \texttt{<score\_1>} tag (Prompt~C.4). Each prompt holds one datapoint and one evaluation criterion, and every datapoint is graded once for each criterion of its environment (Table~\ref{tab:verifier_criteria}). Bin $i \in \{0, \dots, 19\}$ (with \texttt{A} $= 0$) has value $v_i = (19 - i)/19$. At the position of the letter after the last \texttt{<score\_1>} tag, we read the top-$20$ token log-probabilities, keep the alternatives $L$ that are bin letters (optionally followed by part of the closing tag), and compute the score for criterion $c$ as
$$
\operatorname{score}_c(s,a) = \frac{\sum_{i \in L} v_i\, p_i}{\sum_{i \in L} p_i}.
$$
If no bin letter appears among these alternatives, the decoded letter is used: the one inside the score tag or, failing that, the last letter given as the score in the response text. Otherwise $\operatorname{score}_c$ is missing. If the response reaches the length limit, one follow-up request of up to $512$ tokens asks for the grade; if the follow-up is also cut off, $\operatorname{score}_c$ is missing. The method's score is the mean of the per-criterion scores that are not missing, and it is missing only if all of them are.

\begin{table}[h]
\centering
\small
\begin{tabular}{@{}l p{0.74\linewidth}@{}}
\toprule
Criterion & Description given to the model \\
\midrule
\multicolumn{2}{@{}l}{\textit{TerminalBench}} \\
Correctness & Judge how much the chosen action and resulting next state improve the chances of reaching a correct final solution that satisfies the task requirements. Reward actions that move the agent toward the right files, commands, outputs, or validation steps. \\
Error Detection & Judge whether the chosen action helps detect, avoid, or resolve errors rather than introducing or ignoring them. Penalize actions that are likely to produce unresolved failures, misleading output, or brittle workarounds. \\
Efficiency & Judge how much the chosen action and resulting next state advance the solution efficiently. Reward actions that make concrete forward progress and penalize actions that are redundant, low-yield, or likely to waste turns. \\
Resource Usage & Judge whether the chosen action uses compute, memory, runtime, and terminal output responsibly. Reward actions that are targeted and economical, and penalize actions that are likely to be expensive, spammy, or unnecessary. \\
\midrule
\multicolumn{2}{@{}l}{\textit{OpenApps}} \\
Correctness & Judge how much the chosen action and resulting next state advance the target application state. Reward actions that create, edit, or confirm the correct entity (event, todo, message, etc.) using values from the instruction; penalize actions that create wrong entities, stray into unrelated apps, or discard in-progress work. \\
Error Detection & Judge whether the chosen action avoids or resolves UI interaction failures. Penalize actions that submit incomplete forms, click disabled widgets, or are likely to trigger validation errors; reward actions that address error states, correct mistyped fields, or back out of wrong-app navigation. \\
Efficiency & Judge whether the chosen action makes concrete progress toward the goal. Reward clicks, fills, or submissions that directly change application state in the right direction; penalize redundant scrolling, reopening already-visited pages, or re-clicking widgets that produced no effect previously. \\
Input Fidelity & Judge whether the chosen action enters values that exactly match the instruction---correct recipient, date, time, name, or text string, character-for-character where applicable. Reward precise fills and clicks on exactly-matching elements; penalize typos, approximations, wrong-case strings, or selections of adjacent-but-wrong options. \\
\midrule
\multicolumn{2}{@{}l}{\textit{ALFWorld}} \\
Correctness & Judge how much the chosen action and resulting next state advance the household goal. Reward actions that pick up the target object, move toward the target receptacle, or apply the required state change (clean, heat, cool); penalize actions that target wrong objects, travel to irrelevant locations, or undo prior correct progress. \\
Error Detection & Judge whether the chosen action is a valid, effectful command given the current state. Penalize actions likely to fail---manipulating unheld or out-of-reach objects, using wrong receptacles, or issuing malformed commands; reward actions whose preconditions are satisfied and that will produce a meaningful environment change. \\
Efficiency & Judge whether the chosen action advances the task efficiently. Reward actions that directly search likely containers, move between task-relevant locations, or execute the final placement; penalize actions that revisit empty containers, re-enter rooms unnecessarily, or detour into irrelevant inspections. \\
Precondition Awareness & Judge whether the chosen action respects ALFWorld's object-manipulation preconditions. Reward actions that open containers before inspecting, pick up before placing, bring the right object to the right appliance, and sequence unlock/open/close correctly; penalize actions that skip prerequisites (e.g., place without pickup, clean without a sink, cool without a fridge). \\
\midrule
\multicolumn{2}{@{}l}{\textit{FrozenLake}} \\
Correctness & Judge how much the chosen action and resulting next position improve the chances of reaching the goal (G). Reward actions that step along a valid path from the current position toward the goal; penalize actions that step away from the goal or trap the agent in a dead-end region of the grid. \\
Hazard Avoidance & Judge whether the chosen action avoids stepping into a hole (H) or maneuvering into a cell tightly surrounded by holes. Reward actions that keep the agent on ice and move it toward hole-free regions; heavily penalize actions whose resulting cell is a hole. \\
Path Optimality & Judge whether the chosen action shortens the remaining distance to the goal along a valid path rather than backtracking, detouring, or oscillating. Reward actions that make monotonic progress toward the goal; penalize actions that move away, bounce off the grid edge, or return to a previously visited cell. \\
\bottomrule
\end{tabular}
\caption{Evaluation criteria of the \texttt{verifier} method.}
\label{tab:verifier_criteria}
\end{table}

\subsection{\texttt{$\Delta$belief}}
\label{app::methods_dbelief}

We adapt \texttt{$\Delta$belief}~\citep{auzina2026intrinsic} to candidate sets (Prompt~C.5). For each state, a \emph{pre} prompt shows the history and the current state; for each candidate, a \emph{post} prompt additionally shows the candidate action and its next state. Both end with the question whether the agent is likely to eventually succeed, assuming optimal decisions thereafter. We score the fixed continuations \texttt{Yes} and \texttt{No} (summing token log-probabilities) and normalise:
$$
\log p(\text{success} \mid C) = \log p(\texttt{Yes} \mid C) - \operatorname{lse}\big(\log p(\texttt{Yes} \mid C),\, \log p(\texttt{No} \mid C)\big),
$$
$$
\operatorname{score}(s, a) = \log p(\text{success} \mid \text{post}_a) - \log p(\text{success} \mid \text{pre}).
$$
The pre term is shared by all candidates of a state, so the within-state ordering is determined by the post term. No text is generated.

\subsection{Code generation family}
\label{app::methods_code}

\paragraph{\texttt{codegen}.} Our adaptation of Auto MC-Reward~\citep{li_auto_2024} asks the model for one Python function \texttt{signal\_function(state: str, action: str, next\_state: str) -> float} (Prompt~C.6). The prompt contains the shared context and coding constraints but no example states. We sample $16$ functions independently and take the code block from each response. A function is kept if it compiles and \texttt{signal\_function} is callable with the expected number of arguments. Functions run in a restricted namespace that allows only the imports listed in the prompt. A function that fails to parse, compile or validate yields missing scores for all datapoints; a runtime error yields a missing score for that datapoint. The reported \texttt{codegen} value of each metric is the mean over the $16$ functions of the per-function metric, and the reporting rule uses the median valid-state count across functions (Appendix~\ref{app::eval_metrics}).

\paragraph{\texttt{codegen-avg}.} The same $16$ functions, combined before evaluation: each candidate's score is the mean of the predictions of the functions that returned a value, and the metrics are computed once on these averaged scores.

\paragraph{\texttt{eureka}.} Our adaptation of \texttt{eureka}~\citep{ma_eureka_2023} runs an iterative search over the \texttt{codegen} prompt (Prompt~C.10). Each of $16$ iterations samples $8$ candidate functions. From the second iteration on, the prompt also contains the current best function and the judge's feedback. Each valid candidate is run on all evaluation datapoints (the actor's state--action pairs). An LLM judge, the same backbone at temperature $0$, then chooses the best among the new candidates and the current best. For each candidate, the judge sees summary statistics of its predictions over all evaluation datapoints, plus its predictions on $8$ datapoints sampled anew in each iteration. It never sees the code or the reference labels. The $8$ judge datapoints are drawn from the evaluated datapoints; they are not held out. The judge's choice becomes the current best, and its rationale and feedback are shown in the next iteration. The search stops early, keeping the current best, if an iteration yields no valid candidate or the judge's response cannot be used. The final best function is applied to all candidates. The original method instead scores candidates by training RL policies with them.

\subsection{Self-distillation}
\label{app::methods_sdpo}

\paragraph{\texttt{sdpo}.} Our offline adaptation of \texttt{sdpo}~\citep{hubotter_reinforcement_2026} scores a candidate by how much more likely the model is to produce its action once it sees the action's outcome (Prompt~C.7). A \emph{student} context shows the history and the current state and asks for the next action. A \emph{teacher} context additionally shows the candidate's own next state $s'_a$ as feedback. Both contexts are scored on the same backbone with the candidate action $a$ as a fixed continuation:
$$
\operatorname{score}(s, a) = \frac{1}{|a|} \sum_{t=1}^{|a|} \big[\log p(a_t \mid x_{\text{teacher}}, a_{<t}) - \log p(a_t \mid x_{\text{student}}, a_{<t})\big],
$$
where $a_t$ are the action's tokens. The score is missing if the action tokenises differently in the two contexts or a log-probability is not finite. Actions that could not be parsed from the collection model's response are shown as \texttt{[invalid action: could not parse from model response]}.

\paragraph{\texttt{sdpo-gt}.} An oracle-teacher ablation of \texttt{sdpo} (Prompt~C.8). The student is unchanged. The teacher additionally sees evidence from a stored expert continuation that starts after the candidate: the next expert action, whether the continuation reached success, the number of expert steps to success, and the final reward. The stored continuations differ by environment:
\begin{itemize}\setlength{\itemsep}{0pt}
  \item \textbf{ALFWorld}, \textbf{FrozenLake} and \textbf{OpenApps}: the reference policy's continuation from the candidate's next state, i.e., the reference rollouts behind the candidate's reference label.
  \item \textbf{TerminalBench}: DeepSeek v3.2 continuation rollouts.
\end{itemize}

\subsection{\texttt{ranking}}
\label{app::methods_ranking}

The \texttt{ranking} method shows the history, the current state, and every candidate action with its next state, numbered $1..K$ (Prompt~C.9). It asks for the candidates ordered from best to worst as a comma-separated list inside an \texttt{<answer>} tag. We parse the last list that is an exact permutation of $\{1, \dots, K\}$; partial lists and lists with repeated numbers are rejected, and then all scores of the state are missing. The permutation is converted to scores by negating the rank (Appendix~\ref{app::eval_metrics}).

\subsection{Embedding similarity (VLE) methods}
\label{app::methods_vle}

\paragraph{Encoders and goal text.} All four VLE methods use \textbf{CLIP ViT-L/14}~\citep{radford_learning_2021} or \textbf{SigLIP base}~\citep{zhai_sigmoid_2023}, both frozen. State images and goal texts are encoded with each model's own preprocessor, and embeddings are $\ell_2$-normalised. The goal text of each datapoint is:
\begin{itemize}\setlength{\itemsep}{0pt}
  \item \textbf{FrozenLake}: the environment's task text stored with the state, truncated to the encoder's text-length limit.
  \item \textbf{ALFWorld}: the task objective, e.g.\ \texttt{Objective: put a vase in safe.}
  \item \textbf{OpenApps}: the task goal, e.g.\ \texttt{Goal: Save Paris, France to my favorite places}.
\end{itemize}
With $e_s$ the state embedding and $g$ the goal embedding:

\paragraph{\texttt{vlm-rm-cos}.} $\operatorname{score} = \cos(e_s, g)$.

\paragraph{\texttt{vlm-rm}.} Goal--baseline regularisation from \texttt{vlm-rm}~\citep{rocamonde2024visionlanguage}. With $b$ the embedding of an environment-specific baseline text (\texttt{a blank frozen lake grid}, \texttt{a generic household room}, \texttt{a generic web browser home page}) and $\operatorname{proj}(e_s)$ the projection of $e_s$ onto the line through $b$ and $g$,
$$
\operatorname{score} = 1 - \tfrac{1}{2}\,\big\|\alpha \operatorname{proj}(e_s) + (1 - \alpha)\, e_s - g\big\|_2^2, \qquad \alpha = 0.5.
$$

\paragraph{\texttt{vlm-sor-softmax}.} The continuous form of \texttt{vlm-sor}~\citep{baumli2023vision}: the softmax probability of the goal among the goal and a set of negative texts $N$, at temperature $\tau = 0.07$,
$$
\operatorname{score} = \frac{\exp(e_s^\top g / \tau)}{\exp(e_s^\top g / \tau) + \sum_{n \in N} \exp(e_s^\top n / \tau)}.
$$
The negatives are:
\begin{itemize}\setlength{\itemsep}{0pt}
  \item \textbf{FrozenLake}: \texttt{the agent fell into a hole}, \texttt{an empty frozen lake}, \texttt{a random unrelated scene}.
  \item \textbf{ALFWorld}: the $13$ task objectives of the dataset, in the same format as the goal text, minus the evaluated task's own objective ($12$ negatives).
  \item \textbf{OpenApps}: the $8$ task goals, minus the evaluated task's own goal ($7$ negatives).
\end{itemize}

\paragraph{\texttt{vlm-sor}.} The original thresholded \texttt{vlm-sor} reward: $\operatorname{score} = \mathbbm{1}[p > \beta]$ with $p$ the \texttt{vlm-sor-softmax} score and $\beta = 0.5$. Because the score is binary, it is constant on the candidate set of most states.

\subsection{Pretrained value methods}
\label{app::methods_pretrained}

\paragraph{Goal images.} \texttt{Vip} and \texttt{liv} compare the state image with a goal image. For FrozenLake and OpenApps, the goal image is the last frame of the collected trajectory that contains the state. These trajectories do not always end in success (Appendix~\ref{app::environments}), so the goal image can show a failed episode. For ALFWorld, the goal image is a first-person render of a state in which the task is completed, one per game; in $8$ of the $24$ games the placed object is hidden from view (e.g.\ inside a closed cabinet).

\paragraph{\texttt{vip}.} \texttt{Vip}~\citep{ma2022vip} with the ResNet-50 checkpoint pretrained on Ego4D; the score is the negative $\ell_2$ distance between the state and goal embeddings.

\paragraph{\texttt{liv-cos} / \texttt{liv-l2}.} \texttt{Liv}~\citep{ma2023liv} in its image-goal mode, with the CLIP ResNet-50 checkpoint fine-tuned on EpicKitchens. \texttt{Liv-cos} scores by cosine similarity and \texttt{liv-l2} by negative $\ell_2$ distance to the goal-image embedding.

\paragraph{\texttt{liv-txt}.} \texttt{Liv} in its text-goal mode: cosine similarity between the state embedding and the embedding of the datapoint's goal text (as for VLE), encoded with \texttt{liv}'s text tower.

\subsection{Hyperparameter summary}
\label{app::methods_hparams}

Table~\ref{tab:method_hparams} lists the method-specific hyperparameters. Decoding settings (temperature, top-$p$, top-$k$, token limits and thinking budget) depend on the backbone and are given in Appendix~\ref{app::models}.

\begin{table}[h]
\centering
\caption{Method-specific hyperparameters. Counts are per datapoint unless stated otherwise.}
\label{tab:method_hparams}
\footnotesize
\setlength{\tabcolsep}{4pt}
\begin{tabular}{l l l}
\toprule
Method                          & Hyperparameter          & Value \\
\midrule
all LLM methods                 & history                 & last 40 turns (FrozenLake: none) \\
\texttt{direct-single}          & samples                 & 1 \\
\texttt{direct-16}              & samples                 & 16 (mean of parsed answers) \\
\texttt{direct-batched}         & datapoints per prompt   & 4 (one answer list) \\
\texttt{direct-sequential}      & datapoints per conversation & 8 (one per turn) \\
\texttt{gvl}                    & context                 & rest of trajectory, shuffled; no history \\
\midrule
\texttt{verifier}               & score bins              & 20 (\texttt{A}--\texttt{T}) \\
\texttt{verifier}               & readout                 & top-20 log-probabilities at the score letter \\
\texttt{verifier}               & criteria                & 4 per environment (FrozenLake 3), averaged \\
\texttt{$\Delta$belief}         & scored continuations    & \texttt{Yes} / \texttt{No} \\
\midrule
\texttt{codegen}                & functions               & 16 \\
\texttt{codegen-avg}            & aggregation             & mean over functions that return a value \\
\texttt{eureka}                 & iterations              & 16 \\
\texttt{eureka}                 & candidates per iteration & 8 \\
\texttt{eureka}                 & judge datapoints        & 8, resampled each iteration \\
\texttt{eureka}                 & judge                   & same backbone, temperature 0 \\
\midrule
\texttt{sdpo}                   & teacher feedback        & candidate's next state \\
\texttt{sdpo-gt}                & teacher feedback        & next state + stored expert continuation \\
\texttt{ranking}                & parser                  & last exact permutation of $\{1..K\}$ \\
\midrule
\texttt{vlm-rm}                 & $\alpha$                & 0.5 \\
\texttt{vlm-sor-softmax}        & $\tau$                  & 0.07 \\
\texttt{vlm-sor}                & $\tau$, $\beta$         & 0.07, 0.5 \\
\texttt{vlm-sor*}               & negative texts          & FrozenLake 3, ALFWorld 12, OpenApps 7 \\
\midrule
\texttt{vip}                    & backbone                & ResNet-50 (Ego4D) \\
\texttt{liv-*}                  & backbone                & CLIP ResNet-50 (EpicKitchens) \\
\texttt{vip}, \texttt{liv-cos}, \texttt{liv-l2} & goal image & last trajectory frame (ALFWorld: task goal) \\
\texttt{liv-txt}                & goal                    & datapoint's goal text \\
\bottomrule
\end{tabular}
\end{table}

\subsection{Prompt templates}
\label{app::methods_prompts}

We list the $Q$-value prompts of every LLM-based method. \texttt{\textcolor{placeholder}{\{\textit{slot}\}}} marks runtime substitutions; tags such as \texttt{<system>}, \texttt{<user>}, \texttt{<assistant>}, \texttt{<answer>} and \texttt{<score\_1>} are emitted verbatim. Blocks shown in full in Prompt~C.1 appear as slots elsewhere: \texttt{q\_value\_definition} (its first paragraph), \texttt{approximation\_guidance} and \texttt{efficiency\_guidance} (its second and third paragraphs), \texttt{task\_and\_reward\_blocks} (the Task Description, Episode Configuration and Reward Functions sections), and \texttt{state\_action\_history} (the history section, whose layout varies slightly between methods). Prompt~C.7 and~C.8 are shown for TerminalBench; the other environments use analogous environment-specific wording. The encoder methods (\texttt{vip}, \texttt{liv-*}, \texttt{vlm-*}) use no LLM prompt.

\begin{prompt}{Prompt C.1: \texttt{direct-single} and \texttt{direct-16}}
<system>
You are an expert at estimating Q-value functions for reinforcement learning environments.

The Q-value Q(s,a) represents the expected discounted cumulative reward when taking action a in state s and then following the optimal policy thereafter. In other words, Q(s,a) is the expected return given that action a is taken in state s, assuming optimal play thereafter. It depends on both the state and the specific action taken.

Your goal is to produce an informed approximation -- not to compute the exact value. Enumerating all possible future trajectories is intractable. Instead, leverage your understanding of the environment dynamics, the reward structure, and the state's features to reason about how favorable the situation is. A well-reasoned estimate grounded in domain knowledge is more valuable than an attempt at exact computation.

All else being equal, reaching the goal in fewer steps is preferable -- states and actions that lead to shorter successful trajectories should generally be valued higher, reflecting the importance of efficiency.

You will receive the episode's state-action history (if any), the current state, the action taken, and the resulting next state. Estimate Q(s,a) for the **Current Action** in the **Current State** -- the expected discounted cumulative reward when taking this action in this state, assuming optimal play thereafter. The history provides context about how the agent reached the current state.

## Task Description
(*@\promptslot{task_description}@*)

## Episode Configuration
The episode step limit is (*@\promptslot{max_steps}@*).

## Reward Functions
(*@\promptslot{reward_description}@*)

After any reasoning, provide your numeric Q-value estimate as a single number inside <answer> tags, e.g. <answer>NUMBER</answer>.
</system>

<user>
Estimate the Q-value for the **Current Action** in the **Current State** for the following:

### State-Action History

[Note: Only the last (*@\promptslot{num_shown}@*) of (*@\promptslot{num_available}@*) turns are shown.]

[Turn (*@\promptslot{t}@*)/(*@\promptslot{max_steps}@*) - State]
(*@\promptslot{history_state_text}@*)

[Turn (*@\promptslot{t}@*)/(*@\promptslot{max_steps}@*) - Action]
(*@\promptslot{history_action_text}@*)

### Current State
[Turn (*@\promptslot{step}@*)/(*@\promptslot{max_steps}@*)]
(*@\promptslot{state_text}@*)

### Current Action
(*@\promptslot{action_text}@*)

### Next State
(*@\promptslot{next_state_text}@*)

Now, provide your Q-value estimate as a single number inside <answer> tags, e.g. <answer>NUMBER</answer>.
</user>
\end{prompt}
\label{app::prompt_direct_single}

\begin{prompt}{Prompt C.2: \texttt{direct-batched} and \texttt{direct-sequential}, changes to Prompt C.1}
<system>
(*@\textrm{\textit{\texttt{Direct-batched} adds, after the task and reward blocks:}}@*)
You may receive multiple datapoints in one request. Evaluate each datapoint independently and return one estimate per datapoint in the same order they appear.

(*@\textrm{\textit{\texttt{Direct-sequential} adds instead:}}@*)
You may receive multiple datapoints across successive user turns in this conversation. For each turn, evaluate only the datapoint shown in that turn, keep your earlier estimates fixed and mutually consistent, and return exactly one estimate for the current datapoint only.

(*@\textrm{\textit{\texttt{Direct-batched} also replaces the closing line with:}}@*)
After any reasoning, provide exactly (*@\promptslot{n}@*) numeric Q-value estimates as a comma-separated list inside <answer> tags, one per datapoint in the same order.
</system>

<user>
(*@\textrm{\textit{\texttt{Direct-batched}: a single user turn}}@*)
Estimate the Q-value for the **Current Action** in the **Current State** for each datapoint below. Return one estimate per datapoint in the same order.

## Datapoint (*@\promptslot{i}@*)

(*@\promptslot{datapoint_sections}@*)

Now, provide exactly (*@\promptslot{n}@*) numeric Q-value estimates as a comma-separated list inside <answer> tags, one per datapoint in the same order, e.g. <answer>VALUE_1, ..., VALUE_n</answer>.
</user>

<user>
(*@\textrm{\textit{\texttt{Direct-sequential}: turn }}\promptslot{i}\textrm{\textit{ of a multi-turn conversation}}@*)
## Datapoint (*@\promptslot{i}@*) of (*@\promptslot{n}@*)

Keep your earlier answers fixed and answer only for the datapoint below.

Estimate the Q-value for the **Current Action** in the **Current State** for the following:

(*@\promptslot{datapoint_sections}@*)

Now, provide your Q-value estimate as a single number inside <answer> tags, e.g. <answer>NUMBER</answer>.
</user>
\end{prompt}

\begin{completion}{Prompt C.2: replayed earlier turns of \texttt{direct-sequential}}
<assistant>
[thinking not shown]
<answer>(*@\promptslot{previous_estimate}@*)</answer>
</assistant>
\end{completion}

\begin{prompt}{Prompt C.3: \texttt{gvl}, shuffled-trajectory context}
<system>
You are an expert at estimating Q-value functions for reinforcement learning environments.

(*@\promptslot{q_value_definition}@*)

(*@\promptslot{approximation_guidance}@*)

(*@\promptslot{efficiency_guidance}@*)

You will receive the current state, the action taken, and the resulting next state. Estimate Q(s,a) for the **Current Action** in the **Current State** -- the expected discounted cumulative reward when taking this action in this state, assuming optimal play thereafter.

(*@\promptslot{task_and_reward_blocks}@*)

You will receive a shuffled set of surrounding transitions from the same trajectory, followed by one final target datapoint. The surrounding transitions are intentionally out of chronological order and are provided only as contextual evidence. Estimate only the final target datapoint.

Only the target datapoint includes its resulting next state. Surrounding context transitions include only state and action.

After any reasoning, provide your numeric Q-value estimate as a single number inside <answer> tags, e.g. <answer>NUMBER</answer>.
</system>

<user>
Estimate the Q-value for the **Current Action** in the **Current State** for the following:

## Shuffled Trajectory Context
The surrounding transitions below come from the same trajectory as the target datapoint, but are intentionally shuffled and not chronological.

### Context Transition (*@\promptslot{i}@*)
#### State
(*@\promptslot{context_state_text}@*)

#### Action
(*@\promptslot{context_action_text}@*)

## Target Transition

### Target State
(*@\promptslot{state_text}@*)

### Target Action
(*@\promptslot{action_text}@*)

### Target Next State
(*@\promptslot{next_state_text}@*)

Now, provide your Q-value estimate as a single number inside <answer> tags, e.g. <answer>NUMBER</answer>.
</user>
\end{prompt}
\label{app::prompt_gvl}

\begin{prompt}{Prompt C.4: \texttt{verifier}, score bins with log-probability readout}
<system>
You are an expert verifier for reinforcement learning environments. Estimate Q-value using an ordered discrete score scale.

(*@\promptslot{q_value_definition}@*)

(*@\promptslot{approximation_guidance}@*)

(*@\promptslot{efficiency_guidance}@*)

You will receive the episode's state-action history (if any), the current state, the action taken, and the resulting next state. Estimate Q(s,a) for the **Current Action** in the **Current State** -- the expected discounted cumulative reward when taking this action in this state, assuming optimal play thereafter. The history provides context about how the agent reached the current state.

Use an ordered 20-point score scale with the single-letter bins A through T, where A is best and T is worst. Choose exactly one score bin per datapoint.
A = clearly and completely favorable with strong evidence of success (best)
B-D = strongly favorable with only minor remaining concerns
E-G = above average, mostly favorable with some issues
H-J = uncertain, leans favorable
K-M = uncertain, leans unfavorable
N-P = below average, significant issues remain
Q-S = poor, with only limited signs of progress
T = clearly and completely unfavorable or failed (worst)

Interpret the score bins as an ordered scale that will be converted into a scalar value after decoding. Use higher bins for stronger evidence that the state or chosen action is favorable.

(*@\promptslot{task_and_reward_blocks}@*)

Return exactly one score tag for the single datapoint.
</system>

<user>
Estimate a verifier score for each datapoint below. Estimate the Q-value for the **Current Action** in the **Current State**. Judge the chosen action and resulting next state using the ordered score bins.

### Evaluation Criterion: (*@\promptslot{criterion_name}@*)
(*@\promptslot{criterion_description}@*)

## Datapoint 1

(*@\promptslot{datapoint_sections}@*)

Now provide your final scores using one uppercase letter from A through T per datapoint.
Output exactly one tag per datapoint in prompt order:
<score_1>LETTER</score_1>
</user>
\end{prompt}
\label{app::prompt_verifier}

\begin{prompt}{Prompt C.5: \texttt{\(\Delta\)belief}, pre and post belief queries}
<system>
You are an expert at judging how candidate actions change an agent's probability of eventual success in a sequential decision-making environment.

(*@\promptslot{q_value_definition}@*)

Interpret 'successful overall outcome' using the task and reward function below. For sparse-success tasks, this means eventually completing the task. For dense-reward tasks, this means achieving a strong overall return.

Each candidate is scored by the change in log-probability of the one-word answer 'Yes' to a success question after the candidate's observed outcome is revealed.

(*@\promptslot{approximation_guidance}@*)

(*@\promptslot{efficiency_guidance}@*)

(*@\promptslot{task_and_reward_blocks}@*)
</system>

<user>
(*@\textrm{\textit{pre prompt, one per state:}}@*)
Assess the current situation before any candidate outcome is shown.

(*@\promptslot{state_action_history}@*)

**Current State:**
(*@\promptslot{state_text}@*)

Do not answer directly. The assistant continuation being scored is the one-word answer to the success question below.

Question: Given the task and reward function, is the agent likely to eventually achieve a successful overall outcome from here, assuming optimal future decisions thereafter?
Answer with exactly one word: Yes or No.
</user>

<user>
(*@\textrm{\textit{post prompt, one per candidate:}}@*)
Re-assess the same situation after one specific candidate action has been attempted and its observed outcome is available.

(*@\promptslot{state_action_history}@*)

**Current State:**
(*@\promptslot{state_text}@*)

**Candidate Action:**
(*@\promptslot{candidate_action_text}@*)

**Observed Resulting State:**
(*@\promptslot{candidate_next_state_text}@*)

Do not answer directly. The assistant continuation being scored is the one-word answer to the success question below.

Question: Given the task and reward function, is the agent likely to eventually achieve a successful overall outcome from here, assuming optimal future decisions thereafter?
Answer with exactly one word: Yes or No.
</user>
\end{prompt}
\label{app::prompt_dbelief}

\begin{completion}{Prompt C.5: scored continuations}
<assistant>
Yes
</assistant>

<assistant>
No
</assistant>
\end{completion}

\begin{prompt}{Prompt C.6: \texttt{codegen} and \texttt{codegen-avg}, function generation}
<system>
You are an expert at writing Q-value functions for reinforcement learning environments.

(*@\promptslot{q_value_definition}@*)

(*@\promptslot{approximation_guidance}@*)

(*@\promptslot{efficiency_guidance}@*)

(*@\promptslot{task_and_reward_blocks}@*)
</system>

<user>
Write a Python function that estimates the Q-value for a given state, action, and next state.

The function signature must be:
```python
def signal_function(state: str, action: str, next_state: str) -> float:
```

The function receives text representations of the state, the action taken, and the resulting next state. It should return a float representing the estimated Q-value.

Important constraints:
- Do NOT use recursive search, tree expansion, lookahead, or any form of simulation/rollout -- these will time out.
- Base the estimate on direct analysis of the state representation, using your understanding of what features predict good outcomes.

You may use the `collections`, `itertools`, `json`, `math`, `re`, `statistics`, and `string` standard library modules. The function will run in a restricted sandbox that does not expose the introspection builtins `locals`, `globals`, `vars`, `dir`, `eval`, or `exec` -- calling any of them raises NameError at runtime, so do not rely on them (in particular, do not use `locals()` to assemble return values: build any dictionaries with explicit literal keys). After any reasoning, output ONLY a single Python code block containing the function definition. No explanations or commentary after the code block.
</user>
\end{prompt}
\label{app::prompt_codegen}

\begin{prompt}{Prompt C.7: \texttt{sdpo}, student and teacher contexts (TerminalBench)}
<system>
You are an expert at comparing and ranking actions by their Q-value in reinforcement learning environments.

(*@\promptslot{q_value_definition}@*)

You will evaluate one fixed candidate action at a time. The base policy scores the candidate from the original state. A self-teacher may additionally see environment feedback produced after that same action was attempted, and should use that feedback as per-turn evidence about whether the original action improved expected future success from the current state.

For TerminalBench, this SDPO signal is per-turn: the teacher may use only the immediate shell response after the candidate command. Treat that response as evidence about concrete progress toward the task and eventual verifier success, including useful file changes, targeted diagnostics, validation signals, and unresolved errors.

(*@\promptslot{approximation_guidance}@*)

(*@\promptslot{efficiency_guidance}@*)

(*@\promptslot{task_and_reward_blocks}@*)
</system>

<user>
(*@\textrm{\textit{student:}}@*)
Choose the next shell command for the following TerminalBench decision point. The shell session is persistent, so commands should make concrete progress through inspection, file creation or edits, validation, or targeted error diagnosis.

(*@\promptslot{state_action_history}@*)

**Current State:**
(*@\promptslot{state_text}@*)

Respond with exactly one shell command and no explanation. Prefer commands that produce durable task progress, useful diagnostics, or validation evidence rather than redundant probing or noisy output. The assistant continuation being scored is that command itself.

Output exactly one shell command. Output only that command, without explanation, evaluation, or surrounding tags. Preserve the command's required syntax.
</user>

<user>
(*@\textrm{\textit{teacher:}}@*)
Choose again from the same TerminalBench shell state after seeing the immediate shell response produced by one candidate command.

(*@\promptslot{state_action_history}@*)

**Current State:**
(*@\promptslot{state_text}@*)

**Immediate Shell Response After The Candidate Command:**
(*@\promptslot{candidate_next_state_text}@*)

Use this one-turn shell response only as evidence about the candidate command's quality in the current TerminalBench task: whether it made durable file or configuration progress, produced useful diagnostics, revealed or resolved an error, preserved shell state, avoided redundant probes, and moved toward eventual verifier success. A command with no output can still be good if it changed state usefully; an error can be useful only when it provides needed diagnostic information. Do not reward a command merely because its output makes the command easy to reconstruct. The assistant continuation being scored is the original candidate command itself, so do not rewrite or replace it here.

Output exactly one shell command. Output only that command, without explanation, evaluation, or surrounding tags. Preserve the command's required syntax.
</user>
\end{prompt}
\label{app::prompt_sdpo}

\begin{completion}{Prompt C.7: scored continuation, in both contexts}
<assistant>
(*@\promptslot{candidate_action_text}@*)
</assistant>
\end{completion}

\begin{prompt}{Prompt C.8: \texttt{sdpo-gt}, teacher context with stored expert evidence (TerminalBench)}
<system>
(*@\textrm{\textit{As in Prompt C.7, with the two paragraphs after the Q-value definition replaced by:}}@*)
You will evaluate one fixed candidate action at a time. The base policy scores the candidate from the original state. A self-teacher may additionally see environment feedback produced after that same action was attempted, the next expert action from the resulting state, and a compact stored expert rollout summary. It should use that privileged evidence to judge whether the original action improved expected future success from the current state.

For TerminalBench, this SDPO variant is an oracle-teacher ablation: the teacher may use the immediate shell response after the candidate command, the next expert shell command from the resulting persistent shell state, and a structured stored verifier outcome summary. Treat that evidence as information about durable task progress, useful diagnostics, validation signals, shell-state preservation, remaining distance to verifier success, and unresolved errors.
</system>

<user>
(*@\textrm{\textit{teacher (the student turn is that of Prompt C.7):}}@*)
Choose again from the same TerminalBench shell state after seeing the immediate shell response produced by one candidate command, the next expert command from the resulting state, and a structured verifier outcome summary from the stored expert rollout.

(*@\promptslot{state_action_history}@*)

**Current State:**
(*@\promptslot{state_text}@*)

**Immediate Shell Response After The Candidate Command:**
(*@\promptslot{candidate_next_state_text}@*)

**Expert Shell Rollout Evidence After The Candidate Command:**
The expert command below occurs after the candidate command has already been executed in the persistent TerminalBench shell session.
Use it only as evidence about the original candidate command; it is not a replacement command to output.
Best next expert command after candidate result: (*@\promptslot{next_expert_command}@*)
Structured stored expert rollout summary:
Stored expert rollout reached verifier success: (*@\promptslot{yes_or_no_success_observed}@*)
Expert shell steps from candidate result to success: (*@\promptslot{steps}@*)
Remaining distance proxy: (*@\promptslot{steps}@*) expert shell step(s) to verifier success
Final observed verifier reward: (*@\promptslot{final_reward}@*)
Stored expert rollout outcome: (*@\promptslot{success_or_no_success_observed}@*)

Use the immediate shell response, next expert command, and structured verifier outcome summary to judge the original candidate command from the current TerminalBench shell state. Prefer commands that make durable file or configuration progress, produce useful diagnostics or validation evidence, preserve needed shell state, reduce expert distance to verifier success, and avoid redundant probes or noisy dead ends. Treat a short successful stored expert rollout as strong positive evidence; treat no observed success as weaker evidence, not proof that the state is unsolvable. Do not reward a command merely because its output or the next expert command makes it easy to reconstruct. The assistant continuation being scored is the original candidate command itself, so do not rewrite or replace it.

Output exactly one shell command. Output only that command, without explanation, evaluation, or surrounding tags. Preserve the command's required syntax.
</user>
\end{prompt}
\label{app::prompt_sdpo_gt}

\begin{prompt}{Prompt C.9: \texttt{ranking}, candidate permutation}
<system>
You are an expert at comparing and ranking actions by their Q-value in reinforcement learning environments.

(*@\promptslot{q_value_definition}@*)

(*@\promptslot{approximation_guidance}@*)

(*@\promptslot{efficiency_guidance}@*)

You will receive the episode's state-action history (if any), the current state, a list of candidate actions, and the resulting next state for each action. Rank the actions by their expected Q-value Q(s,a) -- the expected discounted cumulative reward when taking each action. Output the action numbers from best to worst.

(*@\promptslot{task_and_reward_blocks}@*)

After any reasoning, output a comma-separated list of action numbers from best to worst inside <answer> tags, e.g. <answer>#, #, #, ...</answer>.
</system>

<user>
Rank the following actions by Q-value (best to worst) for the Current State:

(*@\promptslot{state_action_history}@*)

## Current State
[Turn (*@\promptslot{step}@*)/(*@\promptslot{max_steps}@*)]
(*@\promptslot{state_text}@*)

The following candidate actions are being evaluated in the current state:

## Actions to rank

[**Candidate Action (*@\promptslot{i}@*)**]
(*@\promptslot{candidate_action_text}@*)

[**Resulting State (*@\promptslot{i}@*)**]
(*@\promptslot{candidate_next_state_text}@*)

Rank all listed actions ((*@\promptslot{k}@*) total here) from best (highest Q-value) to worst. Output a comma-separated list of action numbers inside <answer> tags, e.g. <answer>#, #, #, ...</answer>.
</user>
\end{prompt}
\label{app::prompt_ranking}

\begin{prompt}{Prompt C.10: \texttt{eureka}, generation and judge prompts}
<system>
(*@\textrm{\textit{generation:}}@*)
You are an expert at writing Q-value functions for reinforcement learning environments.

(*@\promptslot{q_value_definition}@*)

(*@\promptslot{approximation_guidance}@*)

(*@\promptslot{efficiency_guidance}@*)

(*@\promptslot{task_and_reward_blocks}@*)

## Iterative Search
You are participating in an iterative code-search loop. Each proposed candidate will be executed on benchmark datapoints, judged from its predicted values, and the winning code plus feedback may be shown back to you in the next iteration. Improve usefulness of the signal based on behavior, not code aesthetics.
</system>

<user>
## Search Iteration
Iteration (*@\promptslot{iteration}@*) of (*@\promptslot{total_iterations}@*). Write one candidate Python function for estimating the Q-value.

## Function Contract
The function definition must start with:
```python
def signal_function(state: str, action: str, next_state: str):
```

The function may return either:
1. a single float, which is the actual signal value
2. a tuple `(float, dict[str, float])` where the first float is the actual signal value and the dictionary contains named additive components used only for search-time feedback.

If you return a dictionary, use short stable snake_case keys and make the scalar total equal to the sum of the component values. Build the components dictionary with explicit literal keys -- do NOT use `locals()`, `globals()`, or `vars()` (they are unavailable in the sandbox and raise NameError).

## Return Example
```python
def signal_function(state: str, action: str, next_state: str):
    progress_reward = 0.4
    safety_penalty = -0.1
    total = progress_reward + safety_penalty
    return total, {
        "progress_reward": progress_reward,
        "safety_penalty": safety_penalty,
    }
```

(*@\textrm{\textit{from the second iteration on:}}@*)
## Current Best Code
This is the current best candidate from the previous iteration. Use it as a starting point and improve it.

(*@\promptslot{current_best_code}@*)

## Judge Feedback
### Judge Rationale
(*@\promptslot{judge_rationale}@*)

### Improvement Feedback
(*@\promptslot{judge_feedback}@*)

(*@\textrm{\textit{always:}}@*)
## Constraints
(*@\promptslot{codegen_constraints_and_closing}@*)
</user>

<system>
(*@\textrm{\textit{judge:}}@*)
You are selecting the most useful dense-signal candidate for a reinforcement learning environment.

(*@\promptslot{q_value_definition}@*)

Judge candidate behavior from predicted values and diagnostics. Do not infer quality from code style because candidate code is not shown.

(*@\promptslot{task_and_reward_blocks}@*)
</system>

<user>
## Selection Rule
Choose the candidate whose predictions are the most appropriate and useful Q-value estimates. Prefer candidates that provide discriminative, stable, non-collapsed values and that match the intended semantics of the signal on the sampled datapoints.

## Candidate Summaries

### Candidate (*@\promptslot{candidate_label}@*)
- Finite predictions: (*@\promptslot{finite_count}@*)
- NaN predictions: (*@\promptslot{nan_count}@*)
- Min: (*@\promptslot{min}@*)
- Max: (*@\promptslot{max}@*)
- Mean: (*@\promptslot{mean}@*)
- Std: (*@\promptslot{std}@*)
- Collapsed: (*@\promptslot{yes_or_no}@*)

## Sampled Datapoints

### Datapoint (*@\promptslot{i}@*)

(*@\promptslot{datapoint_sections}@*)

### Candidate Predictions
Candidate (*@\promptslot{candidate_label}@*): (*@\promptslot{prediction}@*)

## Output Format
Respond with:
<winner>INTEGER</winner>
<rationale>Short explanation of why the winner is best.</rationale>
<feedback>Concrete guidance for improving the next iteration.</feedback>
</user>
\end{prompt}
\label{app::prompt_eureka}

\clearpage
\section{Model details}
\label{app::models}

\paragraph{LLM backbones.}
We use six instruction-tuned open-weight LLMs (Table~\ref{tab:llm-sampling-settings}) in two serving roles.
\emph{Generation} methods (\texttt{direct-*}, \texttt{gvl}, \texttt{ranking}, \texttt{codegen}, \texttt{codegen-avg} and \texttt{eureka}) query the models with thinking enabled and a thinking budget of $6{,}144$ tokens. The maximum generation length is $8{,}192$ tokens, or $14{,}336$ tokens for the code methods. If a response reaches this limit without an answer, a single follow-up request, with the truncated response in context, asks for the final answer within $512$ tokens ($2{,}048$ for code). Prompts are budgeted for a context length of $262{,}144$ tokens.
\emph{Likelihood} methods (\texttt{verifier}, \texttt{$\Delta$belief}, \texttt{sdpo} and \texttt{sdpo-gt}) need token log-probabilities; they run with thinking disabled and a context length of $32{,}768$ tokens. \texttt{$\Delta$belief}, \texttt{sdpo} and \texttt{sdpo-gt} only score fixed continuations, so no sampling is needed. The \texttt{verifier} method generates its answer at temperature $0.1$ and top-$p$ $1.0$, with at most $4{,}096$ tokens, and reads the top-$20$ token log-probabilities at the score position.

\begin{table}[h]
\centering
\small
\begin{tabular}{lccc}
\toprule
Model & Temp. & Top-$p$ & Top-$k$ \\
\midrule
\texttt{google/gemma-4-31b-it} & 1.0 & 0.95 & 64 \\
\texttt{google/gemma-4-26b-a4b-it} & 1.0 & 0.95 & 64 \\
\texttt{qwen/qwen3.5-9b} & 0.6 & 0.95 & 20 \\
\texttt{qwen/qwen3.5-27b} & 0.6 & 0.95 & 20 \\
\texttt{qwen/qwen3.5-35b-a3b} & 0.6 & 0.95 & 20 \\
\texttt{qwen/qwen3.5-122b-a10b} & 0.6 & 0.95 & 20 \\
\bottomrule
\end{tabular}
\caption{LLM backbones used in \benchname experiments. Temperature, top-$p$ and top-$k$ apply to the generation methods; the \texttt{verifier} decodes at temperature $0.1$ and top-$p$ $1.0$.}
\label{tab:llm-sampling-settings}
\end{table}

\paragraph{Encoders.}
The embedding methods use \texttt{openai/clip-vit-large-patch14} (CLIP ViT-L/14) and \texttt{google/siglip-base-patch16-224} (SigLIP base). \texttt{Vip} uses the ResNet-50 \texttt{vip} checkpoint pretrained on Ego4D, and the \texttt{liv-*} methods use the \texttt{liv} checkpoint (a CLIP ResNet-50 fine-tuned on EpicKitchens). All encoders are frozen.

\paragraph{Models used to build the benchmark.}
These models are not evaluated. DeepSeek v3.2 collects the TerminalBench and ALFWorld trajectories and proposes their alternative candidate actions (Appendix~\ref{app::environments}). GPT-5.5, run with low reasoning effort, provides the TerminalBench reference rollouts, and Claude Opus 4.7 provides the alternative reference in the reference-policy ablation.

\clearpage
\section{Guided search details}
\label{app::guided_search}

\subsection{Online evaluation}

We use the same actor for unguided and guided runs. The unguided baseline executes one sampled action per step. For guided search, the actor proposes four actions independently and we select to continue with the one that is scored highest by the evaluated signal. Duplicate and invalid proposals are retained; we do not prompt for forbidden actions or resample to enforce diversity. Top-score ties are broken uniformly over proposal slots.

\begin{table}[h]
\centering\small
\caption{Guided search settings. Both environments use text observations and discount factor $\gamma=0.95$.}
\label{tab:guided_search_setup}
\begin{tabular}{lcc}
\toprule
 & FrozenLake & ALFWorld \\
\midrule
Actor / general-purpose scorer & Gemma 4 31B & Qwen3.5 27B \\
Episode limit & 15 steps & 9 steps \\
History turns & 0 & 40 \\
Tasks and repeats per policy & 100 maps $\times$ 1 & 24 games $\times$ 3 \\
Actor temperature / top-$p$ / top-$k$ & 1.0 / 0.95 / 64 & 1.0 / 0.95 / 20 \\
Actor thinking / completion budget & enabled / 8,192 & enabled / 8,192 \\
Proposals per guided decision & 4 & 4 \\
\bottomrule
\end{tabular}
\end{table}

The five signals are \texttt{direct-single}, \texttt{ranking}, \texttt{verifier}, \texttt{codegen-avg} and \texttt{sdpo}. The backend LLM used for the methods matches the actor. \texttt{Codegen-avg} uses one fixed ensemble of 16 generated functions throughout each online run. Matched offline scoring reuses those exact functions, rather than sampling another function set. The actor has a 6,144-token thinking budget and additionally a 512-token follow-up budget.

\begin{table}[htbp]
\centering\small
\caption{\textbf{Candidate diversity in guided search.} Each decision has four independently sampled proposals. We count distinct parsed/resolved actions, so different reasoning texts with the same command count once. The stored invalid-action marker counts as one proposal type. Percentages give the fraction of decisions with each distinct-proposal count. All decisions from each guided run are included; proposals are not removed or resampled.}
\label{tab:guided_search_candidate_diversity}
\setlength{\tabcolsep}{4pt}
\begin{tabular}{llrrrrrr}
\toprule
Environment & Method & Decisions & Mean distinct & 1 (\%) & 2 (\%) & 3 (\%) & 4 (\%) \\
\midrule
FrozenLake & \texttt{direct-single} & 1,378 & 1.44 & 59.4 & 37.2 & 3.4 & 0.0 \\
 & \texttt{ranking} & 1,405 & 1.47 & 56.3 & 40.4 & 3.2 & 0.1 \\
 & \texttt{verifier} & 1,280 & 1.42 & 61.3 & 35.6 & 2.7 & 0.3 \\
 & \texttt{codegen-avg} & 1,383 & 1.50 & 55.5 & 39.6 & 4.9 & 0.1 \\
 & \texttt{sdpo} & 1,318 & 1.50 & 54.8 & 40.6 & 4.6 & 0.0 \\
\midrule
ALFWorld & \texttt{direct-single} & 442 & 1.49 & 66.7 & 19.5 & 12.2 & 1.6 \\
 & \texttt{ranking} & 440 & 1.48 & 64.5 & 24.1 & 10.0 & 1.4 \\
 & \texttt{verifier} & 446 & 1.48 & 65.9 & 21.7 & 11.0 & 1.3 \\
 & \texttt{codegen-avg} & 490 & 1.46 & 65.9 & 23.3 & 10.2 & 0.6 \\
 & \texttt{sdpo} & 507 & 1.61 & 57.8 & 26.4 & 13.0 & 2.8 \\
\bottomrule
\end{tabular}
\end{table}

\subsection{Matching reference values, scoring settings and data}
\label{app::matching_analysis}

We distinguish three matching levels on the original benchmark: (i) original scores and references; (ii) the same scores but reference values recomputed for the guided search horizon setting; and (iii) the same reference values with method scores recomputed under the guided search settings. Matching the horizon changes reference labels as well as horizon information available in the scoring prompt. We then evaluate the matched methods on a shared dataset built from the accepted online trajectories. This changes the state--action distribution, including both visited states and candidate groups.

\begin{table}[t]
\centering\small
\caption{Groups with valid Spearman correlations on the shared matched dataset. All 8,100 FrozenLake and 2,817 ALFWorld groups contribute to primary selection loss, including reference-tied groups.}
\label{tab:guided_search_coverage}
\begin{tabular}{llr}
\toprule
Environment & Method & Spearman groups \\
\midrule
FrozenLake & \texttt{direct-single} & 1,030 \\
 & \texttt{ranking} & 1,207 \\
 & \texttt{verifier} & 1,226 \\
 & \texttt{codegen-avg} & 1,226 \\
 & \texttt{sdpo} & 1,226 \\
\midrule
ALFWorld & \texttt{direct-single} & 256 \\
 & \texttt{ranking} & 257 \\
 & \texttt{verifier} & 257 \\
 & \texttt{codegen-avg} & 234 \\
 & \texttt{sdpo} & 257 \\
\bottomrule
\end{tabular}
\end{table}

\begin{figure}[t]
\centering
\includegraphics[width=\linewidth]{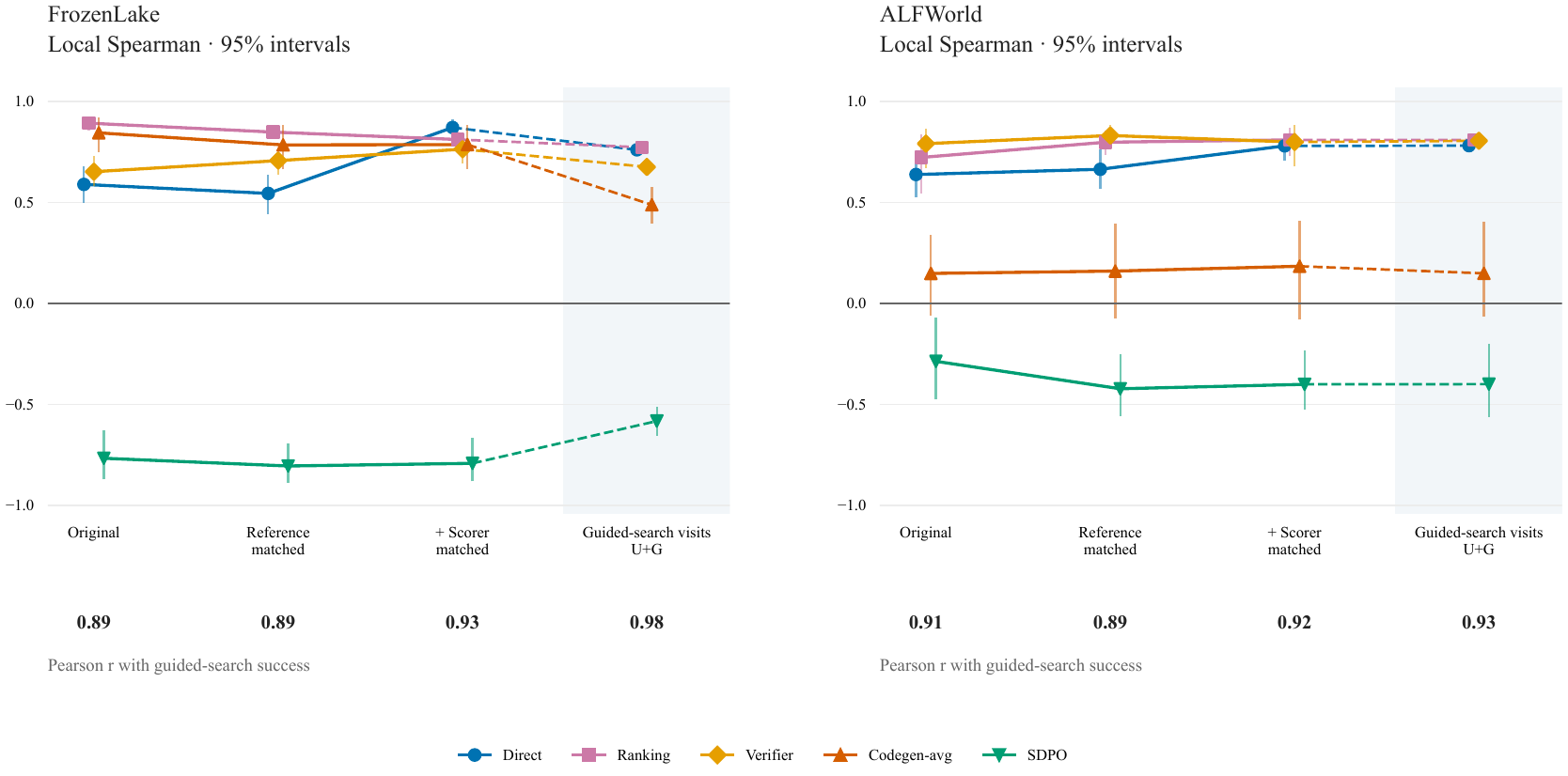}
\caption{\textbf{Matching settings and the state--action distribution.} Each method is retained separately. The matching levels change references and then method settings; the final comparison uses the fully matched setup. Higher values indicate stronger state-local Spearman alignment. "U" is unguided and "G" is guided.}
\label{fig:guided_search_matching}
\end{figure}

\subsection{Reference ties}
\label{app::reference_ties}

On the shared ALFWorld data, all 257 informative groups have ties: 182 have a $1+3$ tie pattern and 75 a $2+2$ pattern. The \texttt{ranking} method's maximum Spearman is therefore $0.81$; \texttt{ranking} attains this ceiling by ordering every unequal-reference pair correctly.

\par\medskip
\noindent\begin{minipage}{\linewidth}
\centering
\includegraphics[width=\linewidth]{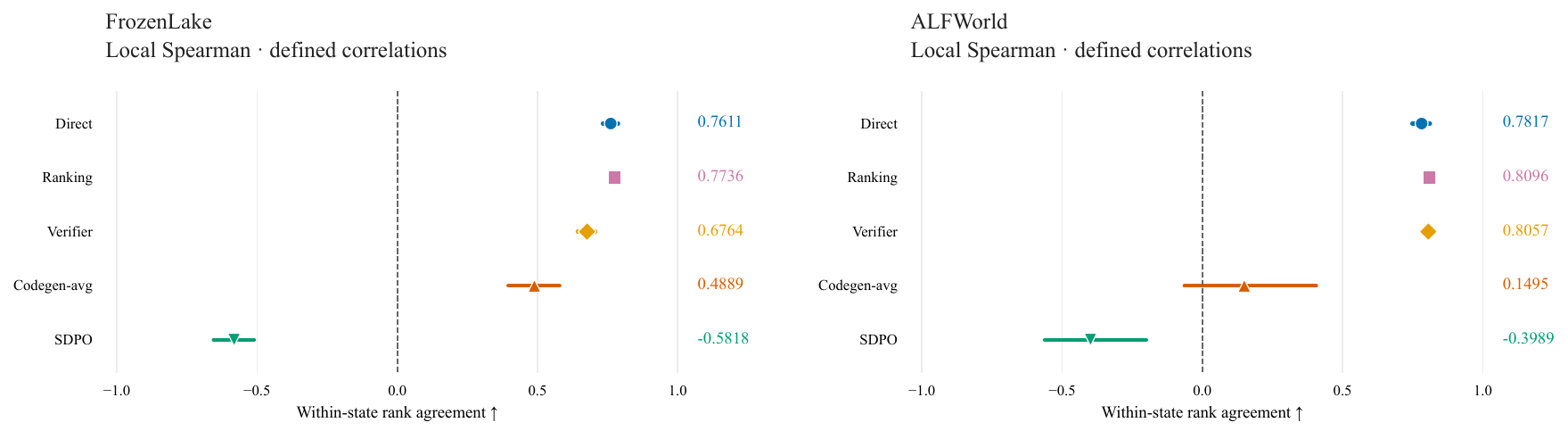}
\captionof{figure}{\textbf{Spearman on the shared matched dataset.}}
\label{fig:guided_search_ordering}
\end{minipage}\par\medskip

\begin{samepage}
\subsection{Decision costs}
\label{app:selection_loss}

Selection loss includes every fully labelled group, including tied groups with zero loss. It complements correlation by measuring the value sacrificed at the chosen action rather than only ordering agreement. On the matched original FrozenLake benchmark, \texttt{codegen-avg} and \texttt{verifier} have similar Spearman means ($0.787$, $0.764$), but mean losses of $0.084$ and $0.019$. On the matched dataset, their Spearman means become $0.489$ and $0.676$. Table~\ref{tab:guided_search_selection_loss} reports selection loss for all five methods on the shared matched dataset.
\par
\end{samepage}

\begin{table}[h]
\centering\small
\caption{Pearson correlation between each metric and guided search success across five methods, on the matched dataset.}
\label{tab:guided_search_metric_agreement}
\begin{tabular}{lrr}
\toprule
Metric & FrozenLake & ALFWorld \\
\midrule
Local Spearman & 0.976 & 0.926 \\
Selection loss & $-0.977$ & $-0.953$ \\
\bottomrule
\end{tabular}
\end{table}

\begin{table}[htbp]
\centering\small
\caption{\textbf{Selection loss on the shared matched dataset.} The means selection loss for each method--environment combination (lower is better), with 95\% task/episode bootstrap confidence intervals from 2,000 draws. Method and environment settings match guided search. Means include all 8,100 FrozenLake and 2,817 ALFWorld candidate groups, including groups with tied reference values.}
\label{tab:guided_search_selection_loss}
\setlength{\tabcolsep}{7pt}
\begin{tabular}{lrrrr}
\toprule
 & \multicolumn{2}{c}{FrozenLake} & \multicolumn{2}{c}{ALFWorld} \\
\cmidrule(lr){2-3}\cmidrule(lr){4-5}
Method & Mean & 95\% CI & Mean & 95\% CI \\
\midrule
\texttt{direct-single} & $0.00215$ & $[0.00161,\,0.00274]$ & $0.00029$ & $[0.00007,\,0.00072]$ \\
\texttt{ranking} & $0.00140$ & $[0.00077,\,0.00211]$ & $0.00000$ & $[0.00000,\,0.00000]$ \\
\texttt{verifier} & $0.00684$ & $[0.00499,\,0.00885]$ & $0.00029$ & $[0.00000,\,0.00115]$ \\
\texttt{codegen-avg} & $0.02360$ & $[0.01833,\,0.02938]$ & $0.00495$ & $[0.00255,\,0.00779]$ \\
\texttt{sdpo} & $0.03779$ & $[0.03303,\,0.04306]$ & $0.00728$ & $[0.00349,\,0.01169]$ \\
\bottomrule
\end{tabular}
\end{table}


\end{document}